\documentclass[10pt,journal,compsoc]{IEEEtran}

\usepackage{amsmath,amsfonts}
\usepackage{algorithmic}
\usepackage{algorithm}
\usepackage{array}
\usepackage[caption=false,font=normalsize,labelfont=sf,textfont=sf]{subfig}
\usepackage{textcomp}
\usepackage{stfloats}
\usepackage{url}
\usepackage{verbatim}
\usepackage{graphicx}

\usepackage{microtype}
\usepackage{amsmath}
\usepackage{amsthm}
\usepackage{mwe}
\usepackage{float}
\usepackage{array}
\usepackage[table]{xcolor}
\usepackage{threeparttable}
\usepackage{hyperref}

\usepackage{makecell}
\usepackage{amssymb}
\usepackage{booktabs}
\usepackage{multirow}
\usepackage{xspace}
\usepackage{bm}
\usepackage{epsfig}
\usepackage{color}
\usepackage{comment}
\usepackage{arydshln}
\usepackage{eqparbox}
\usepackage{subcaption}
\usepackage{pifont}
\usepackage{enumitem}
\ifCLASSOPTIONcompsoc
  \usepackage[nocompress]{cite}
\else
  \usepackage{cite}
\fi

\ifCLASSINFOpdf

\else

\fi

\begin{document}

\title{Human Action Anticipation: A Survey}

\author{Bolin Lai*, Sam Toyer*, Tushar Nagarajan, Rohit Girdhar, Shengxin Zha, James M. Rehg, \\ Kris Kitani, Kristen Grauman, Ruta Desai, Miao Liu

\IEEEcompsocitemizethanks{


\IEEEcompsocthanksitem * Equal contribution.

\IEEEcompsocthanksitem Bolin Lai is with Georgia Institute of Technology, Atlanta, GA 30308, USA (E-mail: bolin.lai@gatech.edu).

\IEEEcompsocthanksitem Sam Toyer is with University of California, Berkeley, CA 94720, USA (E-mail: sdt@berkeley.edu).

\IEEEcompsocthanksitem Tushar Nagarajan, Rohit Girdhar, Shengxin Zha, Ruta Desai, and Miao Liu are with Meta, Menlo Park, CA 94025, USA (E-mail: tusharn@meta.com, rgirdhar@meta.com, szha@meta.com, rutadesai@meta.com, miaoliu@meta.com).

\IEEEcompsocthanksitem James M. Rehg is with University of Illinois Urbana-Champaign, Urbana, IL 61801, USA (E-mail: jrehg@illinois.edu).

\IEEEcompsocthanksitem Kris Kitani is with Carnegie Mellon University, Pittsburgh, PA 15213, USA (E-mail: kmkitani@andrew.cmu.edu).

\IEEEcompsocthanksitem Kristen Grauman is with University of Texas at Austin, Austin, TX 78712, USA (E-mail: grauman@cs.utexas.   edu).
}
}

\markboth{Journal of \LaTeX\ Class Files,~Vol.~14, No.~8, August~2015}%
{Shell \MakeLowercase{\textit{et al.}}: Bare Demo of IEEEtran.cls for Computer Society Journals}

\IEEEtitleabstractindextext{%
\begin{abstract}
Predicting future human behavior is an increasingly popular topic in computer vision, driven by the interest in applications such as autonomous vehicles, digital assistants and human-robot interactions.  The literature on behavior prediction spans various tasks, including action anticipation, activity forecasting, intent prediction, goal prediction, and so on. Our survey aims to tie together this fragmented literature, covering recent technical innovations as well as the development of new large-scale datasets for model training and evaluation. We also summarize the widely-used metrics for different tasks and provide a comprehensive performance comparison of existing approaches on eleven action anticipation datasets. This survey serves as not only a reference for contemporary methodologies in action anticipation, but also a guideline for future research direction of this evolving landscape.
\end{abstract}

}

\maketitle

\IEEEdisplaynontitleabstractindextext

\IEEEpeerreviewmaketitle

\IEEEraisesectionheading{\section{Introduction}\label{sec:introduction}}
\IEEEPARstart{T}{he} problem of predicting the actions a human will take in the future has been studied extensively by the computer vision community, driven by its crucial applications across various domains such as robotics, augmented reality (AR) and surveillance. Existing studies cover a diverse set of action anticipation problems, such as predicting a categorical label representing the next action \cite{pei2011parsing,lan2014hierarchical,han2017human}, predicting both labels and durations of actions \cite{abu2018will,gong2022future,nawhal2022rethinking} or even generates natural language descriptions of future actions \cite{sener2019zero,abdelsalam2023gepsan}. In this survey, we aim to tie together the fragmented literature of human behavior forecasting, summarize the themes of existing research, review the technical development of proposed approaches and identify important gaps to be addressed by future work.

Our survey is timely for two reasons.
The first is a strong \textit{application pull}, in which emerging product categories could benefit from better action anticipation methods. The second is a firm \textit{technical push}, where the diverse anticipative tasks, new datasets and methods have driven rapid progress in academic research on action anticipation.

When it comes to application pull, we see new categories of robots and digital assistants that can better assist or protect humans by forecasting their behaviors. For instance, Mixed reality (XR) devices can benefit from the forecasting capability by calculating and pre-loading virtual objects and scenes before users take actions \cite{damen2018scaling}, which significantly reduces latency. An autonomous vehicle needs to keep pedestrians safe by forecasting when they might step out onto the road, and avoid colliding with other cars by predicting when they might turn or change lanes \cite{rasouli2020pedestrian,liu2020spatiotemporal,girase2021loki}. Anticipation can also enable more fluid social interaction between humans and robots, like inferring that a person who has started extending their hands wants to have a handshake \cite{barquero2022didn}.

\begin{figure*}
    \centering
    \includegraphics[width=\textwidth]{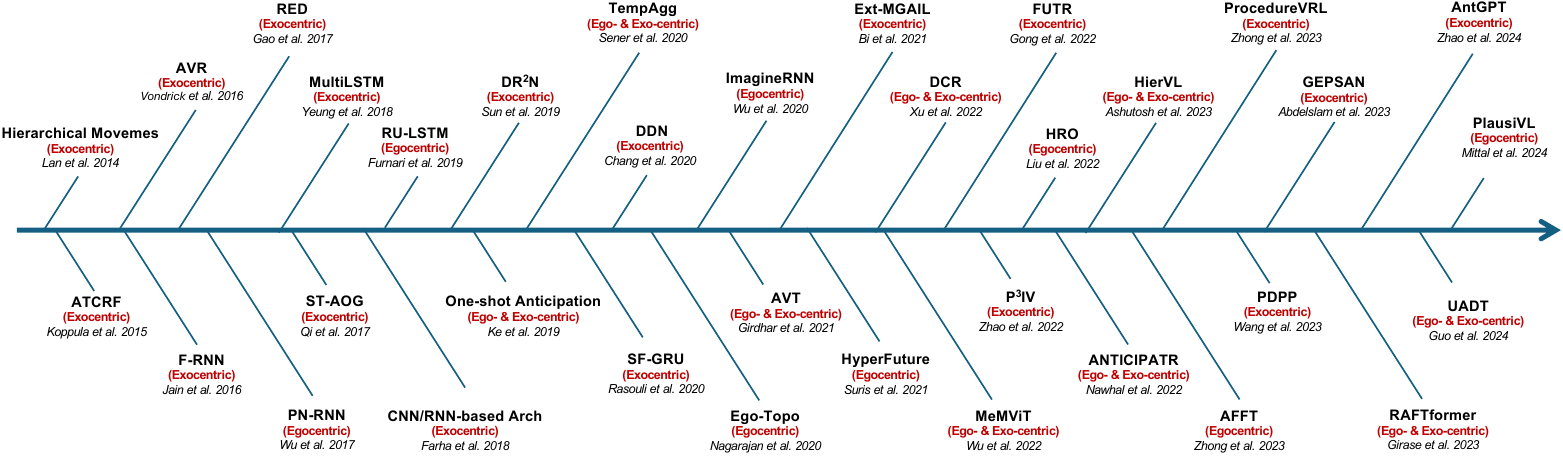}
    \caption{A chronological overview of existing work in action anticipation. We only cover the approaches that make important technical breakthrough or have high impacts in this field. Please refer to Table \ref{tab:action-methods} and Table \ref{tab:action-methods-2} for a thorough review.}
    \label{fig:highlighted_methods}
\end{figure*}

In tandem with the pull of new applications, we have also seen a technical push towards addressing action anticipation on various video datasets with more capable action anticipation methods. Prior works addressed action anticipation problem in both egocentric (first-person) and exocentric (third-person) perspectives. Two most important egocentric action anticipation challenges arise from Epic-Kitchens \cite{damen2019ekchallenges,damen2020ekchallenges,damen2021ekchallenges,damen2022ekchallenges} (2019-present) and Ego4D \cite{grauman2022ego4dcvpr,girdhar2022ego4deccv} (2022-present). The large scale, dense annotations and high diversity in the two datasets have marked a turning point in this field. Prominent models include RU-LSTM \cite{furnari2019would,furnari2020rolling}, AFFT \cite{zhong2023anticipative}, PlausiVL \cite{mittal2024can}, etc. Exocentric action anticipation has a longer studying history because of the early availability of exocentric video datasets, such as 50Salads \cite{stein2013combining}, Breakfast Actions \cite{kuehne2014language}, THUMOS \cite{THUMOS14}, etc. The models in this field span across SVM-based approaches \cite{lan2014hierarchical}, RNN-based sequence models \cite{jain2016recurrent,yeung2018every}, transformer architectures \cite{gong2022future,abdelsalam2023gepsan} and large language models \cite{zhao2024antgpt}. We highlight some important action anticipation methods of each year in Figure \ref{fig:highlighted_methods} to demonstrate the development of egocentric and exocentric action anticipation over time. 

The two closest surveys to ours are the work of Rodin et al. \cite{rodin2021predicting} (2021) and Zhong et al. \cite{zhong2023survey} (2023). Rodin et al. \cite{rodin2021predicting} survey methods for behavior prediction in egocentric videos. Unlike this paper, they include discussion on applications of and hardware for egocentric video recording and survey a broader range of tasks, including trajectory, gaze and interaction tasks.
They limit their focus to egocentric data and methods evaluated on egocentric data, and thus consider only a subset of the methods that we cover (see Table \ref{tab:action-methods} and Table \ref{tab:action-methods-2}). Zhong et al. \cite{zhong2023survey} broaden the scope to deep learning methods applied to both egocentric and exocentric action anticipation. They categorize the prior methods by the strategies, thus ignoring the nuances in the settings of different fine-grained action anticipation tasks and leaving out several related directions (e.g., procedure planning). 

Compared to recent surveys on action anticipation~\cite{rodin2021predicting,zhong2023survey,kong2022human,barquero2022didn,rasouli2020deep}, we provide a formal classification and definition of existing action anticipation tasks based on their anticipative goals and timespans, offering a more structured framework for understanding the field. Furthermore, we conduct an in-depth analysis of existing works for each task, grouping them according to key insights such as inductive biases in model design, input modalities, and training objectives.

There are also many surveys on topics related to action anticipation, including surveys on action recognition and surveys on forecasting of other types of behavioral data.
Herath et al. \cite{herath2017going} (2017) provide an engaging overview of past work in action recognition.
Earlier work from Poppe et al. \cite{poppe2010survey} (2010) provides useful coverage of techniques before the rise of deep learning, while Sun et al. \cite{sun2022human} (2022) cover the most recent work in this area with a particular emphasis on different input modalities.
In addition, relevant yet vastly distinctive work of trajectory forecasting and pose prediction lies beyond the scope of this survey. We refer readers to Rudenko et al. \cite{rudenko2020human} (2020) for coverage of the former and Lyu et al. \cite{lyu20223d} (2022) for the latter.
Our survey will focus only on tasks that directly predict human's future actions or goals, with emphasis on the fine-grained task settings (e.g. definition of input and output) and model settings (e.g., anticipation time and viewpoint).

We briefly outline some of the notations that we use throughout later sections in Table \ref{tab:notation}. There are also some ambiguous terminology in prior papers that may cause misunderstanding. We clarify their definition in this survey which is listed below.

\begin{itemize}
\item ``action'' and ``activity'': We use the world ``action'' to broadly refer to a thing that a person might do for some purpose. Some past work uses ``activity'' as a synonym, or sometimes to denote a higher-level kind of action. In this survey, we use ``activity'' as a high-level concept of actions, where an activity typically consists of many fine-grained actions.

\item ``goal'' and ``intent'': We take a broad view of what a ``goal'' is, and use the term to encompass any objective that a person might aim to achieve. The boundary between a goal and an action is fuzzy, but we use ``goal'' to describe the end state after a series of actions, meaning that goals are usually more high-level than actions. We avoid the word ``intent'' in the following sections, since it is sometimes used as a synonym for ``goal'' or ``desired outcome'' and sometimes as a synonym for ``action that is about to be executed'' \cite{rasouli2019pie}.
\end{itemize}

\begin{table}[t]
\renewcommand{\arraystretch}{1.1}
\caption{Notation used in this paper.}\label{tab:notation}
\centering
\begin{tabular}{cl}
\toprule
\textbf{Symbol}  & \textbf{Meaning} \\
\midrule
$\mathcal V$ & Video \\
$T$ & Number of frames in video \\
$N$ & Number of total actions in video \\
$Z$ & Number of actions to be predicted \\
$a_i$ & Label of the $i$-th action segment \\
$t_i$/$t'_i$ & Start/end time of action $a_i$ \\
$o_t$ & $t$-th observed frame \\
$x_t$ & Latent representation of frame $o_t$ \\
$u_t$ & Latent representation of action $a_t$ \\
$\phi$ & Frame representation encoder \\
$\tau$ & Future time offset (usually in seconds) \\
$s_i$ & The $i$-th state in procedure planning \\
\bottomrule
\end{tabular}
\vspace{-0.3cm}
\end{table}

The structure of this survey is as follows. Section \ref{sec:action-tasks} defines our taxonomy of existing action anticipation tasks and covers existing literature in fine details. Section \ref{sec:datasets} describes popular datasets for action anticipation, and assesses their suitability for different kinds of anticipation tasks. In Section \ref{sec:future}, we come up with many promising directions for future research based on the defects of existing work. Finally, we conclude the survey in Section \ref{sec:conclusion}.

\begin{figure*}[h!]
    \centering
    \vspace{15pt}
    \includegraphics[width=\textwidth]{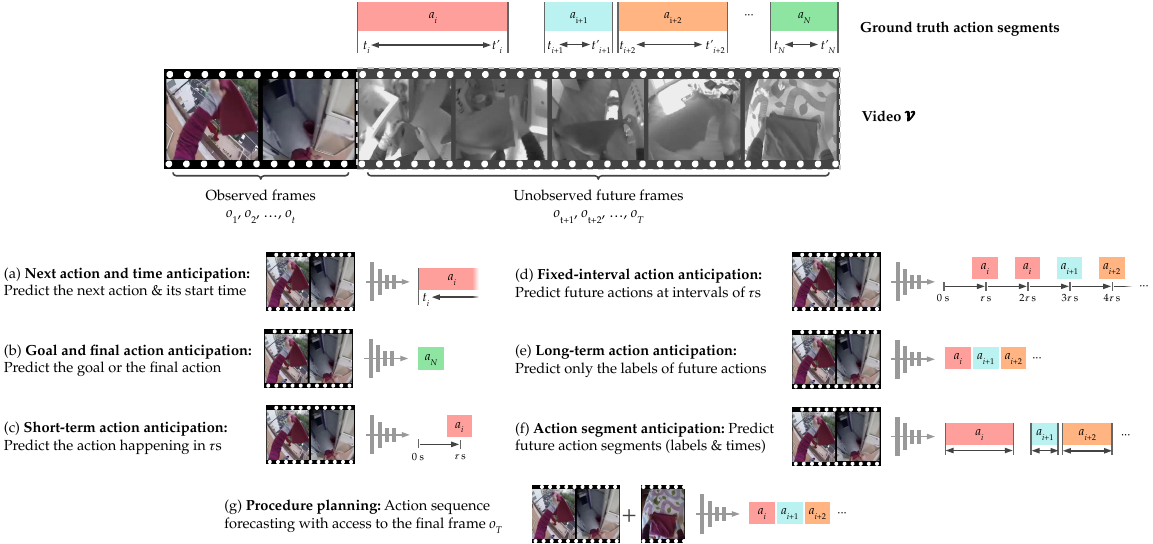}
    \caption{
        Taxonomy of action anticipation tasks. Each of the tasks covered in this paper involves predicting a different portion of the future action segment information (action labels, time or descriptions) for each annotated video. We break down the action anticipation problem into seven fine-grained tasks and show visual depiction of the input/output spec for each task.
        Corresponding notations are explained in Table \ref{tab:notation}, and each task is covered in detail in Section \ref{sec:action-tasks}.
    }
    \label{fig:action-tasks}
\end{figure*}

\begin{table*}[!h]
\scriptsize
\renewcommand{\arraystretch}{1.25}
\caption{
    Comparison of methods in Section \ref{sec:action-tasks}, which continues in Table \ref{tab:action-methods-2}.
    \textit{Tasks} include
    next action or time anticipation (\textbf{NAT}),
    goal and final action prediction (\textbf{GFA}),
    short-term action anticipation (\textbf{STA}),
    fixed-interval action anticipation (\textbf{FIA}),
    long-term action anticipation (\textbf{LTA}),
    action segment anticipation (\textbf{AS}),
    and procedural planning (\textbf{PP}).
    The ``Cat.'' column lists the sub-category, which is either inductive biases (\textbf{IB}), multi-modal models (\textbf{MM}), or pretraining and objectives (\textbf{PO}). \\
    \textit{Datasets} include
    50 Salads (\textbf{50S}),
    Activities of Daily Living (\textbf{ADL}),
    ActivityNet (\textbf{AN}),
    ACTICIPATE (\textbf{ACT}),
    Assembly101 (\textbf{As101}),
    \textbf{AVA},
    BDD100k (\textbf{BDD}),
    Breakfast Actions (\textbf{BA}),
    Charades (\textbf{Ch}),
    \textbf{COIN},
    CrossTask (\textbf{CT}),
    EGTEA Gaze+ (\textbf{EG+}),
    \textbf{Ego4D},
    Epic Kitchens 55 (\textbf{EK55}),
    FineGym (\textbf{FGym}),
    GTEA Gaze (\textbf{GG}),
    GTEA Gaze+ (\textbf{GG+}),
    Ikea Furniture Assembly (\textbf{IFA}),
    \textbf{JAAD},
    Kinetics 400/600/700 (\textbf{K400/K600/K700}),
    Meccano (\textbf{Me}),
    MPII Cooking (\textbf{MC}),
    MPII Cooking 2 (\textbf{MC2}),
    MultiTHUMOS (\textbf{MTH}),
    Narrated Instruction Videos (\textbf{NIV}),
    NCAA Basketball (\textbf{NB}),
    Oops! (\textbf{Oops}),
    Pedestrian Intention Estimation (\textbf{PIE}),
    Recipe1M (\textbf{R1M}),
    Stanford-ECM (\textbf{SE}),
    \textbf{STIP},
    TACoS-ML (\textbf{TM}),
    Tasty Videos (\textbf{Ta}),
    THUMOS (\textbf{TH}),
    TV Human Interaction (\textbf{TVHI}),
    TV Series (\textbf{TVS}),
    VIRAT (\textbf{V}),
    Brains4Cars (\textbf{B4C}),
    and YouCook2 (\textbf{YC2}).\\
    \textit{Ctx} and \textit{Anti} refer to the range of context length (observation time) and range of anticipation time of the method. \\
    \textit{Input modalities} include
    RGB video (\textbf{R}),
    flow (\textbf{F}),
    object detections (\textbf{O}),
    audio (\textbf{A}),
    textual narration (\textbf{T}),
    past action segments (\textbf{PA}),
    poses (\textbf{Po}),
    detected people (\textbf{DP}),
    SLAM/SfM map (\textbf{SM}),
    hand masks (\textbf{H}),
    gaze (\textbf{G}),
    and other proprioceptive data (\textbf{Pr}). \\
    \textit{Auxiliary training tasks} include
    action recognition (\textbf{AR}),
    action segmentation (\textbf{AS}),
    cycle consistency (\textbf{CC}),
    temporal contrastive learning (\textbf{TC}),
    temporal ordering prediction (\textbf{TO}),
    intentionality prediction (\textbf{IT}),
    consistency between branches (\textbf{CB}),
    early-layer prediction (\textbf{EP}),
    frame order recovery (\textbf{FO}),
    frame forecasting (\textbf{FF}),
    bag-of-actions forecasting (\textbf{BA}),
    moveme classification (\textbf{MC}),
    goal prediction (\textbf{GP}),
    autoencoding (\textbf{AE}),
    captioning (\textbf{Ca}),
    predicting gaze-object overlap (\textbf{GO}),
    hand-object interaction mask prediction (\textbf{HI}),
    hand motion forecasting (\textbf{HM}),
    interaction hotspot prediction (\textbf{IH}),
    person bounding box forecasting (\textbf{PB}),
    video progress prediction (\textbf{VP}),
    sample diversity (\textbf{SD}),
    and next active object prediction (\textbf{NAO}).\\
}
\label{tab:action-methods}
\vspace{-0.3cm}
\centering
\begin{tabular}{ccccccccccc}
\toprule
\textbf{Paper} & \textbf{Venue} & \textbf{Tasks} & \textbf{Cat.} & \textbf{Datasets} & \textbf{View} & \textbf{Ctx*} & \textbf{Anti*}                & \textbf{Inputs$^{\text{\textdaggerdbl}}$} & \textbf{Aux.\ tasks}\\
\midrule

Pei et al. \cite{pei2011parsing} & ICCV2011 & NAT & IB & Custom & Exo & 50m & 30s & DP/O & - \\
Lan et al. \cite{lan2014hierarchical} & ECCV2014 & NAT & IB & TVHI/UTI & Exo & 0.3s & 0.8s & R/DP & MC \\
LGCP \cite{mahmud2016poisson} & ICIP2016 & NAT & IB & MC & Exo & 4s & 4s & PA & - \\
Jain et al. \cite{jain2016recurrent} & ICRA2016 & NAT & IB & B4C & Exo & 0.8s & 4s & R & - \\
Zeng et al. \cite{zeng2017visual} & ICCV2017 & NAT & IB & TVHI/TH & Exo & 1s & Unk. & R/F & - \\
Mahmud et al. \cite{mahmud2017joint} & ICCV2017 & NAT & MM & MC/V & Exo & 15s & 26s & R/O & - \\
Shen et al. \cite{shen2018egocentric} & ECCV2018 & NAT & MM & GG/GG+ & Ego & 5s & 4s & R/G/H & GO \\
Rotondo et al. \cite{rotondo2019action} & VISIAPP2019 & NAT & IB & SE & Ego & 2s & 0s & R/PA/Pr & AR \\
Neumann et al. \cite{neumann2019future} & CVPRW2019 & NAT & IB & BDD/NB & Both & 5s & 10s & R/Pr & - \\
DR-VRM \cite{ke2021future} & WACV2021 & NAT & IB & 50S/BA/EK55 & Both & 60s & 61-180s & R/F & - \\
Roy et al. \cite{roy2022action} & WACV2022 & NAT & IB & 50S/BA/EK55 & Both & 2-15s & Unk. & R/F/O & - \\
Zhong et al. \cite{zhong2023learning} & CVPR2023 & NAT & PO & COIN & Exo & Unk. & Unk. & R/T & - \\
NAOGAT \cite{thakur2024leveraging} & WACV2024 & NAT & IB & EK100/Ego4D & Ego & 4s & Unk. & R/O & NAO \\
\arrayrulecolor{black!20}\hline\arrayrulecolor{black}
\arrayrulecolor{black!20}\hline\arrayrulecolor{black}
Wu et al. \cite{wu2017anticipating} & ICCV2017 & GFA & MM & Custom & Ego & Unk. & Unk. & R/Pr & - \\
Epstein et al. \cite{epstein2021learning} & CVPR2021 & GFA & PO & Oops/K600 & Exo & 6-10s & 6s & R & IT/TO \\
Suris et al. \cite{suris2021learning} & CVPR2021 & GFA & PO & FGym & Exo & 20s & 4s & R & TC \\
\arrayrulecolor{black!20}\hline\arrayrulecolor{black}
ATCRF \cite{koppula2015anticipating}  & TPAMI2015 & STA & IB & CAD-120 & Exo & 0-30s & 1-10s & R & - \\
Vondrick et al. \cite{vondrick2016anticipating} & CVPR2016 & STA & PO & TVHI & Both & 0s & 1s & R & FF \\
Multi-LSTM \cite{yeung2018every} & IJCV2018 & STA & IB & MTH & Exo & 1.5s & 0-2s & R & AS \\
Furnari et al. \cite{furnari2018leveraging} & ECCVW2018 & STA & PO & EK55 & Ego & 1s & 1s & R/PA & - \\
Miech et al. \cite{miech2019leveraging} & CVPRW2019 & STA & IB & AN/BA/EK55 & Both & Unk. & 1s & R & - \\
Ke et al. \cite{ke2019time} & CVPR2019 & STA & IB & 50S/EK55 & Both & 30s & 1-60s & R & - \\
Liu et al. \cite{liu2020forecasting} & ECCV2020 & STA & IB & EK55/EG+ & Ego & 2.7s & 0.5-1s & R/O & HM/IH \\
IAI \cite{zhang2020egocentric} & ACM MM2020 & STA & IB & EK55 & Ego & 11s & 1s & R/F/O & AS \\
Guan et al. \cite{guan2020generative} & CVPR2020 & STA & IB & EK55 & Ego & 2s & 5s & R/S/PA & - \\
Ego-Topo \cite{nagarajan2020ego} & CVPR2020 & STA & IB & EK55/EG+ & Ego & 1.5s-40m$^\dagger$ & 1.5s-40m$^\dagger$ & R & - \\
SF-GRU \cite{rasouli2020pedestrian} & BMVC2020 & STA & MM & PIE & Exo & 0.5s & 2s & R/DP/Po/Pr/PA & - \\
ImagineRNN \cite{wu2020learning} & TIP2020 & STA & PO & EK55/EG+ & Ego & 2.5s & 0.25-1.5s & R/F/O & - \\
Fernando et al. \cite{fernando2021anticipating} & CVPR2021 & STA & PO & BA/EK55 & Both & 2s & 1s & R/F/O & - \\
SF-RU-LSTM \cite{osman2021slowfast} & ICCVW2021 & STA & IB & EK55/EG+ & Ego & 3s & 0.5-2s & R/F/O & - \\
Zatsarynna et al. \cite{zatsarynna2021multi} & CVPRW2021 & STA & MM & EK55/EK100 & Ego & 5.25s & 1s & R/F/O & - \\
Ego-OMG \cite{dessalene2020egocentric,dessalene2021forecasting} & TPAMI2021 & STA & IB & EK55 & Ego & 60s & 1s & R & HI \\
LSTR \cite{xu2021long} & NeurIPS2021 & STA & IB & TH/TVS & Exo & 0-1024s & 2s & R/F & AR \\
AVT \cite{girdhar2021anticipative} & ICCV2021 & STA & PO & 50S/EK55/EK100/EG+ & Both & Unk. & 0.5-1s & R & FF/AS \\
DCR \cite{xu2022learning} & CVPR2022 & STA & PO & 50S/EK55/EK100/EG+ & Both & 5-10s & 0.25-1s & R/F/O & FF/FO \\
MeMViT \cite{wu2022memvit} & CVPR2022 & STA & IB & AVA/EK100 & Both & 70s & 1s & R & - \\
VLMAH \cite{manousaki2023vlmah} & ICCVW2023 & STA & MM & As101/Me/50S & Both & Unk. & 0.25-2s & R/G/H/PA & - \\
MAT \cite{wang2023memory} & ICCV2023 & STA & IB & HDD/TVS/TH/EK100 & Both & Unk. & 1s & R/F/O & AR \\
RAFTformer \cite{girase2023latency} & CVPR2023 & STA & PO & EK55/EK100/EG+ & Ego & Unk. & 1s & R & - \\
JOADAA \cite{guermal2024joadaa} & WACV2024 & STA & IB & TH/MTH/Ch & Exo & Unk. & 0.03-0.2s$^\dagger$ & R & - \\
InAViT \cite{roy2024interaction} & WACV2024 & STA & IB & EK100/EG+ & Ego & 2s & 0.5-2s & R/O/H & - \\
UADT \cite{guo2024uncertainty} & CVPR2024 & STA & IB & EK100/EG+  & Both & 2-3s & 0.5-1s & R/F/O & AR \\
S-GEAR \cite{diko2024semantically} & ECCV2024 & STA & PO & 50S/EK55/EK100/EG+  & Both & 10-15s & 0.5-1s & R/F/O & - \\

\bottomrule
\end{tabular}
\end{table*}

\begin{table*}[!h]
\scriptsize
\renewcommand{\arraystretch}{1.3}
\caption{
    Continuation of Table \ref{tab:action-methods}.
    Refer to Table \ref{tab:action-methods} for abbreviation meanings.
}
\label{tab:action-methods-2}
\centering
\begin{threeparttable}
\begin{tabular}{cccccccccccc}
\toprule
\textbf{Paper} & \textbf{Venue} & \textbf{Tasks} & \textbf{Cat.} & \textbf{Datasets} & \textbf{View} & \textbf{Ctx*} & \textbf{Anti*}                & \textbf{Inputs$^{\text{\textdaggerdbl}}$} & \textbf{Aux.\ tasks}\\
\midrule

Chakraborty et al. \cite{chakraborty2015context} & ACCV2015 & FIA & IB & V & Exo & Unk. & Unk. & R/O/DP & - \\
Rhinehart et al. \cite{rhinehart2017first} & ICCV2017 & FIA & IB & Custom & Ego & Unk. & Unk. & R/O/PA/DP/SM & - \\
RED \cite{gao2017red} & BMVC2017 & FIA & IB & TVS/TH/TVHI & Exo & 4s & 0.25-2s & R/O & - \\
RU-LSTM \cite{furnari2019would,furnari2020rolling} & ICCV2019 & FIA & IB & AN/EK55/EG+ & Both & 1.75-3.5s & 0.25-2s & R/F/O & AR \\
DR$^2$N \cite{sun2019relational} & CVPR2019 & FIA & IB & AVA & Exo & 0.3s & 5s & R/DP & PB \\
AGG \cite{piergiovanni2020adversarial} & ECCV2020 & FIA & IB & 50S/MTH/Ch & Both & 1-2m$^\dagger$ & 1-45s & R & - \\
Liu et al. \cite{liu2020spatiotemporal} & ICRA2020 & FIA & IB & JAAD/STIP & Exo & 4s & 3s & R/O/DP & - \\
RESTEP \cite{li2021restep} & IEEE MM2021 & FIA & IB & AVA/V/J-HMDB & Exo & 1s & 5s & R/DP & - \\
SRL \cite{qi2021self} & TPAMI2021 & FIA & PO & 50S/BA/EK55/EG+ & Both & 1.5s & 0.25-2s & R/F/O & TC \\
HRO \cite{liu2022hybrid} & CVPR2022 & FIA & PO & EK55/EG+ & Ego & 1.5-2s & 0.25-2s & R/F/O & - \\
TeSTra \cite{zhao2022real} & ECCV2022 & FIA & IB & TH/EK100 & Both & 0-1024s & 0.25-2s & R/F & AR \\
AFFT \cite{zhong2023anticipative} & WACV2023 & FIA & MM & EK100/EG+ & Ego & 4-20s & 0.5s & R/F/O/A & - \\
\arrayrulecolor{black!20}\hline\arrayrulecolor{black}
ST-AOG \cite{qi2017predicting} & ICCV2017 & LTA & IB & CAD-120 & Exo & Unk. & 3s & R/O & - \\
Schydlo et al. \cite{schydlo2018anticipation} & ICRA2018 & LTA & IB & CAD-120/ACT & Exo & 1s & 1.5s & G/Po & - \\
Gammulle et al. \cite{gammulle2019forecasting} & BMVC2019 & LTA & IB & 50S/BA & Exo & 1-2m$^\dagger$ & 0.5-3m$^\dagger$ & R/PA & - \\
Sener et al. \cite{sener2019zero,sener2022transferring} & ICCV2019 & LTA & PO & R1M/TA/YC2 & Exo & 48s & 24s & PA/R & AE/Ca \\
Ng et al. \cite{ng2020forecasting} & TIP2020 & LTA & IB & 50S/BA/Ch/MC & Exo & 10m & 10m & R/F & AR/FF/BA \\
TempAgg \cite{sener2020temporal,sener2021technical} & ECCV2020 & LTA/STA & IB & 50S/BA/EK55 & Both & 1-2m$^\dagger$ & 0.5-3m$^\dagger$ & R/F/O/PA & - \\
MM-Transformer \cite{roy2021action} & TIP2021 & LTA & IB & 50S/BA/EK55 & Both & 1-5s & 1s-3m & R/DP/O & - \\
GePSAN \cite{abdelsalam2023gepsan} & ICCV2023 & LTA & IB & YC2 & Exo & Unk. & Unk. & R & - \\
I-CVAE \cite{mascaro2023intention} & WACV2023 & LTA & IB & Ego4D & Ego & 10s$^\dagger$ & 50s$^\dagger$ & R & AS/GP \\
HierVL \cite{ashutosh2023hiervl} & CVPR2023 & LTA & PO & Ego4D & Ego & Unk. & 50s$^\dagger$ & R/T & - \\
Zhang et al. \cite{zhang2024object} & WACV2024 & LTA/NTA & IB & 50S/EG+/Ego4D & Both & 1-2m$^\dagger$ & 0.5-3m$^\dagger$ & R & - \\
AntGPT \cite{zhao2024antgpt} & ICLR2024 & LTA & PO & EK55/EG+/Ego4D & Ego & 1.5s-40m$^\dagger$ & 1.5s-40m$^\dagger$ & R/PA & - \\
PlausiVL \cite{mittal2024can} & CVPR2024 & LTA & PO & EK100/Ego4D & Ego & Unk. & 0.5-3m$^\dagger$ & R/PA & - \\

\arrayrulecolor{black!20}\hline\arrayrulecolor{black}

Abu Ferha et al. \cite{abu2018will} & CVPR2018 & AS & IB & 50S/BA & Exo & 1-2m$^\dagger$ & 0.5-3m$^\dagger$ & PA & - \\
UAAA \cite{abu2019uncertainty} & ICCVW2019 & AS & IB & 50S/BA & Exo & 1-2m$^\dagger$ & 0.5-3m$^\dagger$ & PA & - \\
APP-VAE \cite{mehrasa2019variational} & CVPR2019 & AS/NAT & IB & MTH/BA & Exo & Unk. & Unk. & PA & - \\
AAP \cite{zhao2020diverse} & ECCV2020 & AS & PO & 50S/BA/EK55 & Both & 1-2m$^\dagger$ & 0.5-3m$^\dagger$ & PA & SD \\
MAVAP \cite{loh2022long} & CVPRW2022 & AS & PO & 50S/BA & Exo & 1-2m$^\dagger$ & 0.5-3m$^\dagger$ & R/PA & - \\
FUTR \cite{gong2022future} & CVPR2022 & AS & IB & 50S/BA & Exo & 1-2m$^\dagger$ & 0.5-3m$^\dagger$ & R/PA & AR \\
ANTICIPATOR \cite{nawhal2022rethinking} & ECCV2022 & AS & IB & EK100/TH & Both & Unk. & 0.25-2s & R/F/O & AR \\
DIFFANT \cite{zhong2023diffant} & Arxiv2023 & AS & IB & 50S/BA/EK55/EG+ & Both & 1-2m$^\dagger$ & 0.5-3m$^\dagger$ & R & - \\
\arrayrulecolor{black!20}\hline\arrayrulecolor{black}
DDN \cite{chang2020procedure} & ECCV2020 & PP & IB & CT & Exo & 22s & 2.5m & R/F/A & FF \\
Bi et al. \cite{bi2021procedure} & ICCV2021 & PP & IB & CT & Exo & 22s & 2.5m & R/F/A & - \\
PlaTe \cite{sun2022plate} & RAL2022 & PP & IB & CT & Exo & 22s & 2.5m & R/A & FF \\
P${}^3$IV \cite{zhao2022p3iv} & CVPR2022 & PP & PO & CT/NIV/COIN & Exo & 22s & 2.5m & R & FF/Ca \\
PDPP \cite{wang2023pdpp} & CVPR2023 & PP & IB & CT/NIV/COIN & Exo & 22s & 2.5m & R & - \\
\bottomrule
\end{tabular}

\begin{tablenotes}
\item * Most papers do not supply this information, so we had to come up with rough estimates. Where one paper considers several tasks with different contexts or anticipation time, we report the time range. We also denote unknown (\textbf{Unk.}) for papers that don't provide enough information for estimation.
\item $\dagger$ Some papers denote the context and anticipation duration by the percentage of videos, number of actions and number of video frames. We convert them to time duration to our best estimate for a unified format.
\item \textdaggerdbl For methods that learn from pretrained features (e.g., using vision features from a pretrained 3D CNN), we list only the input modality used to generate the features (i.e., RGB video or flow, in the case of a pretrained 3D CNN). For each model, we show all the modalities it can take in for a thorough review. Some models can just use a subset of listed modalities as input. 
\end{tablenotes}
\end{threeparttable}
\end{table*}

\section{Tasks and methods}
\label{sec:action-tasks}

Existing action anticipation research explores a variety of problem settings, which are defined by differences in the observable inputs, the anticipation time span, and the granularity of the anticipated actions. These variations shape the complexity and scope of the task. We compile the variations from existing research and formally define seven distinct anticipative tasks in Figure \ref{fig:action-tasks}. In the following sections, we dicsuss the methods of each task in detail. 

Further, we organize existing works of each task into three sub-categories depending on whether we perceive the primary contribution lies in:\footnote{If the contributions of a paper span several sub-categories, we assign it to the category that aligns with the most salient novelty.}

\begin{itemize}
    \item \textbf{Inductive Biases} -- innovations in model design or novel approaches that incorporate domain-specific knowledge to address challenges within a learning algorithm.
    \item \textbf{Multi-modal Models} -- involvement of a new input modality that has not been studied before, or proposal of novel multi-modal integration methods to enhance learning and inference capabilities.
    \item \textbf{Pretraining and Objectives} -- strategies and criteria used to guide the learning process, including the formulation of loss functions, selection of pretraining datasets and development of novel pretraining methods.
\end{itemize}

To incorporate more \textit{inductive bias}, the prior work shows the trend of integrating stronger backbones in sequence modeling, establishing more hierarchical visual representations, taking in longer observed videos within a fixed embedding size, and learning more fine-grained dependence across past and future actions. In terms of \textit{multi-modal models}, an important thread is leveraging the action labels of the observed video segment as textual modality to complement for visual embeddings. Other modalities including audio and gaze have also been used as extra evidence in a few papers. We also observe that this sub-category is missing in action sequence modeling including forecasting long-term actions, action segments and procedure planning. The reason may be the weak connections of past non-visual features and future actions. For \textit{pretraining and training objective}, prior work commonly adds auxiliary losses that have aligned optimization objects at different granularities, such as predicting the final goal, classifying the observed actions, and localizing next active objects. Pretraining for general video representation on large-scale datasets also shows gains in forecasting tasks. Methods in this sub-category accounts for a higher percentage in goal and final action anticipation, probably because novel loss and more pretraining data bring more prior knowledge that is critical in long-term dependence modeling.

The comprehensive summary is listed in Table \ref{tab:action-methods} and Table \ref{tab:action-methods-2}. The following subsections will also briefly mention some datasets in the context of explaining existing work. However, we leave a detailed discussion and comparison of existing datasets to Section \ref{sec:datasets}.

\subsection{Next Action and Time Anticipation}
\label{ssec:act-nat}

In next action and time prediction, a model takes as input the first $t$ frames of a $T$-frame video $\mathcal V = (o_1, o_2, \ldots, o_T)$, and must predict a subset of:
\begin{itemize}
    \item The label of the next action $a_i$ that starts \textit{after} frame $o_t$.
    \item The start time $t_i$ of action $a_i$, possibly parameterized as an inter-arrival time (i.e., the time $\Delta t = t_i - t_{i-1}$ between the start of the last action $a_{i-1}$ and $a_i$) or just an offset from the time $t$ of the current frame.
    Depending on the structure of the dataset, the start time of the next action might be accompanied by an estimate of the time remaining until the end of the most recent action $a_{i-1}$ (i.e., $t'_{i-1} - t$).
    \item The duration of the next action $a_i$.
\end{itemize}
Some methods predict all of these quantities, and some predict just a subset (e.g., predicting only the label of the next action).
The difficulty of this task can vary greatly across datasets with different distributions of inter-arrival times.

\begin{figure}[!t]
    \centering
    \includegraphics[width=\linewidth]{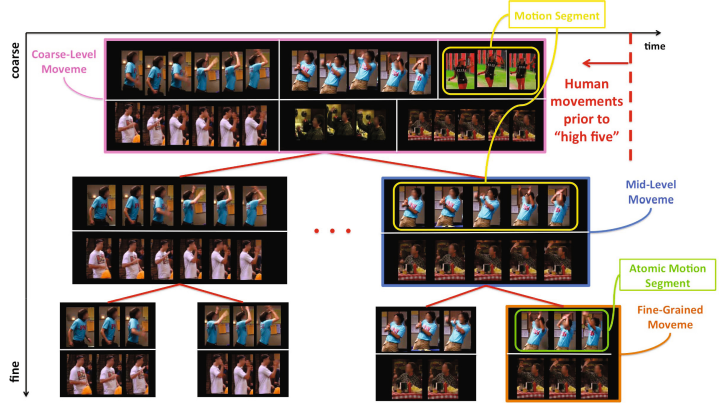}
    \caption{
        An illustration of how the movements that precede a high-level action (a high five in this case) can be decomposed into a hierarchy of coarse, mid-level and fine-grained movemes. In this figure, the $x$-axis represents time, and the $y$-axis represents moveme granularity.
        The frame sequences in each colored box serve as examples of the corresponding moveme.
        \textit{Reproduced from Figure~2 of Lan et al. \cite{lan2014hierarchical}.}
    }
    \label{fig:highlight-lan-moveme}
\end{figure}

\subsubsection{Metrics}

The label prediction component of this task is typically evaluated with ordinary classification metrics like accuracy, precision, recall, AUC, F1 score, and so on.
Time predictions are usually evaluated with Root Mean Squared Error (RMSE) \cite{mahmud2016poisson}, mean absolute error \cite{neumann2019future}, or log likelihood (for probabilistic models).

\subsubsection{Approaches Overview}

Approaches to next action and time anticipation frequently focus only on predicting the label of the next action, or only on the start time or inter-arrival time of the next action.
Early work on predicting labels focuses on building sophisticated prior knowledge into the model architecture, for instance by using and-or graphs to model past action sequences \cite{pei2011parsing} or building classifiers for high-level actions on top of classifiers for low-level motions \cite{lan2014hierarchical}. In contrast, later methods tend to incorporate fewer task-specific inductive biases, such as modeling the temporal relation of the past sequences \cite{jain2016recurrent}, learning the state transition during actions \cite{zeng2017visual}, predicting the final goal \cite{roy2022action} or next active objects \cite{thakur2024leveraging}. Among work that predicts action start times, a common theme is the use of more sophisticated probabilistic models to better model the variation in possible start times: examples include log-Gaussian cox processes \cite{mahmud2016poisson}, VAEs \cite{ke2021future}, and discretized models \cite{neumann2019future}. 
For both action and time prediction, fusing multi-modal inputs can substantially improve prediction accuracy \cite{mahmud2017joint,shen2018egocentric,rotondo2019action} -- as we will see, this is a common finding across all tasks in this paper. Novel pretraining methods and training objectives have not been widely explored for next action and time anticipation. Estimating the progress of an assembly task \cite{han2017human}, pretraining the model with action narrations and enforcing the alignment with the goal \cite{roy2022action} in a contrastive learning scheme \cite{zhong2023learning} are three directions that have been studied in prior work.

\subsubsection{Inductive Biases}

The early work only predicts the next action label \cite{pei2011parsing, lan2014hierarchical}. Pei et al. \cite{pei2011parsing} propose the first exocentric action anticipation model by using a probabilistic and-or graph. They parse a video into a sequence of truth values for unary and binary predicates describing detected people and objects, and then learn weights for a probabilistic and-or graph on top of those predicates. The hierarchical nature of this graph implicitly uses both top-down and bottom-up cues to predict actions. 
Lan et al. \cite{lan2014hierarchical} frame prediction of the next action label as the task of predicting a hierarchy of ``movemes'' \cite{bregler1997learning} (see Figure \ref{fig:highlight-lan-moveme}), which are abstract categories of motion that might occur in a video. Next action prediction is performed by doing inference on a graphical model that assigns movemes to each frame.
More follow-up work focuses on capturing the sequential relation and dynamics of actions. Jain et al. \cite{jain2016recurrent} use a straightforward LSTM \cite{hochreiter1997long} to capture the dependence of actions over time and forecast the driving maneuvers from the exocentric view. Zeng et al. \cite{zeng2017visual} formulate next action anticipation as an imitation learning problem. They propose to learn a policy to simulate the state transition happening in the real world and forecast the behaviors in the short future.
Another strategy to make an accurate anticipation is involving auxiliary tasks. Thakur et al. \cite{thakur2024leveraging} leverage the object features of the past actions and predict the next active objects, which serves an additional evidence for next action anticipation in egocentric view.

Apart from the action label, predicting the moment of the next action is also an important task in real-world problems \cite{mahmud2016poisson,neumann2019future,ke2021future}. The earliest method for predicting action inter-arrival times (but not labels) comes from Mahmud et al. \cite{mahmud2016poisson}, who propose using a Log-Gaussian Cox Process (LGCP) to train a separate model of the inter-arrival time distribution for each action class, conditioned on the start time and label of the last observed action. This idea is further extended to forecasting the action and its occurrence time in a single model \cite{neumann2019future}. Similarly, Ket et al. \cite{ke2021future} also work on this \textit{if-and-when} problem and propose the deterministic residual guided variational regression module. For time prediction, they use a regression head consisting of a VAE-like stochastic branch and a normal deterministic branch.

\subsubsection{Multi-modal Models}

Modalities used for next action and time anticipation may include RGB frames, object features, gaze, biometric information, etc. \cite{mahmud2017joint,shen2018egocentric,rotondo2019action}. Mahmud et al. \cite{mahmud2017joint} propose a simple multi-modal model for joint prediction of the next action label and inter-arrival time. Their model has one MLP stream that conditions on detected objects, a second MLP stream that conditions on extracted visual features, and a recurrent LSTM stream that conditions on visual features from a longer sequence. The experiments on two exocentric datasets suggest the object branch is the most important to the accuracy.

In terms of egocentric perspective, Shen et al. \cite{shen2018egocentric} propose a two-stream video backbone that can process egocentric sensor data both synchronously and asynchronously based on the characteristics of egocentric gaze. The synchronous stream processes a sequence of hand masks and tracks gaze locations with a normal RNN at a fixed frame rate. The asynchronous stream uses gaze tracking and object detection to identify frames where the camera-wearer's gaze drifts into or out of the area of a detected object. It then processes object and gaze features from \textit{only} the frames where those events happen, under the assumption that those frames will be the most useful for action anticipation. The model computes a global video representation by using gaze-based attention to fuse the two streams, and then does next action prediction on top of this representation.
Rotondo et al. \cite{rotondo2019action} further investigate the additional value of heart rate and acceleration modalities for next action prediction, which are more easily to acquire with wearable devices from the egocentric view.

\subsubsection{Pretraining and Objectives}

Some prior work proposes new losses to penalize errors \cite{han2017human} or encourage alignment \cite{roy2022action,zhong2023learning}. Roy et al. \cite{roy2022action} argue that the next action should be a step towards the final goal of this activity. Hence, they utilize a stacked RNN to predict the representation of the final goal, and then propose two losses, action-goal consistency loss and goal completeness loss, to facilitate the anticipation performance. 
Zhong et al. \cite{zhong2023learning} propose to pre-train the backbone with a contrastive learning strategy. They encode the video frames by a learnable visual encoder and encode the action narrations by a fixed text encoder. The visual encoder is pre-trained to align with the text embeddings in the feature space. They also model the temporal dependence of steps in an activity using a diffusion process. To forecast next action, they sample from the gaussian noise and conduct the reverse diffusion process to get the generated visual embedding which is then decoded to specific actions.

\subsection{Goal and Final Action Anticipation}

Final action anticipation is the task of predicting the last action that will be taken by a subject in a video. For example, if a model can tell that the subject is baking a cake, then it might predict ``take cake out of oven'' as the final action. Formally, consider a video of $T$ frames, $\mathcal V = (o_1, o_2, \ldots, o_T)$, where the subject of the video is taking action $a_T$ in frame $o_T$. Given a subsequence of $t < T$ (i.e., frames $o_1, o_2, \ldots, o_t$), final action prediction is the task of predicting the last action $a_T$.
Goal prediction is a similar task, except the label to be predicted represents the goal that the subject achieves at the end of the video (e.g., ``bake cake'').

\subsubsection{Metrics}

Since goal and final action prediction are both classification tasks, they are usually evaluated with generic classification metrics like accuracy or top-$k$ recall. Suris et al. \cite{suris2021learning} propose two additional metrics specifically for datasets with hierarchical label taxonomies. 
Intuitively, these metrics give a model ``partial credit'' for predicting a label that lies within the same coarse-grained category as the true fine-grained label, even if the model does not predict that true fine-grained label.

\subsubsection{Approaches Overview}

Goal and final action anticipation has a simple input/output specification: models only need to take in a snippet of video and produce a single label, and so in principle any video classification model could be applied to this task. This is still an understudied area with very little prior work. Some existing work on this task has focused on multi-modal learning and special loss functions or auxiliary tasks that are appropriate for goal and final action anticipation \cite{epstein2021learning,suris2021learning}, although Wu et al. \cite{wu2017anticipating} also propose an asynchronous forecasting model that could be useful for any video prediction task on a power-constrained device. Roy et al. \cite{roy2022predicting} also propose a short-term action anticipation model that attempts to infer latent goals as an aid to action anticipation (covered in Section \ref{ssec:act-sha}).

\subsubsection{Multi-modal Models}

Wu et al. \cite{wu2017anticipating} experiment with using a wrist-mounted camera and accelerometer to determine the goal of someone going about daily activities in a house. Experiments on their custom Daily Intentions Dataset, which includes goals like ``drink water'', ``go outside'' and ``charge phone'', show that combining video and accelerometer data can improve accuracy over using one modality alone. Moreover, they use accelerometer data to predict whether the the computationally expensive visual backbone should be invoked anew at each time step, as opposed to using cached visual representations for past time steps. By invoking the visual backbone only for frames that are anticipated to be important (according to accelerometer data), they reduce use of the visual backbone by around 70\% with only slight loss of accuracy, which may be useful for power-constrained applications of goal prediction. Note that this technique is most likely to be useful in an egocentric setting, since it requires a subject-mounted accelerometer. Its use of accelerometer data is also similar to the way that gaze data is used by Shen et al. \cite{shen2018egocentric} in their asynchronous next action prediction model (Sectoin \ref{ssec:act-nat}).

\subsubsection{Pretraining and Objectives}

\begin{figure}[!t]
    \centering
    \includegraphics[width=\linewidth]{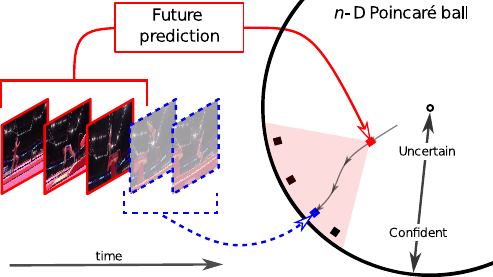}
    \caption{
        Hyperbolic geometry can capture uncertainty over future video sequences \cite{suris2021learning}.
        At left is a video sequence split into observed and unobserved segments.
        At right is a sketch of a hyperbolic space, represented by a circle.
        In this example, the model can infer that the future frame embeddings will be one of the black squares (\textcolor{black}{\rule{0.5em}{0.5em}}), but is uncertain about which one.
        Averaging these possible embeddings produces a point $\hat z$ near the origin of the Poincaré ball.
        After observing more frames, the model will gain enough confidence to update its representation to the more specific value $z$ near the edge of the ball.
        \textit{Adapted from Figure 2 of Suris et al. \cite{suris2021learning}.}
    }
    \label{fig:highlight-suris-hyperbolic}
\end{figure}

Suris et al. \cite{suris2021learning} combine final action prediction with a contrastive pretraining scheme that encourages the model to internally represent abstract hierarchies over possible futures (see Figure \ref{fig:highlight-suris-hyperbolic}). Their contrastive pretraining scheme is reminiscent of Dense Predictive Coding (DPC) \cite{han2019video}: a loss term encourages frames separated by a certain time offset within the same video (e.g., 32 frames) to have closer representations than frame pairs with different offsets or from different videos. Suris et al. replace the Euclidean norm in the original DPC loss with a hyperbolic norm, and also employ network layers that use hyperbolic rather than Euclidean arithmetic. Hyperbolic spaces make it possible to represent tree-like structures in low dimensions, and provide an inductive bias towards capturing semantic hierarchy in data. Epstein et al. \cite{epstein2021learning} use special mistake annotations from the ``Oops!'' dataset to obtain a pretrained representation for goal recognition \cite{epstein2020oops}. They pretrain a neural network on two objectives. The first predicts ``intentionality'' -- whether a given frame occurs before or after the start of the mistake. The second distinguishes normal videos from videos that have been deliberately scrambled (frames appear out of order).

\subsection{Short-Term Action Anticipation}\label{ssec:act-sha}

Short-term ($\tau$-second ahead) action anticipation requires a model to predict what action will occur at a fixed offset from the current time. Specifically, the model observes up to the $t$-th frame $o_t$, and must predict the action that occurs $\tau$ seconds after frame $t$. This differs from next action anticipation in that the offset $\tau$ is constant, even if the ``next action'' starts more than $\tau$ seconds into the future, or finishes fewer than $\tau$ seconds into the future.

\subsubsection{Metrics}

Short-term action anticipation can be evaluated with the same metrics as other classification tasks. Popular choices include top-$k$ accuracy, which is used for evaluations on Epic-Kitchens 55 \cite{damen2018scaling}, and top-5 recall averaged across classes, which is used for Epic-Kitchens 100 \cite{damen2020rescaling}.

\subsubsection{Approaches Overview}

Short-term action anticipation models need to take in a snippet of video (and possibly other modalities) and produce a single label describing the action that will happen $\tau$ seconds after the last observed frame. This is the most common setting for future action anticipation approaches and thus covers plenty of prior work. Existing approaches largely differ along two axes: the specific architecture of the model used to process the input video, and any auxiliary tasks or features used alongside the plain action anticipation problem.

Models used in past work include simple task-agnostic architectures like RNNs \cite{yeung2018every,osman2021slowfast,rasouli2020pedestrian} and transformers \cite{girdhar2021anticipative,wu2022memvit,gu2021transaction,girase2023latency,wang2021oadtr}. Other methods provide a stronger task-specific inductive bias by enforcing that models should use a particular intermediate representation of videos. For example, \cite{miech2019leveraging,zhang2020egocentric} first recognize past actions or objects and then predict a future action accordingly. \cite{zatsarynna2023action,roy2022predicting} forecast the final goal based on the observed video segments and then fill the actions in between. \cite{dessalene2021forecasting,tai2022unified} represent observed video segments as a kind of graph that captures hand-object interactions. Another strategy is first predicting visual features for future frames, and then classifying the action corresponding to those future features \cite{vondrick2016anticipating,fernando2021anticipating,wu2020learning}, which is a theme that we will see in other sections as well.

Auxiliary tasks used in past work include prediction of object affordance \cite{koppula2015anticipating}, modeling hand-object interactions \cite{liu2020forecasting,thakur2023enhancing}, temporal re-ordering of video frames \cite{xu2022learning}, past action recognition \cite{wang2021oadtr} and online action detection \cite{guermal2024joadaa}. Video feature forecasting losses have also been used in models that explicitly use predicted future features as an intermediate representation \cite{vondrick2016anticipating,fernando2021anticipating}, as well as models that merely use feature forecasting as an auxiliary objective \cite{girdhar2021anticipative,xu2022learning}. Finally, Furnari et al. \cite{furnari2018leveraging} have proposed differentiable classification losses that better approximate a top-$k$ recall objective, which they argue is a good fit for situations where future labels are intrinsically ambiguous.

\subsubsection{Inductive Biases}

The architecture for short-term action anticipation evolves from straightforward RNN \cite{yeung2018every,osman2021slowfast} to transformer-based model \cite{xu2021long,wu2022memvit,wang2023memory,guo2024uncertainty}. Yeung et al. \cite{yeung2018every} propose a modified RNN architecture for action segmentation and anticipation. The main innovation is extending the basic LSTM cell so that it can attend over a window of past inputs, rather than just conditioning on the current input and previous hidden state. Recently, more approaches leverage transformers to learn from a longer history actions. Gu et al. \cite{gu2021transaction} design a hierarchical model consisting of a few temporal self-attention modules and one cross-modal self-attention module. Xu et al. \cite{xu2021long} innovatively propose Long Short-term TRansformer (LSTR) to encode long-term and short-term history in separate. The LSTR encoder can embed the long-term history into visual tokens with a fixed size, which allows the model to learn from observed video segments of any length (up to 1024 seconds in the experiments). The short-term history is fed into LSTR decoder directly along with the long-term visual tokens. Similarly, MeMViT \cite{wu2022memvit} is another important approach that involves longer history by adapting the ViT architecture \cite{dosovitskiy2020image} in two ways. First, representations of older frames are ``compressed'' through a learned dimensionality reduction operation. Second, MeMViT stops gradients from propagating into the original video frames representation once they have been compressed, which means that it does not have to keep the uncompressed activations in memory. This approximation allows the model to process substantially longer contexts within a given memory budget. Wang et al. \cite{wang2023memory} propose a novel network to first fuse the long-term and short-term memory in the past and then iteratively generates future action features conditioned on the fused memory.

Another important innovation in terms of the model architecture is modeling the relation of past actions into a graph \cite{dessalene2021forecasting,tai2022unified}. Dessalene et al. \cite{dessalene2021forecasting} propose Ego-Topo to parse videos into a special kind of graph that is useful for egocentric bag-of-actions anticipation. Two frames from a training video are considered to belong to the same node if they occur close together in time, or if they meet a certain visual similarity threshold; otherwise they belong to different nodes. After classifying frames of training videos in this manner, Ego-Topo uses the frames to train a siamese network that detects frame similarity. Likewise, Tai et al. \cite{tai2022unified} also use graph to model observed actions. They further propose a novel graph learning block to update the graph state by message passing.

Leveraging hand-object interactions is another theme for short-term action anticipation \cite{koppula2015anticipating,liu2020forecasting,thakur2023enhancing,roy2024interaction}. ATCRF \cite{koppula2015anticipating} is an early work that forecasts the future action based on object affordance analysis and hand trajectory prediction. Liu et al. \cite{liu2020forecasting} show that predicting the interactions between hands and objects can be complementary to egocentric action anticipation. They use the early, middle and final layers in the model to capture the future motion of head, the future hand-object interaction hotspots and the future actions respectively. Thakur et al. \cite{thakur2023enhancing} propose to use the bounding boxes of objects in observed frames to obtain object-centric embeddings which are fed into the model together with the RGB frames. In the meanwhile, the model learns to forecast the future action and the bounding box of the next active object. Roy et al. \cite{roy2024interaction} use Faster RCNN \cite{ren2015faster} to get the location of objects and hands and extract interaction-centric embeddings. Then they contextualize the interaction-centric embedding with RGB frames and forecast the future action.

In addition to hand and object features, there are also many methods relying on other auxiliary tasks or involving features of different levels to boost the action anticipation performance by using mutli-stream networks \cite{zhang2020egocentric,guermal2024joadaa}. Miech et al. \cite{miech2019leveraging} use two additional branches to involve semantically meaningful high-level attributes based on inferred past actions and object detection probabilities. Zhang et al. \cite{zhang2020egocentric} explicitly learn the transition of verbs and nouns from the current action to the future action. Then they merge this stream to the main encoder-decoder branch to make the prediction. Roy et al. \cite{roy2022predicting} argue that it's easier to predict future actions by first predicting a future goal and then predicting an action conditioned on that goal. Short-term action anticipation can also be conducted as a intermediate step for online action detection \cite{xu2019temporal,qu2020lap}. Guermal et al. \cite{guermal2024joadaa} jointly model the online action detection and action anticipation problems in a single model (i.e., JOADAA). It uses a three-branch architecture to explicitly embed the past observation, present frames and future forecasting.

Different than prior work that uses a fixed time interval in training, Ke et al. \cite{ke2019time} encode the time interval by a learnable layer and then propose time-conditioned skip connections that inject time into the action anticipation network. The model is able to forecast future actions conditioned on any time interval.

\subsubsection{Multi-modal Models}

Merging multiple modalities is another strategy for accurate future action anticipation \cite{rasouli2020pedestrian,zatsarynna2021multi,manousaki2023vlmah,ghosh2023text}. Rasouli et al. \cite{rasouli2020pedestrian} leverage visual cues of different granularities for future action anticipation of pedestrians, including the foreground of pedestrians, local contexts (i.e., background), pedestrian poses, location of pedestrians in each frame and motion of the camera. They propose SF-GRU which is composed of five GRU blocks to integrate these features step by step. Manousaki et al. \cite{manousaki2023vlmah} innovatively leverage texts as an additional modality to better understand the semantics of past actions. Specifically, they propose a two-stream architecture with one branch taking in observed RGB frames as visual input and the other branch taking in past action labels as textual input. The two branches forecast the future action separately. Then they merge the outputs of the two branches as the final results. Ghosh et al. \cite{ghosh2023text} also investigate visual-language learning for egocentric action anticipation and they are the first that leverage the powerful capability of large language models in short-term action anticipation.

\begin{figure}[!t]
    \centering
    \includegraphics[width=\linewidth]{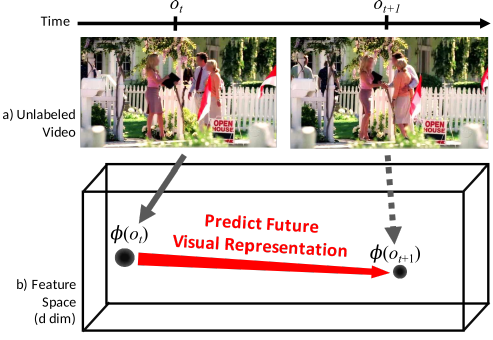}
    \caption{
        Vondrick et al. \cite{vondrick2016anticipating} propose future feature forecasting as an intermediate task in short-term action anticipation. Given a representation of the most recently observed frame $o_t$, their model predicts the representation $\phi(o_{t+1})$ of the target frame, then predicts an action from that representation. 
        The advantage of this approach is that the frame forecasting model can be trained on a large amount of unlabeled data. 
        \textit{Adapted from Figure~1 of Vondrick et al. \cite{vondrick2016anticipating}.}
    }
    \label{fig:highlight-vondrick-forecasting}
\end{figure}

\subsubsection{Pretraining and Objectives}

Vondrick et al. \cite{vondrick2016anticipating} reduce short-term action anticipation to a video forecasting task. Their model first predicts what features AlexNet would assign to the frame at some fixed point in the future (see Figure \ref{fig:highlight-vondrick-forecasting}), and then predicts a future action from that predicted frame representation. The breakthrough of this indirect approach is that only the action classifier needs to be trained on labeled data. Thus the feature anticipation module can be pretrained on a large unlabeled dataset, like a set of YouTube videos. This important innovation has become a regular learning strategy for many following tasks \cite{gupta2022act,zhong2018unsupervised,wu2020learning,girdhar2021anticipative}. Wu et al. \cite{wu2020learning} propose ImagineRNN that combines the idea of anticipation via feature forecasting \cite{vondrick2016anticipating} with the RU-LSTM anticipation architecture \cite{furnari2018leveraging,furnari2020rolling} for egocentric action anticipation. ImagineRNN makes the unusual choice of predicting \textit{differences} between features for adjacent future frames with a contrastive loss. The authors argue that predicting differences between frame features encourages the model to capture changes in a scene rather than static appearance, and that difference prediction acts somewhat like a skip connection during training. Along similar lines, the Anticipative Video Transformer (AVT) \cite{girdhar2021anticipative} uses a transformer to do action anticipation via future frame forecasting. Specifically, in their base architecture, a pretrained ViT is used to embed a subset of past frames, and then a causally masked transformer is used to predict future frame features, which can be classified with a linear head. In addition to an action anticipation loss, the model uses two auxiliary losses: a future frame forecasting loss, and an action classification loss (for actions before the final observed frame).

As noted in the discussion on metrics, Furnari et al. \cite{furnari2018leveraging} have argued that top-1 metrics are not a good fit for action anticipation because anticipation is inherently ambiguous. They argue that using cross-entropy loss with the true future action as a label can exacerbate this problem by encouraging models to predict only that one label, rather than all plausible labels. Thus, in addition to proposing the use of top-$k$ metrics (with $k>1$) for evaluation, they also propose several differentiable training objectives that approximate the top-$k$ recall. Similarly, Camporese et al. \cite{camporese2021knowledge} also take into account the ambiguity of possible future actions. They propose data-driven label smoothing methods based on linguistic similarity of action labels and temporal co-occurrence statistics.

In terms of multi-task learning, an important work is Dynamic Context Removal (DCR) from Xu et al. \cite{xu2022learning}. They involve two auxiliary tasks to improve action anticipation performance. The first task is recovering frame ordering: given a video with scrambled frame order, the model is trained to predict how close together each pair of frames was in the original video. The second task is feature forecasting: the model is jointly trained to predict vision features of future frames. Unlike past applications of this idea, such as Vondrick et al. Vondrick et al. \cite{vondrick2016anticipating}, DCR initially makes the feature forecasting task easier by including features from some future frames as input, and then gradually drops out the future frames over the course of training. Zatsarynna et al. \cite{zatsarynna2023action} add one more branch to forecast the final goal of the observed actions. They also use a consistency loss between the final goal and the predicted next action.

\subsection{Fixed-Interval Action Anticipation}
\label{ssec:act-fixed-iv}

Fixed-interval action anticipation is a recursive counterpart to short-term action anticipation. After observing the first $t$ frames $o_1, o_2, \ldots, o_t$, future actions are predicted at regular intervals of $\tau$ seconds. This means predicting the action that occurs $\tau$s after frame $t$, $2\tau$s seconds later, $3\tau$ seconds later, and so on. Short-term action anticipation approaches mentioned in Section \ref{ssec:act-sha} can be applied to fixed-interval anticipation problem simply by forecasting the action after $\tau$ seconds recursively. In this section, we focus on the models specifically designed for fixed-interval action anticipation, which usually differ from short-term anticipation in that neural-network-based methods must have some kinds of sequence decoders in order to generate predictions for all the future offsets. Furthermore, models are sometimes evaluated with sequence prediction metrics instead of normal classification metrics.

\subsubsection{Metrics}

Fixed-interval action anticipation methods are most commonly evaluated at a fixed temporal offset (e.g., 1s in the future) to enable comparison with short-term anticipation methods. These comparisons use the same metrics described in Section \ref{ssec:act-sha}.
Fixed-interval anticipation methods are also sometimes used to generate dense predictions for a long block of future frames, which allows them to be compared against action segment anticipation using the metrics described in Section \ref{ssec:act-seg}.

\subsubsection{Approaches Overview}

Models for fixed-interval action methods need a component for processing observed video segments, and a component for predicting future actions. The input processing component of these models look similar to the short-term action anticipation models in the previous section. For instance, existing work uses Markov Random Field (MRF) \cite{chakraborty2015context}, RNNs \cite{furnari2019would,furnari2020rolling,qi2021self,gao2017red}, transformers \cite{zhao2022real,zhong2023anticipative} or graph representations of videos \cite{sun2019relational,liu2020spatiotemporal}. The prediction component is typically another sequence model, like an RNN \cite{sun2019relational,furnari2020rolling,qi2021self} or transformer \cite{zhong2023anticipative}. Using a sequence model as a prediction head enables the model to make predictions on arbitrary horizons. Reinforcement learning is also used to reason about the semantic states and future goal states \cite{gao2017red,rhinehart2017first} of an action sequence. Auxiliary losses for fixed-interval anticipation also look similar to those for other anticipation tasks, including feature forecasting losses and temporal cycle consistency losses \cite{qi2021self}.

\begin{figure}[!t]
    \centering
    \includegraphics[width=\linewidth]{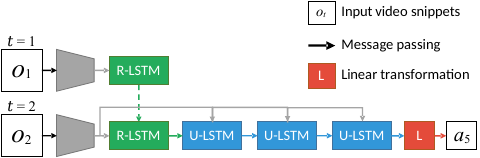}
    \caption{
        The RU-LSTM \cite{furnari2019would,furnari2020rolling} is typical of sequence-to-sequence action forecasting architectures.
        In this diagram, we see the \textit{rolling} LSTM (R-LSTM) process two observed frames, $o_1$ and $o_2$.
        The \textit{unrolling} LSTM (U-LSTM) is then unrolled for three steps to predict the action $a_5$ that happens three time steps into the future.
        \textit{Adapted from Figure~4 of Furnari et al. \cite{furnari2020rolling}.}
    }
    \label{fig:highlight-ru-lstm}
\end{figure}

\subsubsection{Inductive Biases}

Recurrent sequence-to-sequence models are the most common architecture for fixed-interval action anticipation \cite{gao2017red,furnari2019would,furnari2020rolling,piergiovanni2020adversarial}. Gao et al. \cite{gao2017red} are the first to use LSTM-based encoder and decoder to model multiple past actions and forecast the future action sequence. The regular regression loss and cross-entropy loss are applied on the future visual features and future action class labels. In addition, they propose to use an reinforcement module to provide guidance on the entire predicted action sequence. The Rolling-Unrolling LSTM (RU-LSTM) \cite{furnari2019would,furnari2020rolling} is an important milestone which applies a sequence-to-sequence RNN architecture -- a \textit{rolling} LSTM encoding features from a sequence of frames, and an \textit{unrolling} LSTM predicting future actions at fixed rate (Figure \ref{fig:highlight-ru-lstm}). A separate instance of this architecture is learned for each modality, and then the resulting predictions are fused using learned attention over the modalities. In order to provide additional supervision for the intermediate representation, the unrolling LSTM is pretrained to do action recognition conditioned on true future frame features at each time step (the paper refers to this pretraining strategy as ``sequence completion pretraining'').

Most prior work focuses on action anticipation for a single person, forecasting the action sequence of multiple identities remains a challenging problem. DR$^2$N \cite{sun2019relational} performs multi-person action anticipation by combining a frame-level graph network with inter-frame RNNs for each actor. Specifically, the model uses Faster R-CNN \cite{ren2015faster} to track people across observed frames. Each person is associated with a GRU, and the GRU's state is updated at each frame with a combination of RoI-pooled appearance features and graph-like attention over representations for other detected people. The idea of extracting identity specific features with pretrained detection models and involving them in action anticipation networks is also adopted by Li et al. \cite{li2021restep}. Similarly, Liu et al. \cite{liu2020spatiotemporal} propose a spiritually similar graph-structured neural network for predicting pedestrian actions from car-mounted cameras. Their proposed method first parses each frame into a series of bounding boxes corresponding to pedestrians, traffic signs and other cars, and then models them with a graph convnet followed by a GRU to capture temporal dependence.

\subsubsection{Multi-modal Models}
The only model focuses on multi-modal fusion in this task is AFFT from Zhong et al. \cite{zhong2023anticipative}. They are the first to involve the audio stream besides visual input to boost the anticipation performance in egocentric vision. They compare three fusion strategies based on the transformer architecture: (1) fusing multiple modalities at each time step independently, (2) fusing multiple modalities by self-attention layer at all time steps at once, and (3) using visual input as the main modality and fusing the other modalities by cross-attention layers. The first strategy performs the best in their experiments. They argue that this strategy decouples the temporal modeling from modality fusion and thus makes the problem easier to handle.

\subsubsection{Pretraining and Objectives}

SRL \cite{qi2021self} aims to improve the prediction accuracy of RNN-based models over long anticipation horizons.
The authors argue that fixed-interval predictions are particularly difficult for RNNs because they suffer from compounding error. They propose three ways to alleviate this problem, including an attention mechanism over past frames, a contrastive feature forecasting loss, and auxiliary losses for noun and verb anticipation. These modifications together result in higher accuracy than prior methods in several video datasets. In order to avoid compounding errors, Liu et al. \cite{liu2022hybrid} use an additional memory bank to preserve accurate long-term connections. The memory bank is composed of $K$ learnable vectors of $d$ dimensions. The features of the observed segment are represented by a weighted sum of these memory vectors, where the weights are obtained by calculating the cosine similarity of the observed segment embedding and each memory vector. The matrix is optimized towards sparsity and diversity by two regularization terms. They also use a one-shot similarity maximization strategy to maximize the similarity of target frame feature and the feature predicted from prior frames.

\subsection{Long-Term Action Anticipation}
\label{ssec:act-seq-fc}

Long-term action anticipation is the task of predicting the labels of the next $Z$ unobserved action segments, but not the associated start or end times \cite{grauman2022ego4d}. Specifically, given a sequence of observed video up to action segment $a_i$, the algorithm must predict the labels $a_{i+1}, a_{i+2}, \ldots, a_{i+Z}$ of the next $Z$ actions.
Typically models are rewarded for predicting future actions in the correct order, although some methods also do bag-of-actions forecasting in which the aim is merely to predict which labels will occur in the future without predicting the order \cite{sener2019zero}.

\subsubsection{Metrics}

Several metrics have been proposed for long-term action anticipation. The simplest one ignores the sequential nature of the task. It evaluates the prediction at each time step independently \cite{ng2020forecasting} or treats the sequence as an unordered set. Other work uses metrics from the natural language community, such as BLEU and METEOR, that reward correctly ordered predictions without requiring an exact temporal match between predicted and ground truth labels \cite{ng2020forecasting,sener2019zero}. In this same vein, Grauman et al. \cite{grauman2022ego4d} propose to use a normalized Levenshtein string distance. Levenshtein distance is a common sequence accuracy metric, yet lacks the free parameters of the more complex metrics from the NLP community. Further, for models that can produce several sequence predictions, they suggest taking the \textit{minimum} string distance across $N$ samples, which makes the forecasting task slightly easier.

\subsubsection{Approaches Overview}

Like fixed-interval action anticipation models, long-term action anticipation models need to both process a sequence of observed video frames and produce a sequence of future actions.
Observed video is typically processed with some combination of CNNs \cite{grauman2022ego4d}, RNNs \cite{ng2020forecasting,schydlo2018anticipation,gammulle2019forecasting} and transformers \cite{sener2020temporal,roy2021action,zhang2024object}. There also exist some methods parsing the video into more structured representations, such a 3D map \cite{bokhari2016long}, and a set of high-level scene attributes or a graph that captures transitions between scenes with similar affordances \cite{qi2017predicting}.
Sequence prediction can be achieved with common sequence models like RNNs \cite{ng2020forecasting,sener2019zero} or transformers \cite{mascaro2023intention}, but some work has also used MRFs \cite{chakraborty2015context}, VAE \cite{mascaro2023intention} or RL policies \cite{bokhari2016long}.
Aside from model architecture, long-term action anticipation methods also differ in their choice of sequence prediction and pretraining objectives. Notable design choices in this regard include adversarial training for sequence prediction \cite{piergiovanni2020adversarial}, CLIP pretraining \cite{das2022video+}, reconstruction objectives \cite{mascaro2023intention}, and pretraining on text-based action sequence datasets \cite{sener2019zero,sener2022transferring}.

\subsubsection{Inductive Biases}

Like fixed-interval anticipation, RNN-based architecture is most commonly applied in long-term action anticipation \cite{ng2020forecasting,schydlo2018anticipation,gammulle2019forecasting,liu2022hybrid}. Schydlo et al. \cite{schydlo2018anticipation} use the gaze and posture as visual cues (instead of RGB frames) to forecast future action sequences by implementing a straightforward RNN-based architecture. They then input the final hidden state of the encoder to an RNN-based decoder. Ng et al. \cite{ng2020forecasting} propose a recurrent encoder-decoder model for long-term action anticipation. First, they use a bidirectional GRU encoder to process a sequence of input video frames. Then, a unidirectional GRU decoder can output a sequence of future actions by making use of a learnable module that allows it to attend over the encoder activations. Sener et al. \cite{sener2020temporal,sener2021technical} propose a hierarchical pooling scheme (TempAgg) that makes it possible to process long contexts at variable resolution. TempAgg summarizes frame embeddings from the ``recent past'' (the last few seconds or minutes) by max-pooling different durations of recent frames, and summarizes the ``spanning past'' (the entire duration of observed video) by max-pooling the entire observed video at various temporal resolutions. The resulting features are further aggregated by learned transformer-based neural networks in order to produce future action predictions. 

Considering the uncertainty in the future action sequence, Qi et al. \cite{qi2017predicting} propose spatial-temporal And-Or graph (ST-AOG) to capture the hierarchical characteristics of actions over time. Mascaro et al. \cite{mascaro2023intention} choose to model the uncertainty by using a variational auto-encoder (VAE) which shows a great advantage in modeling egocentric actions.

Human-object interaction is another feature that can be leveraged \cite{roy2021action,liu2022joint,zhang2024object}. Roy et al. \cite{roy2021action} use Mask-RCNN \cite{he2017mask} to detect the human body and objects involved in the past actions and extract the corresponding features. They further learn the weight for each object by concatenating object features with human body features followed by a non-linear layer. An outer production is used between the weights and concatenated features to get a cross-attention map which is the input to a transformer model.

Different than prior work that predicts action labels, Abdelsalam et al. \cite{abdelsalam2023gepsan} generate natural language to describe the following steps given the observed video frames. The model is trained with texts and then transfers to videos in zero-shot setting.

\subsubsection{Pretraining and Objectives}

\begin{figure}[!t]
    \centering
    \includegraphics[width=\linewidth]{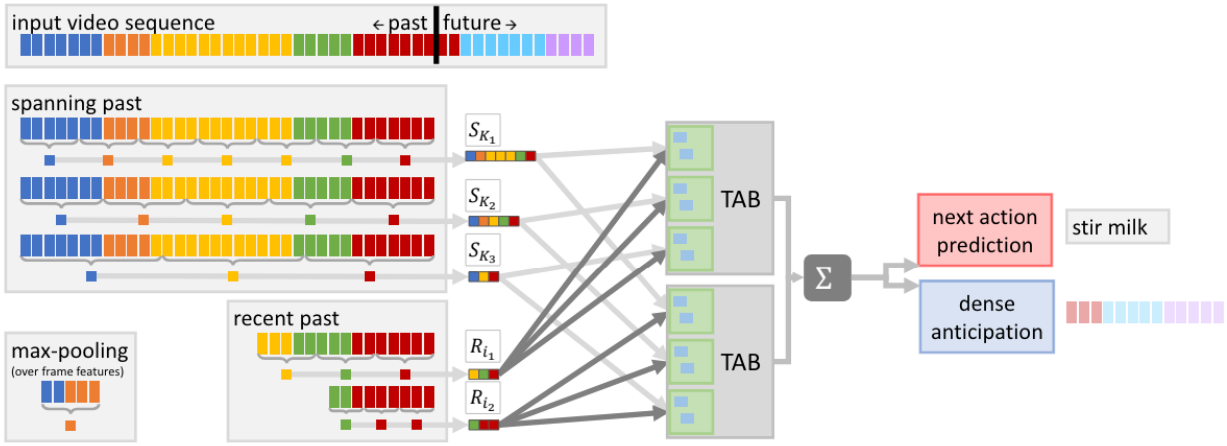}
    \caption{
        Schematic view of Sener et al.'s text-based long-term action anticipation model \cite{sener2020temporal}.
        The model uses three scales to aggregate the whole past video contexts and two starting points to compute recent observed history by max-pooling over frame features. 
        There are six (3$\times$2) combinations of the two kinds of features in total, which are fused in their proposed Temporal Aggregation Block (TAB). 
        \textit{Adapted from Figure~2 of Sener et al. \cite{sener2020temporal}.}
    }
    \label{fig:highlight-sener-zero-shot}
\end{figure}

As a baseline for the Ego4D long-term action anticipation benchmark, Grauman et al. \cite{grauman2022ego4d} propose to use a SlowFast network \cite{feichtenhofer2019slowfast} to extract features from a fixed-length video clip, then use $Z$ independent MLP prediction heads to predict each future action. Das et al. \cite{das2022video+} augment this model with two additional CLIP-based inputs.

Other novel pretraining strategies for long-term action anticipation include only using visual features \cite{narasimhan2023learning,xue2023egocentric} and leveraging texts as an additional modality \cite{sener2019zero,sener2022transferring,tan2023multiscale,ashutosh2023hiervl,zhao2024antgpt}. Sener et al. \cite{sener2019zero,sener2022transferring} show how to use a purely text-based recipe dataset as pretraining data for visual action forecasting in cooking videos (Figure \ref{fig:highlight-sener-zero-shot}). They first use the text-based data to learn three components: a sentence encoder that can represent individual recipe steps, a recipe RNN to predict the next recipe step, and a sentence decoder to decode the predicted steps into text. They then use their video data to train a video encoder that produces recipe step representations from videos and replace the sentence encoder (text input) with the video encoder (visual input) at test time. Ashutosh et al. \cite{ashutosh2023hiervl} propose to learn short video representation with contrastive learning scheme, and then integrate them in a long video embedding. The learned representation is used for many downstream tasks including long-term action forecasting. Zhao et al. \cite{zhao2024antgpt} use an off-the-shelf LLM to infer the final target by in-context learning based on the action labels of observed actions. The final target is fed into a temporal model together with observed RGB frames to forecast long-term actions. Mittal et al. \cite{mittal2024can} directly finetune a multi-modal large language model (MLLM) with two losses to generate more plausible action sequences.

\subsection{Action Segment Anticipation}\label{ssec:act-seg}

Action segment anticipation extends long-term action anticipation to also predicting the start times and durations of future actions. Formally, we input the first $i$ frames $o_1, o_2, \ldots, o_i$ of a $T$-frame video $\mathcal V = (o_1, o_2, \ldots, o_T)$ into a model, and denote the next $Z$ actions that start after time $i$ as $a_{i+1}, a_{i+2}, \ldots, a_{i+Z}$. In action segment anticipation, the model must predict the future label sequence $a_{i+1}, a_{i+2}, \ldots, a_{i+Z}$, as well as corresponding start and end times $(t_{i+1}, t_{i+1}'), (t_{i+2}, t_{i+2}'), \ldots, (t_{i+Z}, t_{i+Z}')$. Depending on the model, these start and end times might be parameterized in different ways. For example, predicting the inter-arrival times for each pair of future actions rather than predicting start times directly. Furthermore, some work predicts only the start time or the duration of future actions, and not both.\footnote{Some datasets provide a single action label for every frame in the dataset, in which case start times can be deduced from durations and vice versa.}

Action segment anticipation models and fixed-interval anticipation models are sometimes evaluated head-to-head.
These two types of models can be viewed as different ways of assigning an action label to future frames, which is sometimes called \textit{dense} action anticipation. Consequently, it's possible to evaluate both kinds of models with the same evaluation datasets and metrics.

\subsubsection{Metrics}

Abu Farha et al. \cite{abu2018will,abu2019uncertainty} propose to feed the first $x\%$ of a video into the model, and then measuring the accuracy of the actions predicted for the next $y\%$ of the video. For each chosen value of $x$ and $y$, this top-1 accuracy calculation repeated for each class to report a Mean over Classes (MoC) accuracy. Performing well under this metric requires a model produce accurate labels, and also reward models for producing accurate temporal boundaries so that the predicted action segments line up with the ground truth action segments.

Morais et al. \cite{morais2020learning} argue that the implicit evaluation of temporal boundaries via MoC accuracy is inadequate. To substantiate this claim, they show that a simple model that predicts just one long future action segment (i.e., a single label for the rest of the video) can obtain competitive results with prior methods, particularly for smaller values of the temporal offset $y$ (e.g., predicting 10\% into the future). Instead, they propose a metric in which a given segment prediction is ``correct'' if it has IoU $\geq k$ with a ground truth segment that has the same action label.
They then use this notion of correctness to compute precision and recall over the full dataset, which is aggregated into a single F1 score that they refer to as \textit{F1@$k$}.
In experiments, F1@$k$ appears to produce a greater gap between methods that predict only one long future segment and methods that predict multiple segments.

\subsubsection{Approaches Overview}

At a high level, action segment anticipation is a very similar task to long-term action anticipation and fixed-interval action anticipation. Thus models for action segment anticipation look broadly similar: a sequence model processes some observed video, and then a second sequence model conditions on the first model's output to predict a sequence of future actions. These models include various combinations of CNNs \cite{abu2018will}, sequence-to-sequence RNNs \cite{abu2018will,abu2019uncertainty,morais2020learning,zhao2020diverse,abu2021long,loh2022long}, and transformers \cite{nawhal2022rethinking,gong2022future}. The main additional complication is that action segment anticipation models must explicitly predict action durations, and not just labels.

The problem of compounding error, which we discussed previously in the context of other sequence forecasting tasks, has received special attention in the action segment forecasting literature. Potential ways of addressing this problem include simultaneously predicting a block of future actions with a bidirectional \cite{nawhal2022rethinking,gong2022future} or multi-head \cite{mahmud2021prediction} model, incorporating hierarchy into a prediction model \cite{morais2020learning}, or using cycle consistency losses \cite{abu2021long}.

Another issue that has received special attention is the problem of modeling stochastic sequences of action labels and durations. The simplest auto-regressive models make deterministic predictions of action duration \cite{abu2018will}, but action duration can vary greatly based on hard-to-model factors. Past work models the resulting stochasticity by endowing the predicted duration \cite{abu2019uncertainty} or the latent representation \cite{loh2022long} with a probability distribution, or by combining a probabilistic model with diversity losses to encourage a range of possible sequence predictions \cite{zhao2020diverse}.

\subsubsection{Inductive Biases}

Sequence-to-sequence (seq2seq) RNNs are a common way to perform action segment forecasting. In these models, one RNN is used to process observed frame features or action labels and durations, and then a second RNN conditions on the output of the first RNN to forecast a sequence of future action labels and durations. Abu Farha et al. \cite{abu2018will} are the first to study action segment anticipation problem by using a simple RNN. It also predicts the duration of the currently occurring (partially-observed) action, since the model may be invoked before the most recent action has ended. Predictions from this model can be auto-regressively fed back into the RNN in order to forecast more than one action segment ahead. Given the duration predictions are deterministic, the model cannot capture all of the inherent randomness in action anticipation. To complement for the defect of \cite{abu2018will}, Abu Farhar et al. \cite{abu2019uncertainty} propose a variation on this model -- Uncertainty-Aware Action Anticipation (UAAA). They model the duration with a Gaussian distribution rather than providing only a point estimate, which makes it possible to model stochasticity in the action duration and statistical dependence between the action label and duration at a given time step. The uncertainty can also be captured by a diffusion model \cite{zhong2023diffant}.

\begin{figure}[!t]
    \centering
    \includegraphics[width=0.95\linewidth]{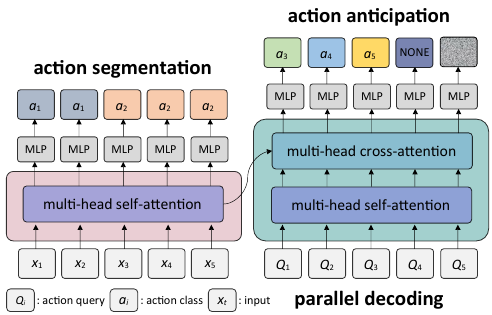}
    \caption{
        FUTR \cite{gong2022future} is an example of a model that avoids the problem of compounding error in auto-regressive models by predicting a whole batch of future actions at the same time.
        It processes observed frame embeddings with a bidirectional transformer encoder that is trained for action segmentation.
        It then applies a bidirectional transformer decoder to simultaneously predict a fixed number of future actions.
        \textit{Adapted from Figure~1 of Gong et al. \cite{gong2022future}.}
    }
    \label{fig:highlight-futr}
\end{figure}

Most methods that we've covered so far predict future action segments auto-regressively, one at a time. One fallback of this strategy is compounding error. Another strategy, which we discuss here, is to have a model simultaneously predict a fixed number of future action segments at the same time, rather than doing auto-regressive prediction \cite{mahmud2021prediction,nawhal2022rethinking,gong2022future}.

Two examples of such models are ANTICIPATR \cite{nawhal2022rethinking} and FUTR \cite{gong2022future}. They both use bidirectional transformers to simultaneously predict an entire sequence of future actions, aiming to avoid the problem of compounding error. ANTICIPATR treats the output as an unordered set of segments that must be matched to ground truth segments using predicted time information (like DETR \cite{carion2020end}), while FUTR outputs an ordered sequence that is meant to have a one-to-one correspondence to the ground truth future segments. ANTICIPATR is pretrained for bag-of-actions forecasting, while the decoder in FUTR is jointly trained to do action segmentation (Figure \ref{fig:highlight-futr}).

\subsubsection{Pretraining and Objectives}

Zhao et al. \cite{zhao2020diverse} propose a sequence-to-sequence RNN trained with a modified GAN loss. They use the Gumbel-Softmax trick to enable them to backpropagate through the sampling step for their categorical actions. They also predict discretized, rather than continuous, time intervals. In order to produce diverse predictions, their GAN generator uses a diversity loss proposed by Yang et al. \cite{yang2019diversity} to ensure that different noise vectors map to different output sequences. Loh et al. \cite{loh2022long} propose MAVAP which is a variant of the VRNN \cite{chung2015recurrent} (a kind of recurrent VAE). They argue that the stochastic representation aspect of their model is a better fit for the unpredictable nature of action sequences, and also that it is useful to model the dependence between action segment labels and durations.

\subsection{Procedure Planning}

Given a video clip depicting an initial state $s_0$ and a video clip depicting a goal state $s_{Z+1}$, procedure planning is the task of predicting a single sequence of $Z$ actions that a human would use to get from $s_0$ to $s_{Z+1}$ \cite{chang2020procedure}. The length $Z$ of the action sequence is assumed to be known and constant.

Procedure planning can be viewed as a goal-conditioned version of long-term action anticipation.
The general task of predicting an intervening sequence of actions conditioned on a start and end states has also been investigated in other settings like robotic imitation learning and animal behavior modeling, as noted by Chang et al. \cite{chang2020procedure}. Moreover, it's similar to the task of \textit{walkthrough planning}, which requires a model predict visual representations of intermediate states instead of predicting intermediate actions \cite{kurutach2018learning}. In our survey, we narrow down our survey to predicting the procedures of human's daily actions. The general policy prediction in reinforcement learning for robots is beyond our scope.

\subsubsection{Metrics}

Chang et al. \cite{chang2020procedure} propose three metrics for procedure planning. The first is \textit{success rate}: the fraction of predicted plans that match the ground truth plan at \textit{every} action, averaged across the entire evaluation dataset. The second is \textit{accuracy}: the fraction of predicted actions that match the ground truth actions at the corresponding time step, averaged across the dataset. The third is \textit{mean intersection over union} (mIoU). If $\hat{A}$ is the set of actions in the predicted plan $\hat{P}$, and $A$ is the set of actions in the ground truth plan $P$, then the IoU for $P$ and $\hat{P}$ is
\begin{equation}
    \operatorname{IoU}(P, \hat P) = \frac{|A \cap \hat A|}{|A \cup \hat A|}.
\end{equation}
mIoU averages the IoU across all procedure planning problems in a dataset.

\subsubsection{Benchmarks}

Procedure planning has a slightly different input-output specification to the previous tasks, so we'll briefly describe how existing action recognition datasets can be repurposed for procedure planning. Given an annotated video $\mathcal V$ with $N \geq Z+1$ annotated action segments, one can sort the action segments by start time to produce a segment list
\begin{equation}
\small
    \mathcal S = \langle(a_1, o_1, t_1, t'_1), (a_2, o_2, t_2, t'_2), \ldots, (a_N, o_N, t_N, t'_N)\rangle,
\end{equation}
where $a_i$ is the $i$-th action which runs from time $t_i$ to $t'_i$, and $o_i$ is the visual observation obtained by trimming $\mathcal V$ from $t_i$ to $t_i'$.
One can then enumerate all subsequences with the length of $Z+1$ in a sliding window fashion to produce $N-Z$ procedure planning problems. The $i$-th such problem will have an initial state observation $o_i$, a goal state observation $o_{i+Z+1}$, and a ground truth plan $\langle a_{i+1}, a_{i+2}, \ldots, a_{i+Z}\rangle$.
This procedure can be repeated for all videos in the dataset to produce a complete benchmark.\footnote{
    Sun et al. \cite{sun2022plate} propose an alternative method of generating procedure planning problems that produces one problem per video $\mathcal V$ by doing random sampling instead of sliding-window enumeration, among with other small changes documented by \cite{zhao2022p3iv}.
    The numbers resulting from this method should be roughly comparable to the original method, but we highlight them with an ``alt eval'' in tables for clarity.
}

The standard dataset used to evaluate existing procedure planning work is CrossTask \cite{zhukov2019cross}. In addition, Zhao et al. \cite{zhao2022p3iv} evaluate their method on COIN \cite{tang2019coin} and NIV \cite{alayrac2016unsupervised}, while Sun et al. \cite{sun2022plate} even evaluate their method on a real robot arm \cite{sun2022plate}.

\subsubsection{Approaches Overview}

Existing methods for procedure planning generally have two components.
The first is a probabilistic generative model that generates a sequence of observations and actions, perhaps represented by an RNN \cite{chang2020procedure,bi2021procedure} or a transformer \cite{sun2022plate,zhao2022p3iv,wang2023pdpp}.
The second is a specific inference procedure that takes a (initial observation, goal state) pair $(o_0, o_{T+1})$ and returns a plan $\langle a_1, a_2, \ldots, a_Z \rangle$ of reaching the goal from the initial state under the sequence model.
For instance, existing work uses direct sampling from a policy \cite{bi2021procedure}, beam search \cite{chang2020procedure,sun2022plate}, Viterbi decoding \cite{zhao2022p3iv} or diffusion models \cite{wang2023pdpp}.

\subsubsection{Inductive Biases}

\begin{figure}[!t]
    \centering
    \includegraphics[width=\linewidth]{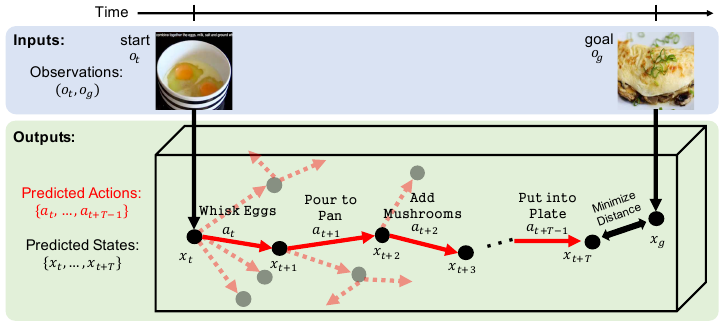}
    \caption{
        DDN \cite{chang2020procedure} is the prototypical approach to procedure planning.
        At a high level, the model is able to predict a next latent representation (``state'') from a latent representation and action, and a next action from the current state and prior action.
        At inference time, a policy or search algorithm must find a sequence of actions that transition from an initial observation to a supplied goal observation.
        \textit{Adapted from Figure~1 of Chang et al. \cite{chang2020procedure}.}
    }
    \label{fig:highlight-ddn}
\end{figure}

Chang et al. \cite{chang2020procedure} introduce the task of procedure planning and propose the Dual Dynamics Network (DDN) to solve it. The DDN's sequence model is composed of two parts. The first is a forward dynamics model $p(x_{i+1}|x_i,u_i)$ that predicts a latent representation $x_{i+1}$ of the next observation $o_{i+1}$ from latent representations $x_i$ and $u_i$ of the current observation $o_i$ and action $a_i$, respectively. The second is a conjugate dynamics model that predicts $u_i$ from $x_i$ and $u_{i-1}$. At inference time, DDN uses a procedure similar to beam search to find a goal-reaching plan (Figure \ref{fig:highlight-ddn}).
PlaTe \cite{sun2022plate} from Sun et al. is similar to DDN, but uses a transformer that conditions on goal representations to model state-action sequences, and does a normal beam search at inference time instead of using a custom procedure.
The Ext-GAIL method \cite{bi2021procedure} of Bi et al. similarly learns a goal-conditioned latent dynamics model $p(x_{i+1}|x_i,u_i,x_1,x_{T+1})$. However, it uses Generative Adversarial Imitation Learning (GAIL) \cite{ho2016generative} to obtain the conditional distribution $p(u_i|x_i)$ over latent actions. Wang et al. \cite{wang2023pdpp} propose PDPP that leverages the advanced diffusion models to learn the distribution of future actions. The denoising steps are applied to infer a sample of procedural steps.

\subsubsection{Pretraining and Objectives}

The P$^3$IV method of Zhao et al. \cite{zhao2022p3iv} is able to use simpler supervision sources at training time, as well as a faster inference procedure at test time. Instead of supervising training of the observation transition model using labeled observation intervals, P$^3$IV supervises the transition model using language descriptions of intermediate states. In principle, this allows their method to learn from sequences of paired symbolic action labels and natural language action descriptions, without labeled time intervals for each intermediate action. At inference time, P$^3$IV also differs from past work in that it computes a transition probability matrix for adjacent actions at each time step, and then applies a Viterbi decoder to recover an action sequence.

\section{Datasets}\label{sec:datasets}

\begin{table*}[!t]

\renewcommand{\arraystretch}{1.3}
\scriptsize
\caption{Datasets used for action anticipation in prior work. The official statistics of some datasets are unavailable which are denoted as unknown (Unk.). Kinetics-700 is not used for action anticipation problem, but it's widely used for pretraining. Thus we also list it in this table for a thorough review.}
\label{tab:datasets}
\centering
\begin{threeparttable}
\resizebox{\linewidth}{!}{
\begin{tabular}{lccccccc}
\toprule
\textbf{Dataset} & \textbf{Year} & \textbf{View} & \textbf{\# Hours} & \textbf{\# Segments} & \textbf{\# Videos} & \textbf{\# Classes} & \textbf{Data \& Annotations} \\
\midrule
TV Human Interaction \cite{patron2010high,patron2012structured} & 2010  & Exo & Unk. & 200 & 300 & 4 & RGB videos, action classes, upper body bboxes, head orientation \\

VIRAT \cite{oh2011large} & 2011 & Exo & 29 & $\sim$8,000 & $\sim$119 & 46 
& RGB videos, action classes, object bboxes\\

GTEA \cite{fathi2011learning,li2015delving} & 2011 & Ego & 0.6 & 525 & 22 & 71 & RGB videos, action classes, hand masks \\

GTEA Gaze \cite{fathi2012learning} & 2012 & Ego & 1 & 331 & 17 & 40 & RGB videos, gaze, action classes \\

MPII Cooking \cite{rohrbach2012database} & 2012 & Exo & $\sim$8 & 5,609 & 44 & 65 & RGB videos, action segments, upper-body poses\\

MPII Composites \cite{rohrbach2012script} & 2012 & Exo & 15.9 & 8,818 & 212 & 218$^{\text{\textdaggerdbl}}$ & RGB videos, action segments, dish labels, scripts \\
\quad + TACoS(-ML) \cite{regneri2013grounding,rohrbach2014coherent} & 2013 & Exo & 15.9 & 8,818 & 212 & 218$^{\text{\textdaggerdbl}}$  & Adding captions (temporally aligned), caption similarity, tree-level scripts \\

50Salads \cite{stein2013combining} & 2013 & Exo & 4.5 & 966 & 50 & 17 & RGB-D videos, accelerometer data of objects, action segments, scripts \\

Breakfast Actions \cite{kuehne2014language} & 2014 & Exo & 77$^{\S}$ & 11,267$^{\S}$ 
& 1,989$^{\S}$ & 48$^{\S}$ & RGB videos, action segments, dish names \\

THUMOS \cite{THUMOS14} & 2014 & Exo & 20 & 6,365 & 2500 & 20 & RGB videos, action segments, action attributes \\

GTEA Gaze+ \cite{fathi2012learning,li2015delving} & 2015 & Ego & 9 & 1,958 & 37 & 44  & RGB videos, gaze, action classes, hand masks \\

MPII Cooking 2 \cite{rohrbach2015recognizing} & 2015 & Exo & 27 & 14,105 & 273 & 222$^{\text{\textdaggerdbl}}$ & RGB videos, action segments, upper-body poses, hand locations, scripts\\

ActivityNet \cite{caba2015activitynet} & 2015 & Mix & 849 & 39,200 & 27,811 & 203 & RGB videos, action segments, ATUS-derived label taxonomy \\

TVSeries \cite{de2016online} & 2016 & Exo & 16 & 6231 & Unk. & 30 & RGB videos, action segments, metadata (e.g., occlusion, viewpoint, etc.)\\

Charades \cite{sigurdsson2016hollywood} & 2016 & Exo & $\sim$82 & 66,500 & 9,848 & 157 & RGB videos, action segments, active object classes, scripts, video descriptions \\

Narrated Instruction Videos \cite{alayrac2016unsupervised} & 2016 & Exo & $\sim$7.4 & Unk. & 150 & Unk. & RGB videos, action segments, ASR captions, scripts \\

EGTEA Gaze+ \cite{li2018eye} & 2018 & Ego & 29 & 10,325 & 86 & 106 & RGB videos, gaze, action classes, hand masks \\

AVA \cite{gu2018ava} & 2018 & Exo & 107.5 & Unk. & 430 & 80 & RGB videos, multi-person actions, person bboxes and tracks \\

MultiTHUMOS \cite{yeung2018every} & 2018 & Exo & 30 & 38,690 & 400 & 65 & RGB videos, action segments \\

YouCook2 \cite{zhou2018towards} & 2018 & Exo & 176 & $\sim$15,400 & 2000 & N/A & RGB videos, audio, action segment descriptions (natural language) \\

Epic-Kitchens 55 \cite{damen2018scaling} & 2018 & Ego & 55 & 39,596 & 432 & 149$^\dagger$ & RGB videos, actions segments, active object bboxes \\

CrossTask \cite{zhukov2019cross} & 2019 & Exo & 376 & 34,780 & 4,700 & 83 & RGB videos, audio, action segments, scripts \\

COIN \cite{tang2019coin} & 2019 & Mix & 477 & 46,354 & 11,827 & 778 & RGB videos, action segments (semantic hierarchy) \\

Epic-Kitchens 100 \cite{damen2020rescaling} & 2020 & Ego & 100 & 89,977 & 700 & 4,053 & RGB videos, proprioceptive data, actions segments, active objects bboxes/masks\\

Assembly101 \cite{sener2022assembly101} & 2022 & Mix & 513 & 1,013,523 & 4321 & 1380/202$^*$ & RGB videos, actions classes, 3D hand poses \\

Ego4D \cite{grauman2022ego4d} & 2022 & Ego & 120 & 77,002 & Unk. & Unk. & RGB videos, action classes, object bboxes, narrations \\

\quad + Ego4D Goal-Step \cite{song2024ego4d} & 2023 & Ego & 430 & 48,000 & 851 & 514 & Providing goals and fine-grained steps annotations for videos in Ego4D \\

\arrayrulecolor{black!20}\midrule\arrayrulecolor{black}

Kinetics-700-2020 \cite{smaira2020short} & 2020 & Mix & $\sim$1,800 & $\sim$650,000 & $\sim$650,000 & 700 & RGB videos, action classes (loose semantic hierarchy), audio \\
\bottomrule
\end{tabular}
}

\begin{tablenotes}
\item $\dagger$ Class counted for EK55 only includes classes with $>$50 samples.
\item \textdaggerdbl \ MPII Composites and derivative datasets report only the number of attributes, which is the number of verb classes plus the number of noun classes.
\item $\S$ Video statistics for Breakfast Actions count different views of the same action as distinct videos, which inflates the number of hours. The number of action \\ classes and segments refers only to the coarse-grained annotations \cite{kuehne2014language}.
\item $^*$ Assembly101 has 202 coarse-grained classes and 1380 fine-grained classes.
\end{tablenotes}
\end{threeparttable}
\end{table*}

There are many datasets of egocentric and exocentic videos used for action anticipation training and evaluation, spanning TV series \cite{patron2010high,patron2012structured}, cooking activities \cite{stein2013combining,rohrbach2012database,kuehne2014language}, driving videos \cite{kotseruba2016joint}, furniture assembly, etc. We summarize frequently-used datasets in Table \ref{tab:datasets} and elaborate additional details about video recording, annotations of each dataset and commonly-used metrics in the supplementary.

\subsection{Exocentric Video Datasets}

The earliest dataset for action anticipation is TV Human Interaction (TVHI) \cite{patron2010high,patron2012structured}. TVHI consists of 200 video clips from 20 different TV shows covering four actions -- hand shakes, high fives, hugs and kisses. The remaining 100 video clips don't contain the four actions (i.e., negative samples). Similarly, TV Series \cite{de2016online} also comprises 16-hour videos from 6 TV series covering 30 actions. This dataset contains extra metadata such as whether the person is occluded, whether the video is shot from frontal or side view, etc.

Apart from TV shows, another important data source for action anticipation is cooking videos \cite{stein2013combining,rohrbach2012database,kuehne2014language,fathi2012learning,li2015delving,li2018eye,damen2020rescaling,damen2018scaling,zhou2018towards,salvador2017learning,marin2021recipe1m+}. 50 Salads \cite{stein2013combining} consists of 50 videos (totaling 4 hours) of 25 participants preparing salads. Labels of 17 actions and accelerometer data of objects are provided in this dataset. Breakfast Actions \cite{kuehne2014language} is another widely-used cooking video dataset composed of 52 participants performing 10 cooking activities. They also annotate 48 fine-grained actions for all videos. YouCook2 \cite{zhou2018towards} collects cooking videos from YouTube and provide natural language descriptions for each action segment.

In terms of driving, there are some datasets including driver's actions (e.g., JAAD \cite{kotseruba2016joint}, Brain4Car \cite{jain2016brain4cars} and HDD \cite{ramanishka2018toward}) and pedestrian's actions (e.g., PIE \cite{rasouli2019pie}). To include more diverse activities, many datasets cover multiple daily activities \cite{oh2011large,THUMOS14,THUMOS15,caba2015activitynet,gu2018ava}, such as furniture assembling, reading, exercising and car repairing. The most widely-used dataset is THUMOS \cite{THUMOS14}. However, the training set only has trimmed videos that cannot be used for temporal action anticipation. Thus prior work uses the untrimmed videos in validation and test sets which totals 20-hour videos covering 20 action classes. Yeung et al. \cite{yeung2018every} further expand this dataset with denser annotations and more action classes (65 action classes). CrossTask \cite{zhukov2019cross} contains activities of cooking, car maintenance and craft. Each activity has a script describing the involved actions in sequence, which is a key difference compared with other datasets. Hence, CrossTask is mainly used for procedure planning task.

\subsection{Egocentric Video Datasets}

The study of egocentric video dataset used for action anticipation begins with GTEA \cite{fathi2011learning,li2015delving} and draws more attention in recent years \cite{yu2020bdd100k,nakamura2017jointly} because of the popularity and commercialization of wearable devices. More large-scale egocentric video datasets \cite{damen2018scaling,damen2020rescaling,grauman2022ego4d} are released to foster novel approaches applied to the egocentric view.

GTEA dataset \cite{fathi2011learning} includes 22 egocentric videos totaling 0.6 hours of cooking scenarios. All videos are untrimmed thus making it an ideal vehicle for action anticipation. GTEA Gaze \cite{fathi2012learning} captures videos in a similar way. They additionally provide eye-tracking data to reveal human's attention. GTEA Gaze+ \cite{li2015delving} and EGTEA Gaze+ \cite{li2018eye} are two expansions to the original datasets by involving longer videos and well-instructed cooking steps. The final dataset contains 29-hour untrimmed videos covering 106 action classes. Another well-known egocentric video dataset of cooking activities is Epic-Kitchen 55 \cite{damen2018scaling}. It includes 55-hour egocentric videos of subjects preparing food in different kitchens. The labels of action classes, start time and end time are provided, which can be used for both action- and time-related anticipation. Epic-Kitchen 100 \cite{damen2020rescaling} expands it by involving additional 45 hours of videos. Recently, Ego4D \cite{grauman2022ego4d} is released with a specific benchmark for action anticipation with 120 hours of videos. The annotations include high-level action classes, object bounding boxes and narrations in natural language.

\section{What's next for forecasting?}
\label{sec:future}

Though many action anticipation approaches and datasets have been proposed, we still point out some weaknesses and understudied directions in this area. 

\textbf{Avoiding Error Accumulation.} In action sequence forecasting (e.g., fixed-interval action anticipation and action segment anticipation), many existing models tend to predict one action at a time and then feed the predicted action to the model as input for the next action prediction \cite{abu2018will,furnari2019would,furnari2020rolling,piergiovanni2020adversarial}. The important drawback of these models is the compounding error. One solution is forecasting the entire action sequence at once, such as ANTICIPATR \cite{nawhal2022rethinking} and FUTR \cite{gong2022future}. More advanced model architecture for sequence prediction demands future exploration. Another possible way is developing an auto-correction mechanism to correct some errors while predicting the next action iteratively. Using powerful large language models may be a promising way to detect and correct those errors.

\textbf{Quantitative Long-history Modeling.} The future actions are closely relevant with the most recent observed behaviors, whereas the long history still contributes to the forecasting. Several models are proposed to embed longer history in an efficient way \cite{manousaki2023vlmah,wu2022memvit,sener2020temporal}. They simply use the long history as a black box and rely on the optimizer to automatically learn the relationship of future and past actions. The quantitative analysis is still missing. It will be an important milestone if a novel approach is able to learn the quantitative contribution of each past action to the future actions and visualize the temporal dependence accordingly.

\textbf{More Investigation in Procedure Planning and Goal Anticipation.} In the seven sub-categories listed in this paper, final action and goal anticipation and procedure planning are still understudied problems. One possible reason is the scarcity of proper datasets having both high-level labels of goals and action labels of each step, which deserves more efforts. In addition, novel approaches specifically designed for the two problems are also necessary in the future research.

\textbf{Novel Design for Egocentric Vision and Multi-view Anticipation.} Egocentric action anticipation gets an increasing attention because of the release of large-scale egocentric video datasets \cite{damen2018scaling,damen2020rescaling,grauman2022ego4d}. The videos in these datasets are untrimmed thus serving as an ideal vehicle for learning the dependence of actions over time. However, most prior approaches evaluated on egocentric video datasets lack the specific design for egocentric view. The challenging nature, such as moving viewpoint, changing background, partial visibility of human body, is still unsolved. Another promising direction is leveraging the advantages of multiple views, including egocentric and exocentric views. The egocentric perspective contains many embodied clues such as gaze shift, head movement and head-hand coordination, which implies people's attention and next possible action. In contrast, the exocentric perspective has richer contexts of body movements and environment. How to exploit the synergy of multi-view representation deserves more attention in future work.

\textbf{Social Scenarios.} Most of prior work focuses on action anticipation of a single person. The social scenarios involving near-field interactions of multiple people are not fully studied. Though Li et al.\cite{li2021restep} propose an novel approach for multi-person action modeling based on bounding boxes, their datasets are still composed of single-person actions, such as walking, running, etc. The inter-person actions are neglected. TVHI \cite{patron2010high} and TV Series \cite{de2016online} contain a few social actions (e.g., hugging, shaking hands), but they only cover very few social interactions without bounding boxes of subjects. Forecasting the future social interactions is a promising research direction and many social interaction datasets are already established \cite{zadeh2019social,lai2023werewolf,lee2024modeling}.

\textbf{Common Sense Involvement.} When humans are forecasting the future actions someone may take, we can leverage some common senses for accurate inference. For example, when someone takes off clothes in an indoor situation, we may foresee that they will take a shower or do the laundry. If the same action is observed outdoor, we may predict that they are going to swim. On the contrary, existing models learn the causal dependence of actions from only the training data, making them subject to overfitting. How to obtain and involve common sense is a key step for addressing this issue.

\textbf{Utilization of Foundation Models.} Recently, the emerging foundation models have shown the effectiveness in solving many traditional computer vision problems. In terms of action anticipation, some pioneer work includes using large language models \cite{ghosh2023text,zhao2024antgpt} and vision-language models \cite{zhang2024object,mittal2024can}. These models are pretrained with large-scale corpora or image-text pairs so that they learn better semantic representations and image-text associations. How to leverage these models more effectively deserves more investigation. Another unstudied direction is using purely vision-based foundation model such as DINO \cite{caron2021emerging} and SAM \cite{kirillov2023segment}. The vision foundation models show superiority in learning embeddings of various objects which can benefit object-centric action anticipation \cite{roy2021action,liu2022joint,zhang2024object}.

\textbf{Action Anticipation with Variable Time Interval.} The existing problem setting of short-term and long-term action anticipation requires the model forecast the future action $\tau$ seconds after the last observed video frame. Existing work typically compares with baselines under different values of $\tau$. However, they usually have to train a new model corresponding to different anticipation time. How to input $\tau$ as a variable into the model is an important research direction. Ke et al. \cite{ke2019time} propose the only method that forecasts future actions conditioned on a given $\tau$ by using a time-conditioned skip connection. There still exists large space for improvement by exploiting better time embedding and fusion strategies.

\textbf{Multi-granularity Action Anticipation.} The majority of previous models predicts the action classes of future actions. Nevertheless, a more fine-grained forecast is necessary in many cases. For example, the autonomous driving system has to know the size of pedestrians by predicting bounding boxes rather than the action labels alone. Other action anticipation of different granularities includes future hand masks \cite{jia2022generative}, gaze \cite{lai2024listen,lai2022eye,lai2024eye}, hand-object interactions \cite{lai2024lego,zhang2024hoidiffusion,ye2023affordance}, etc. Another research line is generating descriptions of future actions in natural language, so that the model is not limited to a pre-defined vocabulary of actions. Though there exist some progress in this direction \cite{sener2022transferring,sener2019zero,abdelsalam2023gepsan}, more attention should be paid to this area, especially considering the fast development of multi-modal large language models.

\section{Conclusion}\label{sec:conclusion}

In this survey, we categorize the action anticipation problem into seven fine-grained tasks and  make a thorough review of all the important papers in each task. We make a comparison across the prior methods in the viewpoint, duration of observed videos, anticipation time, modalities, etc. For each task, we group the primary contribution of the existing work into three groups: inductive biases, multi-modal models, and pretraining or objectives. The relation and difference of these methods are also summarized in this paper. We also list all datasets used for action anticipation and specify some important statistics. For a better understanding of the current model performance, we summarize the metrics and quantitative results of prior models on widely-used datasets in the supplementary. More importantly, we provide our outlook into the future promising research directions based on the unsolved challenges, defects of datasets and recent technical advance. We believe this survey can provide necessary assistance to researchers working on action anticipation problem and inspire more future studies in the community.

\ifCLASSOPTIONcaptionsoff
  \newpage
\fi

\bibliographystyle{IEEEtran}
\bibliography{references}

\begin{thebibliography}{100}
\providecommand{\url}[1]{#1}
\csname url@samestyle\endcsname
\providecommand{\newblock}{\relax}
\providecommand{\bibinfo}[2]{#2}
\providecommand{\BIBentrySTDinterwordspacing}{\spaceskip=0pt\relax}
\providecommand{\BIBentryALTinterwordstretchfactor}{4}
\providecommand{\BIBentryALTinterwordspacing}{\spaceskip=\fontdimen2\font plus
\BIBentryALTinterwordstretchfactor\fontdimen3\font minus \fontdimen4\font\relax}
\providecommand{\BIBforeignlanguage}[2]{{%
\expandafter\ifx\csname l@#1\endcsname\relax
\typeout{** WARNING: IEEEtran.bst: No hyphenation pattern has been}%
\typeout{** loaded for the language `#1'. Using the pattern for}%
\typeout{** the default language instead.}%
\else
\language=\csname l@#1\endcsname
\fi
#2}}
\providecommand{\BIBdecl}{\relax}
\BIBdecl

\bibitem{pei2011parsing}
M.~Pei, Y.~Jia, and S.-C. Zhu, ``Parsing video events with goal inference and intent prediction,'' in \emph{ICCV}, 2011, pp. 487--494.

\bibitem{lan2014hierarchical}
T.~Lan, T.-C. Chen, and S.~Savarese, ``A hierarchical representation for future action prediction,'' in \emph{ECCV}, 2014, pp. 689--704.

\bibitem{han2017human}
T.~Han, J.~Wang, A.~Cherian, and S.~Gould, ``Human action forecasting by learning task grammars,'' \emph{arXiv preprint arXiv:1709.06391}, 2017.

\bibitem{abu2018will}
Y.~Abu~Farha, A.~Richard, and J.~Gall, ``When will you do what?-anticipating temporal occurrences of activities,'' in \emph{CVPR}, 2018, pp. 5343--5352.

\bibitem{gong2022future}
D.~Gong, J.~Lee, M.~Kim, S.~J. Ha, and M.~Cho, ``Future transformer for long-term action anticipation,'' in \emph{CVPR}, 2022, pp. 3052--3061.

\bibitem{nawhal2022rethinking}
M.~Nawhal, A.~A. Jyothi, and G.~Mori, ``Rethinking learning approaches for long-term action anticipation,'' in \emph{ECCV}, 2022, pp. 558--576.

\bibitem{sener2019zero}
F.~Sener and A.~Yao, ``Zero-shot anticipation for instructional activities,'' in \emph{ICCV}, 2019, pp. 862--871.

\bibitem{abdelsalam2023gepsan}
M.~A. Abdelsalam, S.~B. Rangrej, I.~Hadji, N.~Dvornik, K.~G. Derpanis, and A.~Fazly, ``Gepsan: Generative procedure step anticipation in cooking videos,'' in \emph{ICCV}, 2023, pp. 2988--2997.

\bibitem{damen2018scaling}
D.~Damen, H.~Doughty, G.~M. Farinella, S.~Fidler, A.~Furnari, E.~Kazakos, D.~Moltisanti, J.~Munro, T.~Perrett, W.~Price \emph{et~al.}, ``Scaling egocentric vision: The {EPIC-KITCHENS} dataset,'' in \emph{ECCV}, 2018, pp. 720--736.

\bibitem{rasouli2020pedestrian}
A.~Rasouli, I.~Kotseruba, and J.~K. Tsotsos, ``Pedestrian action anticipation using contextual feature fusion in stacked rnns,'' \emph{BMVC}, 2020.

\bibitem{liu2020spatiotemporal}
B.~Liu, E.~Adeli, Z.~Cao, K.-H. Lee, A.~Shenoi, A.~Gaidon, and J.~C. Niebles, ``Spatiotemporal relationship reasoning for pedestrian intent prediction,'' \emph{IEEE Robotics and Automation Letters}, vol.~5, no.~2, pp. 3485--3492, 2020.

\bibitem{girase2021loki}
H.~Girase, H.~Gang, S.~Malla, J.~Li, A.~Kanehara, K.~Mangalam, and C.~Choi, ``{LOKI}: Long term and key intentions for trajectory prediction,'' in \emph{ICCV}, 2021, pp. 9803--9812.

\bibitem{barquero2022didn}
G.~Barquero, J.~N{\'u}nez, S.~Escalera, Z.~Xu, W.-W. Tu, I.~Guyon, and C.~Palmero, ``Didn’t see that coming: a survey on non-verbal social human behavior forecasting,'' in \emph{Understanding Social Behavior in Dyadic and Small Group Interactions}, 2022, pp. 139--178.

\bibitem{damen2019ekchallenges}
\BIBentryALTinterwordspacing
D.~Damen, W.~Price, E.~Kazakos, A.~Furnari, and G.~M. Farinella, ``{EPIC-KITCHENS} - 2019 challenges report,'' 2019. [Online]. Available: \url{https://epic-kitchens.github.io/Reports/EPIC-Kitchens-Challenges-2019-Report.pdf}
\BIBentrySTDinterwordspacing

\bibitem{damen2020ekchallenges}
\BIBentryALTinterwordspacing
D.~Damen, E.~Kazakos, W.~Price, J.~Ma, and H.~Doughty, ``{EPIC-KITCHENS-55} - 2020 challenges report,'' 2020. [Online]. Available: \url{https://epic-kitchens.github.io/Reports/EPIC-KITCHENS-Challenges-2020-Report.pdf}
\BIBentrySTDinterwordspacing

\bibitem{damen2021ekchallenges}
\BIBentryALTinterwordspacing
D.~Damen, A.~Fragomeni, J.~Munro, T.~Perrett, D.~Whettam, and M.~Wray, ``{EPIC-KITCHENS-100} - 2021 challenges report,'' 2021. [Online]. Available: \url{https://epic-kitchens.github.io/Reports/EPIC-KITCHENS-Challenges-2021-Report.pdf}
\BIBentrySTDinterwordspacing

\bibitem{damen2022ekchallenges}
\BIBentryALTinterwordspacing
D.~Damen, A.~Fragomeni, T.~Perrett, D.~Whettam, M.~Wray, and B.~Zhu, ``{EPIC-KITCHENS-100} - 2022 challenges report,'' 2022. [Online]. Available: \url{https://epic-kitchens.github.io/Reports/EPIC-KITCHENS-Challenges-2022-Report.pdf}
\BIBentrySTDinterwordspacing

\bibitem{grauman2022ego4dcvpr}
\BIBentryALTinterwordspacing
K.~Grauman, J.~Malik, M.~Z. Shou, R.~Girdhar, G.~M. Farinella, J.~Rehg, A.~Westbury, D.~Damen, and H.~S. Park, ``1st {Ego4D} workshop @ {CVPR} 2022,'' 2022. [Online]. Available: \url{https://ego4d-data.org/workshops/cvpr22/}
\BIBentrySTDinterwordspacing

\bibitem{girdhar2022ego4deccv}
\BIBentryALTinterwordspacing
R.~Girdhar, A.~Westbury, M.~Wray, A.~Furnari, S.~Bansal, D.~Kukreja, K.~Grauman, J.~Malik, D.~Damen, G.~M. Farinella, J.~Rehg, D.~Crandall, H.~S. Park, M.~Z. Shou, C.~Jawahar, K.~Kitani, B.~Ghanem, J.~Shi, Y.~Sato, P.~Arbelaez, A.~Oliva, and A.~Torralba, ``2nd international {Ego4D} workshop @ {ECCV} 2022,'' 2022. [Online]. Available: \url{https://ego4d-data.org/workshops/eccv22/}
\BIBentrySTDinterwordspacing

\bibitem{furnari2019would}
A.~Furnari and G.~M. Farinella, ``What would you expect? anticipating egocentric actions with rolling-unrolling {LSTM}s and modality attention,'' in \emph{ICCV}, 2019, pp. 6252--6261.

\bibitem{furnari2020rolling}
A.~Furnari and G.~Farinella, ``Rolling-unrolling {LSTM}s for action anticipation from first-person video,'' \emph{TPAMI}, vol.~43, no.~11, pp. 4021--4036, 2020.

\bibitem{zhong2023anticipative}
Z.~Zhong, D.~Schneider, M.~Voit, R.~Stiefelhagen, and J.~Beyerer, ``Anticipative feature fusion transformer for multi-modal action anticipation,'' in \emph{WACV}, 2023, pp. 6068--6077.

\bibitem{mittal2024can}
H.~Mittal, N.~Agarwal, S.-Y. Lo, and K.~Lee, ``Can't make an omelette without breaking some eggs: Plausible action anticipation using large video-language models,'' in \emph{CVPR}, 2024, pp. 18\,580--18\,590.

\bibitem{stein2013combining}
S.~Stein and S.~J. McKenna, ``Combining embedded accelerometers with computer vision for recognizing food preparation activities,'' in \emph{ACM international joint conference on Pervasive and ubiquitous computing}, 2013, pp. 729--738.

\bibitem{kuehne2014language}
H.~Kuehne, A.~Arslan, and T.~Serre, ``The language of actions: Recovering the syntax and semantics of goal-directed human activities,'' in \emph{CVPR}, 2014, pp. 780--787.

\bibitem{THUMOS14}
Y.-G. Jiang, J.~Liu, A.~Roshan~Zamir, G.~Toderici, I.~Laptev, M.~Shah, and R.~Sukthankar, ``{THUMOS} challenge: Action recognition with a large number of classes,'' \url{http://crcv.ucf.edu/THUMOS14/}, 2014.

\bibitem{jain2016recurrent}
A.~Jain, A.~Singh, H.~S. Koppula, S.~Soh, and A.~Saxena, ``Recurrent neural networks for driver activity anticipation via sensory-fusion architecture,'' in \emph{ICRA}, 2016, pp. 3118--3125.

\bibitem{yeung2018every}
S.~Yeung, O.~Russakovsky, N.~Jin, M.~Andriluka, G.~Mori, and L.~Fei-Fei, ``Every moment counts: Dense detailed labeling of actions in complex videos,'' \emph{IJCV}, vol. 126, pp. 375--389, 2018.

\bibitem{zhao2024antgpt}
Q.~Zhao, C.~Zhang, S.~Wang, C.~Fu, N.~Agarwal, K.~Lee, and C.~Sun, ``Antgpt: Can large language models help long-term action anticipation from videos?'' \emph{ICLR}, 2024.

\bibitem{rodin2021predicting}
I.~Rodin, A.~Furnari, D.~Mavroeidis, and G.~M. Farinella, ``Predicting the future from first person (egocentric) vision: A survey,'' \emph{Computer Vision and Image Understanding}, vol. 211, p. 103252, 2021.

\bibitem{zhong2023survey}
Z.~Zhong, M.~Martin, M.~Voit, J.~Gall, and J.~Beyerer, ``A survey on deep learning techniques for action anticipation,'' \emph{arXiv preprint arXiv:2309.17257}, 2023.

\bibitem{kong2022human}
Y.~Kong and Y.~Fu, ``Human action recognition and prediction: A survey,'' \emph{IJCV}, vol. 130, no.~5, pp. 1366--1401, 2022.

\bibitem{rasouli2020deep}
A.~Rasouli, ``Deep learning for vision-based prediction: A survey,'' \emph{arXiv preprint arXiv:2007.00095}, 2020.

\bibitem{herath2017going}
S.~Herath, M.~Harandi, and F.~Porikli, ``Going deeper into action recognition: A survey,'' \emph{Image and vision computing}, vol.~60, pp. 4--21, 2017.

\bibitem{poppe2010survey}
R.~Poppe, ``A survey on vision-based human action recognition,'' \emph{Image and vision computing}, vol.~28, no.~6, pp. 976--990, 2010.

\bibitem{sun2022human}
Z.~Sun, Q.~Ke, H.~Rahmani, M.~Bennamoun, G.~Wang, and J.~Liu, ``Human action recognition from various data modalities: A review,'' \emph{TPAMI}, 2022.

\bibitem{rudenko2020human}
A.~Rudenko, L.~Palmieri, M.~Herman, K.~M. Kitani, D.~M. Gavrila, and K.~O. Arras, ``Human motion trajectory prediction: A survey,'' \emph{The International Journal of Robotics Research}, vol.~39, no.~8, pp. 895--935, 2020.

\bibitem{lyu20223d}
K.~Lyu, H.~Chen, Z.~Liu, B.~Zhang, and R.~Wang, ``{3D} human motion prediction: A survey,'' \emph{Neurocomputing}, vol. 489, pp. 345--365, 2022.

\bibitem{rasouli2019pie}
A.~Rasouli, I.~Kotseruba, T.~Kunic, and J.~K. Tsotsos, ``{PIE}: A large-scale dataset and models for pedestrian intention estimation and trajectory prediction,'' in \emph{ICCV}, 2019.

\bibitem{mahmud2016poisson}
T.~Mahmud, M.~Hasan, A.~Chakraborty, and A.~K. Roy-Chowdhury, ``A poisson process model for activity forecasting,'' in \emph{ICIP}, 2016, pp. 3339--3343.

\bibitem{zeng2017visual}
K.-H. Zeng, W.~B. Shen, D.-A. Huang, M.~Sun, and J.~Carlos~Niebles, ``Visual forecasting by imitating dynamics in natural sequences,'' in \emph{ICCV}, 2017, pp. 2999--3008.

\bibitem{mahmud2017joint}
T.~Mahmud, M.~Hasan, and A.~K. Roy-Chowdhury, ``Joint prediction of activity labels and starting times in untrimmed videos,'' in \emph{ICCV}, 2017, pp. 5773--5782.

\bibitem{shen2018egocentric}
Y.~Shen, B.~Ni, Z.~Li, and N.~Zhuang, ``Egocentric activity prediction via event modulated attention,'' in \emph{ECCV}, 2018, pp. 197--212.

\bibitem{rotondo2019action}
T.~Rotondo, G.~M. Farinella, V.~Tomaselli, and S.~Battiato, ``Action anticipation from multimodal data.'' in \emph{VISIGRAPP (4: VISAPP)}, 2019, pp. 154--161.

\bibitem{neumann2019future}
L.~Neumann, A.~Zisserman, and A.~Vedaldi, ``Future event prediction: If and when,'' in \emph{CVPR Workshops}, 2019.

\bibitem{ke2021future}
Q.~Ke, M.~Fritz, and B.~Schiele, ``Future moment assessment for action query,'' in \emph{WACV}, 2021, pp. 3219--3228.

\bibitem{roy2022action}
D.~Roy and B.~Fernando, ``Action anticipation using latent goal learning,'' in \emph{WACV}, 2022, pp. 2745--2753.

\bibitem{zhong2023learning}
Y.~Zhong, L.~Yu, Y.~Bai, S.~Li, X.~Yan, and Y.~Li, ``Learning procedure-aware video representation from instructional videos and their narrations,'' in \emph{CVPR}, 2023, pp. 14\,825--14\,835.

\bibitem{thakur2024leveraging}
S.~Thakur, C.~Beyan, P.~Morerio, V.~Murino, and A.~Del~Bue, ``Leveraging next-active objects for context-aware anticipation in egocentric videos,'' in \emph{WACV}, 2024, pp. 8657--8666.

\bibitem{wu2017anticipating}
T.-Y. Wu, T.-A. Chien, C.-S. Chan, C.-W. Hu, and M.~Sun, ``Anticipating daily intention using on-wrist motion triggered sensing,'' in \emph{ICCV}, 2017, pp. 48--56.

\bibitem{epstein2021learning}
D.~Epstein and C.~Vondrick, ``Learning goals from failure,'' in \emph{CVPR}, 2021, pp. 11\,194--11\,204.

\bibitem{suris2021learning}
D.~Sur{\'\i}s, R.~Liu, and C.~Vondrick, ``Learning the predictability of the future,'' in \emph{CVPR}, 2021, pp. 12\,607--12\,617.

\bibitem{koppula2015anticipating}
H.~S. Koppula and A.~Saxena, ``Anticipating human activities using object affordances for reactive robotic response,'' \emph{TPAMI}, vol.~38, no.~1, pp. 14--29, 2015.

\bibitem{vondrick2016anticipating}
C.~Vondrick, H.~Pirsiavash, and A.~Torralba, ``Anticipating visual representations from unlabeled video,'' in \emph{CVPR}, 2016, pp. 98--106.

\bibitem{furnari2018leveraging}
A.~Furnari, S.~Battiato, and G.~Maria~Farinella, ``Leveraging uncertainty to rethink loss functions and evaluation measures for egocentric action anticipation,'' in \emph{ECCV Workshops}, 2018.

\bibitem{miech2019leveraging}
A.~Miech, I.~Laptev, J.~Sivic, H.~Wang, L.~Torresani, and D.~Tran, ``Leveraging the present to anticipate the future in videos,'' in \emph{CVPR Workshops}, 2019.

\bibitem{ke2019time}
Q.~Ke, M.~Fritz, and B.~Schiele, ``Time-conditioned action anticipation in one shot,'' in \emph{CVPR}, 2019, pp. 9925--9934.

\bibitem{liu2020forecasting}
M.~Liu, S.~Tang, Y.~Li, and J.~M. Rehg, ``Forecasting human-object interaction: joint prediction of motor attention and actions in first person video,'' in \emph{ECCV}, 2020, pp. 704--721.

\bibitem{zhang2020egocentric}
T.~Zhang, W.~Min, Y.~Zhu, Y.~Rui, and S.~Jiang, ``An egocentric action anticipation framework via fusing intuition and analysis,'' in \emph{ACM MM}, 2020, pp. 402--410.

\bibitem{guan2020generative}
J.~Guan, Y.~Yuan, K.~M. Kitani, and N.~Rhinehart, ``Generative hybrid representations for activity forecasting with no-regret learning,'' in \emph{CVPR}, 2020, pp. 173--182.

\bibitem{nagarajan2020ego}
T.~Nagarajan, Y.~Li, C.~Feichtenhofer, and K.~Grauman, ``{E}go-{T}opo: Environment affordances from egocentric video,'' in \emph{CVPR}, 2020, pp. 163--172.

\bibitem{wu2020learning}
Y.~Wu, L.~Zhu, X.~Wang, Y.~Yang, and F.~Wu, ``Learning to anticipate egocentric actions by imagination,'' \emph{IEEE Transactions on Image Processing}, vol.~30, pp. 1143--1152, 2020.

\bibitem{fernando2021anticipating}
B.~Fernando and S.~Herath, ``Anticipating human actions by correlating past with the future with jaccard similarity measures,'' in \emph{CVPR}, 2021, pp. 13\,224--13\,233.

\bibitem{osman2021slowfast}
N.~Osman, G.~Camporese, P.~Coscia, and L.~Ballan, ``Slowfast rolling-unrolling lstms for action anticipation in egocentric videos,'' in \emph{ICCV}, 2021, pp. 3437--3445.

\bibitem{zatsarynna2021multi}
O.~Zatsarynna, Y.~Abu~Farha, and J.~Gall, ``Multi-modal temporal convolutional network for anticipating actions in egocentric videos,'' in \emph{CVPR}, 2021, pp. 2249--2258.

\bibitem{dessalene2020egocentric}
E.~Dessalene, M.~Maynord, C.~Devaraj, C.~Fermuller, and Y.~Aloimonos, ``Egocentric object manipulation graphs,'' \emph{arXiv preprint arXiv:2006.03201}, 2020.

\bibitem{dessalene2021forecasting}
E.~Dessalene, C.~Devaraj, M.~Maynord, C.~Ferm{\"u}ller, and Y.~Aloimonos, ``Forecasting action through contact representations from first person video,'' \emph{TPAMI}, vol.~45, no.~6, pp. 6703--6714, 2021.

\bibitem{xu2021long}
M.~Xu, Y.~Xiong, H.~Chen, X.~Li, W.~Xia, Z.~Tu, and S.~Soatto, ``Long short-term transformer for online action detection,'' \emph{NeurIPS}, vol.~34, pp. 1086--1099, 2021.

\bibitem{girdhar2021anticipative}
R.~Girdhar and K.~Grauman, ``Anticipative video transformer,'' in \emph{ICCV}, 2021, pp. 13\,505--13\,515.

\bibitem{xu2022learning}
X.~Xu, Y.-L. Li, and C.~Lu, ``Learning to anticipate future with dynamic context removal,'' in \emph{CVPR}, 2022, pp. 12\,734--12\,744.

\bibitem{wu2022memvit}
C.-Y. Wu, Y.~Li, K.~Mangalam, H.~Fan, B.~Xiong, J.~Malik, and C.~Feichtenhofer, ``{MeMViT}: Memory-augmented multiscale vision transformer for efficient long-term video recognition,'' in \emph{CVPR}, 2022, pp. 13\,587--13\,597.

\bibitem{manousaki2023vlmah}
V.~Manousaki, K.~Bacharidis, K.~Papoutsakis, and A.~Argyros, ``Vlmah: Visual-linguistic modeling of action history for effective action anticipation,'' in \emph{ICCV}, 2023, pp. 1917--1927.

\bibitem{wang2023memory}
J.~Wang, G.~Chen, Y.~Huang, L.~Wang, and T.~Lu, ``Memory-and-anticipation transformer for online action understanding,'' in \emph{ICCV}, 2023, pp. 13\,824--13\,835.

\bibitem{girase2023latency}
H.~Girase, N.~Agarwal, C.~Choi, and K.~Mangalam, ``Latency matters: Real-time action forecasting transformer,'' in \emph{CVPR}, 2023, pp. 18\,759--18\,769.

\bibitem{guermal2024joadaa}
M.~Guermal, A.~Ali, R.~Dai, and F.~Br{\'e}mond, ``Joadaa: joint online action detection and action anticipation,'' in \emph{WACV}, 2024, pp. 6889--6898.

\bibitem{roy2024interaction}
D.~Roy, R.~Rajendiran, and B.~Fernando, ``Interaction region visual transformer for egocentric action anticipation,'' in \emph{WACV}, 2024, pp. 6740--6750.

\bibitem{guo2024uncertainty}
H.~Guo, N.~Agarwal, S.-Y. Lo, K.~Lee, and Q.~Ji, ``Uncertainty-aware action decoupling transformer for action anticipation,'' in \emph{CVPR}, 2024, pp. 18\,644--18\,654.

\bibitem{diko2024semantically}
A.~Diko, D.~Avola, B.~Prenkaj, F.~Fontana, and L.~Cinque, ``Semantically guided representation learning for action anticipation,'' \emph{arXiv preprint arXiv:2407.02309}, 2024.

\bibitem{chakraborty2015context}
A.~Chakraborty and A.~K. Roy-Chowdhury, ``Context-aware activity forecasting,'' in \emph{ACCV}, 2015, pp. 21--36.

\bibitem{rhinehart2017first}
N.~Rhinehart and K.~M. Kitani, ``First-person activity forecasting with online inverse reinforcement learning,'' in \emph{ICCV}, 2017, pp. 3696--3705.

\bibitem{gao2017red}
J.~Gao, Z.~Yang, and R.~Nevatia, ``Red: Reinforced encoder-decoder networks for action anticipation,'' \emph{BMVC}, 2017.

\bibitem{sun2019relational}
C.~Sun, A.~Shrivastava, C.~Vondrick, R.~Sukthankar, K.~Murphy, and C.~Schmid, ``Relational action forecasting,'' in \emph{CVPR}, 2019, pp. 273--283.

\bibitem{piergiovanni2020adversarial}
A.~Piergiovanni, A.~Angelova, A.~Toshev, and M.~S. Ryoo, ``Adversarial generative grammars for human activity prediction,'' in \emph{ECCV}, 2020, pp. 507--523.

\bibitem{li2021restep}
Y.~Li, P.~Wang, and C.-Y. Chan, ``Restep into the future: relational spatio-temporal learning for multi-person action forecasting,'' \emph{IEEE Transactions on Multimedia}, vol.~25, pp. 1954--1963, 2021.

\bibitem{qi2021self}
Z.~Qi, S.~Wang, C.~Su, L.~Su, Q.~Huang, and Q.~Tian, ``Self-regulated learning for egocentric video activity anticipation,'' \emph{TPAMI}, 2021.

\bibitem{liu2022hybrid}
T.~Liu and K.-M. Lam, ``A hybrid egocentric activity anticipation framework via memory-augmented recurrent and one-shot representation forecasting,'' in \emph{CVPR}, 2022, pp. 13\,904--13\,913.

\bibitem{zhao2022real}
Y.~Zhao and P.~Kr{\"a}henb{\"u}hl, ``Real-time online video detection with temporal smoothing transformers,'' in \emph{ECCV}, 2022, pp. 485--502.

\bibitem{qi2017predicting}
S.~Qi, S.~Huang, P.~Wei, and S.-C. Zhu, ``Predicting human activities using stochastic grammar,'' in \emph{ICCV}, 2017, pp. 1164--1172.

\bibitem{schydlo2018anticipation}
P.~Schydlo, M.~Rakovic, L.~Jamone, and J.~Santos-Victor, ``Anticipation in human-robot cooperation: A recurrent neural network approach for multiple action sequences prediction,'' in \emph{ICRA}, 2018, pp. 5909--5914.

\bibitem{gammulle2019forecasting}
H.~Gammulle, S.~Denman, S.~Sridharan, and C.~Fookes, ``Forecasting future action sequences with neural memory networks,'' \emph{BMVC}, 2019.

\bibitem{sener2022transferring}
F.~Sener, R.~Saraf, and A.~Yao, ``Transferring knowledge from text to video: Zero-shot anticipation for procedural actions,'' \emph{TPAMI}, 2022.

\bibitem{ng2020forecasting}
Y.~B. Ng and B.~Fernando, ``Forecasting future action sequences with attention: a new approach to weakly supervised action forecasting,'' \emph{IEEE Transactions on Image Processing}, vol.~29, pp. 8880--8891, 2020.

\bibitem{sener2020temporal}
F.~Sener, D.~Singhania, and A.~Yao, ``Temporal aggregate representations for long-range video understanding,'' in \emph{ECCV}, 2020, pp. 154--171.

\bibitem{sener2021technical}
F.~Sener, D.~Chatterjee, and A.~Yao, ``Technical report: Temporal aggregate representations,'' \emph{arXiv preprint arXiv:2106.03152}, 2021.

\bibitem{roy2021action}
D.~Roy and B.~Fernando, ``Action anticipation using pairwise human-object interactions and transformers,'' \emph{IEEE Transactions on Image Processing}, vol.~30, pp. 8116--8129, 2021.

\bibitem{mascaro2023intention}
E.~V. Mascar{\'o}, H.~Ahn, and D.~Lee, ``Intention-conditioned long-term human egocentric action anticipation,'' in \emph{WACV}, 2023, pp. 6048--6057.

\bibitem{ashutosh2023hiervl}
K.~Ashutosh, R.~Girdhar, L.~Torresani, and K.~Grauman, ``Hiervl: Learning hierarchical video-language embeddings,'' in \emph{CVPR}, 2023, pp. 23\,066--23\,078.

\bibitem{zhang2024object}
C.~Zhang, C.~Fu, S.~Wang, N.~Agarwal, K.~Lee, C.~Choi, and C.~Sun, ``Object-centric video representation for long-term action anticipation,'' in \emph{WACV}, 2024, pp. 6751--6761.

\bibitem{abu2019uncertainty}
Y.~Abu~Farha and J.~Gall, ``Uncertainty-aware anticipation of activities,'' in \emph{ICCV Workshops}, 2019.

\bibitem{mehrasa2019variational}
N.~Mehrasa, A.~A. Jyothi, T.~Durand, J.~He, L.~Sigal, and G.~Mori, ``A variational auto-encoder model for stochastic point processes,'' in \emph{CVPR}, 2019, pp. 3165--3174.

\bibitem{zhao2020diverse}
H.~Zhao and R.~P. Wildes, ``On diverse asynchronous activity anticipation,'' in \emph{ECCV}, 2020, pp. 781--799.

\bibitem{loh2022long}
S.~B. Loh, D.~Roy, and B.~Fernando, ``Long-term action forecasting using multi-headed attention-based variational recurrent neural networks,'' in \emph{CVPR}, 2022, pp. 2419--2427.

\bibitem{zhong2023diffant}
Z.~Zhong, C.~Wu, M.~Martin, M.~Voit, J.~Gall, and J.~Beyerer, ``Diffant: Diffusion models for action anticipation,'' \emph{arXiv preprint arXiv:2311.15991}, 2023.

\bibitem{chang2020procedure}
C.-Y. Chang, D.-A. Huang, D.~Xu, E.~Adeli, L.~Fei-Fei, and J.~C. Niebles, ``Procedure planning in instructional videos,'' in \emph{ECCV}, 2020, pp. 334--350.

\bibitem{bi2021procedure}
J.~Bi, J.~Luo, and C.~Xu, ``Procedure planning in instructional videos via contextual modeling and model-based policy learning,'' in \emph{ICCV}, 2021, pp. 15\,611--15\,620.

\bibitem{sun2022plate}
J.~Sun, D.-A. Huang, B.~Lu, Y.-H. Liu, B.~Zhou, and A.~Garg, ``{PlaTe}: Visually-grounded planning with transformers in procedural tasks,'' \emph{IEEE Robotics and Automation Letters}, vol.~7, no.~2, pp. 4924--4930, 2022.

\bibitem{zhao2022p3iv}
H.~Zhao, I.~Hadji, N.~Dvornik, K.~G. Derpanis, R.~P. Wildes, and A.~D. Jepson, ``{P3IV}: Probabilistic procedure planning from instructional videos with weak supervision,'' in \emph{CVPR}, 2022, pp. 2938--2948.

\bibitem{wang2023pdpp}
H.~Wang, Y.~Wu, S.~Guo, and L.~Wang, ``Pdpp: Projected diffusion for procedure planning in instructional videos,'' in \emph{CVPR}, 2023, pp. 14\,836--14\,845.

\bibitem{bregler1997learning}
C.~Bregler, ``Learning and recognizing human dynamics in video sequences,'' in \emph{CVPR}, 1997, pp. 568--574.

\bibitem{hochreiter1997long}
S.~Hochreiter and J.~Schmidhuber, ``Long short-term memory,'' \emph{Neural computation}, vol.~9, no.~8, pp. 1735--1780, 1997.

\bibitem{roy2022predicting}
D.~Roy and B.~Fernando, ``Predicting the next action by modeling the abstract goal,'' \emph{arXiv preprint arXiv:2209.05044}, 2022.

\bibitem{han2019video}
T.~Han, W.~Xie, and A.~Zisserman, ``Video representation learning by dense predictive coding,'' in \emph{ICCV Workshops}, 2019, pp. 0--0.

\bibitem{epstein2020oops}
D.~Epstein, B.~Chen, and C.~Vondrick, ``Oops! predicting unintentional action in video,'' in \emph{CVPR}, 2020, pp. 919--929.

\bibitem{damen2020rescaling}
D.~Damen, H.~Doughty, G.~M. Farinella, A.~Furnari, E.~Kazakos, J.~Ma, D.~Moltisanti, J.~Munro, T.~Perrett, W.~Price \emph{et~al.}, ``Rescaling egocentric vision,'' \emph{arXiv preprint arXiv:2006.13256}, 2020.

\bibitem{gu2021transaction}
X.~Gu, J.~Qiu, Y.~Guo, B.~Lo, and G.-Z. Yang, ``{T}rans{A}ction: {ICL-SJTU} submission to {EPIC}-{K}itchens action anticipation challenge 2021,'' \emph{arXiv preprint arXiv:2107.13259}, 2021.

\bibitem{wang2021oadtr}
X.~Wang, S.~Zhang, Z.~Qing, Y.~Shao, Z.~Zuo, C.~Gao, and N.~Sang, ``Oadtr: Online action detection with transformers,'' in \emph{ICCV}, 2021, pp. 7565--7575.

\bibitem{zatsarynna2023action}
O.~Zatsarynna and J.~Gall, ``Action anticipation with goal consistency,'' in \emph{ICIP}, 2023, pp. 1630--1634.

\bibitem{tai2022unified}
T.-M. Tai, G.~Fiameni, C.-K. Lee, S.~See, and O.~Lanz, ``Unified recurrence modeling for video action anticipation,'' in \emph{ICPR}, 2022, pp. 3273--3279.

\bibitem{thakur2023enhancing}
S.~Thakur, C.~Beyan, P.~Morerio, V.~Murino, and A.~Del~Bue, ``Enhancing next active object-based egocentric action anticipation with guided attention,'' in \emph{ICIP}, 2023, pp. 1450--1454.

\bibitem{dosovitskiy2020image}
A.~Dosovitskiy, L.~Beyer, A.~Kolesnikov, D.~Weissenborn, X.~Zhai, T.~Unterthiner, M.~Dehghani, M.~Minderer, G.~Heigold, S.~Gelly \emph{et~al.}, ``An image is worth 16x16 words: Transformers for image recognition at scale,'' in \emph{ICLR}, 2020.

\bibitem{ren2015faster}
S.~Ren, K.~He, R.~Girshick, and J.~Sun, ``Faster r-cnn: Towards real-time object detection with region proposal networks,'' \emph{NeurIPS}, vol.~28, 2015.

\bibitem{xu2019temporal}
M.~Xu, M.~Gao, Y.-T. Chen, L.~S. Davis, and D.~J. Crandall, ``Temporal recurrent networks for online action detection,'' in \emph{ICCV}, 2019, pp. 5532--5541.

\bibitem{qu2020lap}
S.~Qu, G.~Chen, D.~Xu, J.~Dong, F.~Lu, and A.~Knoll, ``Lap-net: Adaptive features sampling via learning action progression for online action detection,'' \emph{arXiv preprint arXiv:2011.07915}, 2020.

\bibitem{ghosh2023text}
S.~Ghosh, T.~Aggarwal, M.~Hoai, and N.~Balasubramanian, ``Text-derived knowledge helps vision: A simple cross-modal distillation for video-based action anticipation,'' in \emph{EACL Findings}, 2023, pp. 1882--1897.

\bibitem{gupta2022act}
A.~Gupta, J.~Liu, L.~Bo, A.~K. Roy-Chowdhury, and T.~Mei, ``{A-ACT}: Action anticipation through cycle transformations,'' \emph{arXiv preprint arXiv:2204.00942}, 2022.

\bibitem{zhong2018unsupervised}
Y.~Zhong and W.-S. Zheng, ``Unsupervised learning for forecasting action representations,'' in \emph{ICIP}, 2018, pp. 1073--1077.

\bibitem{camporese2021knowledge}
G.~Camporese, P.~Coscia, A.~Furnari, G.~M. Farinella, and L.~Ballan, ``Knowledge distillation for action anticipation via label smoothing,'' in \emph{ICPR}, 2021, pp. 3312--3319.

\bibitem{grauman2022ego4d}
K.~Grauman, A.~Westbury, E.~Byrne, Z.~Chavis, A.~Furnari, R.~Girdhar, J.~Hamburger, H.~Jiang, M.~Liu, X.~Liu \emph{et~al.}, ``{E}go{4D}: Around the world in 3,000 hours of egocentric video,'' in \emph{CVPR}, 2022, pp. 18\,995--19\,012.

\bibitem{bokhari2016long}
S.~Z. Bokhari and K.~M. Kitani, ``Long-term activity forecasting using first-person vision,'' in \emph{ACCV}, 2016, pp. 346--360.

\bibitem{das2022video+}
S.~Das and M.~S. Ryoo, ``Video+ clip baseline for ego4d long-term action anticipation,'' \emph{arXiv preprint arXiv:2207.00579}, 2022.

\bibitem{liu2022joint}
S.~Liu, S.~Tripathi, S.~Majumdar, and X.~Wang, ``Joint hand motion and interaction hotspots prediction from egocentric videos,'' in \emph{CVPR}, 2022, pp. 3282--3292.

\bibitem{he2017mask}
K.~He, G.~Gkioxari, P.~Doll{\'a}r, and R.~Girshick, ``Mask r-cnn,'' in \emph{ICCV}, 2017, pp. 2961--2969.

\bibitem{feichtenhofer2019slowfast}
C.~Feichtenhofer, H.~Fan, J.~Malik, and K.~He, ``Slowfast networks for video recognition,'' in \emph{ICCV}, 2019, pp. 6202--6211.

\bibitem{narasimhan2023learning}
M.~Narasimhan, L.~Yu, S.~Bell, N.~Zhang, and T.~Darrell, ``Learning and verification of task structure in instructional videos,'' \emph{arXiv preprint arXiv:2303.13519}, 2023.

\bibitem{xue2023egocentric}
Z.~Xue, Y.~Song, K.~Grauman, and L.~Torresani, ``Egocentric video task translation,'' in \emph{CVPR}, 2023, pp. 2310--2320.

\bibitem{tan2023multiscale}
R.~Tan, M.~De~Lange, M.~Iuzzolino, B.~A. Plummer, K.~Saenko, K.~Ridgeway, and L.~Torresani, ``Multiscale video pretraining for long-term activity forecasting,'' \emph{arXiv preprint arXiv:2307.12854}, 2023.

\bibitem{morais2020learning}
R.~Morais, V.~Le, T.~Tran, and S.~Venkatesh, ``Learning to abstract and predict human actions,'' \emph{BMVC}, 2020.

\bibitem{abu2021long}
Y.~Abu~Farha, Q.~Ke, B.~Schiele, and J.~Gall, ``Long-term anticipation of activities with cycle consistency,'' in \emph{DAGM German Conference on Pattern Recognition}, 2021, pp. 159--173.

\bibitem{mahmud2021prediction}
T.~Mahmud, M.~Billah, M.~Hasan, and A.~K. Roy-Chowdhury, ``Prediction and description of near-future activities in video,'' \emph{Computer Vision and Image Understanding}, vol. 210, p. 103230, 2021.

\bibitem{carion2020end}
N.~Carion, F.~Massa, G.~Synnaeve, N.~Usunier, A.~Kirillov, and S.~Zagoruyko, ``End-to-end object detection with transformers,'' in \emph{ECCV}, 2020, pp. 213--229.

\bibitem{yang2019diversity}
D.~Yang, S.~Hong, Y.~Jang, T.~Zhao, and H.~Lee, ``Diversity-sensitive conditional generative adversarial networks,'' in \emph{ICLR}, 2019.

\bibitem{chung2015recurrent}
J.~Chung, K.~Kastner, L.~Dinh, K.~Goel, A.~C. Courville, and Y.~Bengio, ``A recurrent latent variable model for sequential data,'' \emph{NeurIPS}, vol.~28, 2015.

\bibitem{kurutach2018learning}
T.~Kurutach, A.~Tamar, G.~Yang, S.~J. Russell, and P.~Abbeel, ``Learning plannable representations with causal {InfoGAN},'' \emph{NeurIPS}, vol.~31, 2018.

\bibitem{zhukov2019cross}
D.~Zhukov, J.-B. Alayrac, R.~G. Cinbis, D.~Fouhey, I.~Laptev, and J.~Sivic, ``Cross-task weakly supervised learning from instructional videos,'' in \emph{CVPR}, 2019, pp. 3537--3545.

\bibitem{tang2019coin}
Y.~Tang, D.~Ding, Y.~Rao, Y.~Zheng, D.~Zhang, L.~Zhao, J.~Lu, and J.~Zhou, ``{COIN}: A large-scale dataset for comprehensive instructional video analysis,'' in \emph{CVPR}, 2019, pp. 1207--1216.

\bibitem{alayrac2016unsupervised}
J.-B. Alayrac, P.~Bojanowski, N.~Agrawal, J.~Sivic, I.~Laptev, and S.~Lacoste-Julien, ``Unsupervised learning from narrated instruction videos,'' in \emph{CVPR}, 2016, pp. 4575--4583.

\bibitem{ho2016generative}
J.~Ho and S.~Ermon, ``Generative adversarial imitation learning,'' \emph{NeurIPS}, vol.~29, 2016.

\bibitem{patron2010high}
A.~Patron-Perez, M.~Marszalek, A.~Zisserman, and I.~Reid, ``High five: Recognising human interactions in {TV} shows,'' \emph{BMVC}, 2010.

\bibitem{patron2012structured}
A.~Patron-Perez, M.~Marszalek, I.~Reid, and A.~Zisserman, ``Structured learning of human interactions in {TV} shows,'' \emph{TPAMI}, vol.~34, no.~12, pp. 2441--2453, 2012.

\bibitem{oh2011large}
S.~Oh, A.~Hoogs, A.~Perera, N.~Cuntoor, C.-C. Chen, J.~T. Lee, S.~Mukherjee, J.~Aggarwal, H.~Lee, L.~Davis \emph{et~al.}, ``A large-scale benchmark dataset for event recognition in surveillance video,'' in \emph{CVPR}, 2011, pp. 3153--3160.

\bibitem{fathi2011learning}
A.~Fathi, X.~Ren, and J.~Rehg, ``Learning to recognize objects in egocentric activities,'' in \emph{CVPR}, 2011, pp. 3281--3288.

\bibitem{li2015delving}
Y.~Li, Z.~Ye, and J.~M. Rehg, ``Delving into egocentric actions,'' in \emph{CVPR}, 2015, pp. 287--295.

\bibitem{fathi2012learning}
A.~Fathi, Y.~Li, and J.~M. Rehg, ``Learning to recognize daily actions using gaze,'' in \emph{ECCV}, 2012, pp. 314--327.

\bibitem{rohrbach2012database}
M.~Rohrbach, S.~Amin, M.~Andriluka, and B.~Schiele, ``A database for fine grained activity detection of cooking activities,'' in \emph{CVPR}, 2012, pp. 1194--1201.

\bibitem{rohrbach2012script}
M.~Rohrbach, M.~Regneri, M.~Andriluka, S.~Amin, M.~Pinkal, and B.~Schiele, ``Script data for attribute-based recognition of composite activities,'' in \emph{ECCV}, 2012, pp. 144--157.

\bibitem{regneri2013grounding}
M.~Regneri, M.~Rohrbach, D.~Wetzel, S.~Thater, B.~Schiele, and M.~Pinkal, ``Grounding action descriptions in videos,'' \emph{ACL}, vol.~1, pp. 25--36, 2013.

\bibitem{rohrbach2014coherent}
A.~Rohrbach, M.~Rohrbach, W.~Qiu, A.~Friedrich, M.~Pinkal, and B.~Schiele, ``Coherent multi-sentence video description with variable level of detail,'' in \emph{Pattern Recognition: 36th German Conference}, 2014, pp. 184--195.

\bibitem{rohrbach2015recognizing}
\BIBentryALTinterwordspacing
M.~Rohrbach, A.~Rohrbach, M.~Regneri, S.~Amin, M.~Andriluka, M.~Pinkal, and B.~Schiele, ``Recognizing fine-grained and composite activities using hand-centric features and script data,'' \emph{IJCV}, pp. 1--28, 2015. [Online]. Available: \url{http://dx.doi.org/10.1007/s11263-015-0851-8}
\BIBentrySTDinterwordspacing

\bibitem{caba2015activitynet}
B.~G. Fabian Caba~Heilbron, Victor~Escorcia and J.~C. Niebles, ``{A}ctivity{N}et: A large-scale video benchmark for human activity understanding,'' in \emph{CVPR}, 2015, pp. 961--970.

\bibitem{de2016online}
R.~De~Geest, E.~Gavves, A.~Ghodrati, Z.~Li, C.~Snoek, and T.~Tuytelaars, ``Online action detection,'' in \emph{ECCV}, 2016, pp. 269--284.

\bibitem{sigurdsson2016hollywood}
G.~A. Sigurdsson, G.~Varol, X.~Wang, A.~Farhadi, I.~Laptev, and A.~Gupta, ``Hollywood in homes: Crowdsourcing data collection for activity understanding,'' in \emph{ECCV}, 2016, pp. 510--526.

\bibitem{li2018eye}
Y.~Li, M.~Liu, and J.~M. Rehg, ``In the eye of beholder: Joint learning of gaze and actions in first person video,'' in \emph{ECCV}, 2018, pp. 619--635.

\bibitem{gu2018ava}
C.~Gu, C.~Sun, D.~A. Ross, C.~Vondrick, C.~Pantofaru, Y.~Li, S.~Vijayanarasimhan, G.~Toderici, S.~Ricco, R.~Sukthankar \emph{et~al.}, ``{AVA}: A video dataset of spatio-temporally localized atomic visual actions,'' in \emph{CVPR}, 2018, pp. 6047--6056.

\bibitem{zhou2018towards}
L.~Zhou, C.~Xu, and J.~Corso, ``Towards automatic learning of procedures from web instructional videos,'' in \emph{AAAI}, vol.~32, no.~1, 2018.

\bibitem{sener2022assembly101}
F.~Sener, D.~Chatterjee, D.~Shelepov, K.~He, D.~Singhania, R.~Wang, and A.~Yao, ``Assembly101: A large-scale multi-view video dataset for understanding procedural activities,'' in \emph{CVPR}, 2022, pp. 21\,096--21\,106.

\bibitem{song2024ego4d}
Y.~Song, E.~Byrne, T.~Nagarajan, H.~Wang, M.~Martin, and L.~Torresani, ``Ego4d goal-step: Toward hierarchical understanding of procedural activities,'' \emph{NeurIPS}, vol.~36, 2024.

\bibitem{smaira2020short}
L.~Smaira, J.~Carreira, E.~Noland, E.~Clancy, A.~Wu, and A.~Zisserman, ``A short note on the {K}inetics-700-2020 human action dataset,'' \emph{arXiv preprint arXiv:2010.10864}, 2020.

\bibitem{kotseruba2016joint}
I.~Kotseruba, A.~Rasouli, and J.~K. Tsotsos, ``Joint attention in autonomous driving (jaad),'' \emph{arXiv preprint arXiv:1609.04741}, 2016.

\bibitem{salvador2017learning}
A.~Salvador, N.~Hynes, Y.~Aytar, J.~Marin, F.~Ofli, I.~Weber, and A.~Torralba, ``Learning cross-modal embeddings for cooking recipes and food images,'' in \emph{CVPR}, 2017.

\bibitem{marin2021recipe1m+}
J.~Mar{\i}n, A.~Biswas, F.~Ofli, N.~Hynes, A.~Salvador, Y.~Aytar, I.~Weber, and A.~Torralba, ``Recipe1m+: A dataset for learning cross-modal embeddings for cooking recipes and food images,'' \emph{TPAMI}, vol.~43, no.~1, pp. 187--203, 2021.

\bibitem{jain2016brain4cars}
A.~Jain, H.~S. Koppula, S.~Soh, B.~Raghavan, A.~Singh, and A.~Saxena, ``Brain4cars: Car that knows before you do via sensory-fusion deep learning architecture,'' \emph{arXiv preprint arXiv:1601.00740}, 2016.

\bibitem{ramanishka2018toward}
V.~Ramanishka, Y.-T. Chen, T.~Misu, and K.~Saenko, ``Toward driving scene understanding: A dataset for learning driver behavior and causal reasoning,'' in \emph{CVPR}, 2018, pp. 7699--7707.

\bibitem{THUMOS15}
A.~Gorban, H.~Idrees, Y.-G. Jiang, A.~Roshan~Zamir, I.~Laptev, M.~Shah, and R.~Sukthankar, ``{THUMOS} challenge: Action recognition with a large number of classes,'' \url{http://www.thumos.info/}, 2015.

\bibitem{yu2020bdd100k}
F.~Yu, H.~Chen, X.~Wang, W.~Xian, Y.~Chen, F.~Liu, V.~Madhavan, and T.~Darrell, ``{BDD}100k: A diverse driving dataset for heterogeneous multitask learning,'' in \emph{CVPR}, 2020, pp. 2636--2645.

\bibitem{nakamura2017jointly}
K.~Nakamura, S.~Yeung, A.~Alahi, and L.~Fei-Fei, ``Jointly learning energy expenditures and activities using egocentric multimodal signals,'' in \emph{CVPR}, 2017, pp. 1868--1877.

\bibitem{zadeh2019social}
A.~Zadeh, M.~Chan, P.~P. Liang, E.~Tong, and L.-P. Morency, ``Social-iq: A question answering benchmark for artificial social intelligence,'' in \emph{CVPR}, 2019, pp. 8807--8817.

\bibitem{lai2023werewolf}
B.~Lai, H.~Zhang, M.~Liu, A.~Pariani, F.~Ryan, W.~Jia, S.~A. Hayati, J.~Rehg, and D.~Yang, ``Werewolf among us: Multimodal resources for modeling persuasion behaviors in social deduction games,'' in \emph{ACL Findings}, 2023, pp. 6570--6588.

\bibitem{lee2024modeling}
S.~Lee, B.~Lai, F.~Ryan, B.~Boote, and J.~M. Rehg, ``Modeling multimodal social interactions: New challenges and baselines with densely aligned representations,'' in \emph{CVPR}, 2024.

\bibitem{caron2021emerging}
M.~Caron, H.~Touvron, I.~Misra, H.~J{\'e}gou, J.~Mairal, P.~Bojanowski, and A.~Joulin, ``Emerging properties in self-supervised vision transformers,'' in \emph{ICCV}, 2021, pp. 9650--9660.

\bibitem{kirillov2023segment}
A.~Kirillov, E.~Mintun, N.~Ravi, H.~Mao, C.~Rolland, L.~Gustafson, T.~Xiao, S.~Whitehead, A.~C. Berg, W.-Y. Lo \emph{et~al.}, ``Segment anything,'' in \emph{ICCV}, 2023, pp. 4015--4026.

\bibitem{jia2022generative}
W.~Jia, M.~Liu, and J.~M. Rehg, ``Generative adversarial network for future hand segmentation from egocentric video,'' in \emph{ECCV}, 2022, pp. 639--656.

\bibitem{lai2024listen}
B.~Lai, F.~Ryan, W.~Jia, M.~Liu, and J.~M. Rehg, ``Listen to look into the future: Audio-visual egocentric gaze anticipation,'' in \emph{ECCV}, 2024.

\bibitem{lai2022eye}
B.~Lai, M.~Liu, F.~Ryan, and J.~M. Rehg, ``In the eye of transformer: Global-local correlation for egocentric gaze estimation,'' in \emph{BMVC}, 2022.

\bibitem{lai2024eye}
B.~Lai, M.~Liu, F.~Ryan, and J.~Rehg, ``In the eye of transformer: Global--local correlation for egocentric gaze estimation and beyond,'' \emph{IJCV}, vol. 132, no.~3, pp. 854--871, 2024.

\bibitem{lai2024lego}
B.~Lai, X.~Dai, L.~Chen, G.~Pang, J.~M. Rehg, and M.~Liu, ``Lego: Learning egocentric action frame generation via visual instruction tuning,'' in \emph{ECCV}, 2024.

\bibitem{zhang2024hoidiffusion}
M.~Zhang, Y.~Fu, Z.~Ding, S.~Liu, Z.~Tu, and X.~Wang, ``Hoidiffusion: Generating realistic 3d hand-object interaction data,'' in \emph{CVPR}, 2024, pp. 8521--8531.

\bibitem{ye2023affordance}
Y.~Ye, X.~Li, A.~Gupta, S.~De~Mello, S.~Birchfield, J.~Song, S.~Tulsiani, and S.~Liu, ``Affordance diffusion: Synthesizing hand-object interactions,'' in \emph{CVPR}, 2023, pp. 22\,479--22\,489.

\bibitem{bls2023atus}
\BIBentryALTinterwordspacing
``American time use survey,'' 2023, accessed: May 30, 2023. [Online]. Available: \url{https://www.bls.gov/tus/}
\BIBentrySTDinterwordspacing

\bibitem{ramanathan2016detecting}
V.~Ramanathan, J.~Huang, S.~Abu-El-Haija, A.~Gorban, K.~Murphy, and L.~Fei-Fei, ``Detecting events and key actors in multi-person videos,'' in \emph{CVPR}, 2016, pp. 3043--3053.

\bibitem{toyer2017human}
S.~Toyer, A.~Cherian, T.~Han, and S.~Gould, ``Human pose forecasting via deep {M}arkov models,'' in \emph{International Conference on Digital Image Computing: Techniques and Applications}, 2017, pp. 1--8.

\bibitem{carreira2017quo}
J.~Carreira and A.~Zisserman, ``Quo vadis, action recognition? {A} new model and the {K}inetics dataset,'' in \emph{CVPR}, 2017, pp. 6299--6308.

\bibitem{kay2017kinetics}
W.~Kay, J.~Carreira, K.~Simonyan, B.~Zhang, C.~Hillier, S.~Vijayanarasimhan, F.~Viola, T.~Green, T.~Back, P.~Natsev \emph{et~al.}, ``The {K}inetics human action video dataset,'' \emph{arXiv preprint arXiv:1705.06950}, 2017.

\bibitem{carreira2018short}
J.~Carreira, E.~Noland, A.~Banki-Horvath, C.~Hillier, and A.~Zisserman, ``A short note about {K}inetics-600,'' \emph{arXiv preprint arXiv:1808.01340}, 2018.

\bibitem{carreira2019short}
J.~Carreira, E.~Noland, C.~Hillier, and A.~Zisserman, ``A short note on the {K}inetics-700 human action dataset,'' \emph{arXiv preprint arXiv:1907.06987}, 2019.

\bibitem{shao2020finegym}
D.~Shao, Y.~Zhao, B.~Dai, and D.~Lin, ``{F}ine{G}ym: A hierarchical video dataset for fine-grained action understanding,'' in \emph{CVPR}, 2020, pp. 2616--2625.

\bibitem{ragusa2021meccano}
F.~Ragusa, A.~Furnari, S.~Livatino, and G.~M. Farinella, ``The meccano dataset: Understanding human-object interactions from egocentric videos in an industrial-like domain,'' in \emph{WACV}, 2021, pp. 1569--1578.

\bibitem{richard2017weakly}
A.~Richard, H.~Kuehne, and J.~Gall, ``Weakly supervised action learning with {RNN} based fine-to-coarse modeling,'' in \emph{CVPR}, 2017, pp. 754--763.

\bibitem{lea2017temporal}
C.~Lea, M.~D. Flynn, R.~Vidal, A.~Reiter, and G.~D. Hager, ``Temporal convolutional networks for action segmentation and detection,'' in \emph{CVPR}, 2017, pp. 156--165.

\bibitem{li2020ms}
S.-J. Li, Y.~AbuFarha, Y.~Liu, M.-M. Cheng, and J.~Gall, ``{MS-TCN++}: Multi-stage temporal convolutional network for action segmentation,'' \emph{TPAMI}, pp. 1--1, 2020.

\bibitem{vaswani2017attention}
A.~Vaswani, N.~Shazeer, N.~Parmar, J.~Uszkoreit, L.~Jones, A.~N. Gomez, {\L}.~Kaiser, and I.~Polosukhin, ``Attention is all you need,'' \emph{NeurIPS}, vol.~30, 2017.

\bibitem{papineni2002bleu}
K.~Papineni, S.~Roukos, T.~Ward, and W.-J. Zhu, ``Bleu: a method for automatic evaluation of machine translation,'' in \emph{ACL}, 2002, pp. 311--318.

\bibitem{banerjee2005meteor}
S.~Banerjee and A.~Lavie, ``Meteor: An automatic metric for mt evaluation with improved correlation with human judgments,'' in \emph{ACL workshop on intrinsic and extrinsic evaluation measures for machine translation and/or summarization}, 2005, pp. 65--72.

\end{thebibliography}
\label{sec:refs}

\clearpage

\setcounter{section}{0}

\qquad \\
\begin{center}
    \textbf{\LARGE Supplementary} \\ [1.0cm]
\end{center}

\renewcommand\thesection{\Alph {section}}
\renewcommand\thesubsection{\thesection.\arabic{subsection}}

\section{Datasets}\label{app:datasets}

In this section, we provide more details for each dataset listed in Table \ref{tab:datasets}. For a thorough review, we also introduce some datasets that are used in only a few action anticipation papers. The datasets are introduced in chronological order but we will group some strongly relevant datasets as a series.

\textbf{TV Human Interaction (TVHI)} \cite{patron2010high,patron2012structured} is an early dataset of human social actions. All videos in TVHI are from 20 different TV shows. There are only four action categories annotated in this dataset -- hand shakes, high fives, hugs and kisses. Additional labels include the bounding boxes of upper bodies and head orientations.

\textbf{VIRAT} \cite{oh2011large} is a dataset of outdoor surveillance videos. It includes both stationary CCTV footage and aerial footage, as well as a mix of scripted and unscripted behavior. The action classes are focused on events of interest to surveillance systems, like a person entering or leaving a building, interacting with another person or interacting with a vehicle.

The \textbf{Georgia Tech Egocentric Activities (GTEA) datasets} are a series of egocentric action recognition datasets featuring subjects preparing food in various environments. All datasets include untrimmed videos annotated with actions in the familiar verb-noun form (i.e., ``subject performs \emph{verb} using \emph{noun \#1, noun \#2, \ldots{}}''). However, they differ along several other axes, including total length, whether or not gaze and audio data are available, and whether or not the dataset includes hand masks, as well as the environment used for recording.

The original \textbf{GTEA dataset} \cite{fathi2011learning,li2015delving} dataset includes GoPro videos of four participants in a lab environment making several types of food and drink. The set list of recipes include things like hotdog sandwiches and coffee with honey. 
\textbf{GTEA Gaze} \cite{fathi2012learning} is a slightly larger dataset with a similar theme, but with gaze data from eye-tracking glasses.
GTEA Gaze also differs in that subjects are simply given a table of ingredients and asked to make a meal for themselves, with no constraints on the precise type of meals.

\textbf{GTEA Gaze+} \cite{fathi2012learning,li2015delving} and its extension, \textbf{EGTEA Gaze+} \cite{li2018eye}, are larger egocentric eye-tracking datasets.
The data collection methodology for these datasets differs in two substantial ways.
One is that the scenes are recorded in a full kitchen rather than on a table in a lab.
The other is that each subject is given a specific recipe with step-by-step instructions, and the resulting activities are both longer and more tightly scripted than GTEA Gaze.

\textbf{MPII Cooking} \cite{rohrbach2012database} is a dataset of untrimmed third-person cooking videos and action annotations, with an emphasis on distinguishing fine-grained actions (like cutting versus peeling a vegetable). Every video in the dataset is recorded from the same stationary camera in a specially-equipped kitchen, and features recruited subjects making various dishes. The videos are loosely scripted: subjects are told a dish along with some suggested ingredients, tools and verbal instructions, but they sometimes deviate from those suggestions. The dataset also includes pose annotations for a small subset of around 2,400 frames.

\textbf{MPII Composites} \cite{rohrbach2012script} is a similar dataset filmed in the same kitchen environment, but is slightly larger and has richer action annotations.
Specifically, the fine-grained action classes are each decomposed into a verb and a list of nouns (objects) that the verb applies to. Each video in the dataset also has a global label describing the type of dish being cooked. The authors refer to the dishes as \textit{composite} activities because they can be decomposed into sequences of fine-grained actions. MPII Composites also includes ``scripts'': step-by-step natural language instructions for how to cook each of the dishes, which are generated by Mechanical Turk workers. The scripts are not aligned with specific videos (and sometimes depict dishes for which no video is recorded), but in principle could be useful for modeling the expected high-level structure of the cooking videos.
In this sense they are like the recipes.

MPII Composites has since been extended with further natural language annotations in \textbf{TACoS} \cite{regneri2013grounding} and \textbf{TACoS-ML} \cite{rohrbach2014coherent}.
TACoS adds temporally-aligned single-sentence captions to each video that describe the low-level actions taken by the subject.
It also has a set of human-annotated subjective similarity annotations for natural language caption pairs.
TACoS-ML further adds temporally-unaligned natural language descriptions of each video.
These temporally-unaligned descriptions are collected at three levels of detail ranging between one sentence and 15 sentences, which may be useful for hierarchical language-based action modeling.

\textbf{MPII Cooking 2} unifies MPII Cooking with MPII Composites \cite{rohrbach2015recognizing}.
In addition to the videos and annotations from those datasets, MPII Cooking 2 adds a handful of new videos and updates some of the annotations for accuracy.
The authors note that the diverse types of annotation in the resulting dataset make it useful for a range of tasks, including fine-grained action recognition, recursive decomposition of composite activities (i.e., decomposing dishes into fine-grained actions), and pose-guided modeling of actions using either upper body or hand pose annotations.

\textbf{50Salads} \cite{stein2013combining} depicts subjects repeatedly preparing the same salad in an instrumented test kitchen.
It includes RGB and depth video from a stationary, downward-facing Kinect camera, as well as data from accelerometers mounted on various kitchen objects.
The activities are semi-scripted: the dataset authors sampled desired action orderings from a generative model and asked the subjects to each follow a specific ordering, but in practice the subjects deviated from their supplied orderings.

\textbf{Breakfast Actions} \cite{kuehne2014language} is another dataset of cooking videos.
Videos depict recruited subjects preparing various food items in ordinary kitchens, typically filmed from 3-5 stationary camera views.
The subjects were each instructed to cook a specific dish, but were not given a specific sequence of steps to follow.
Unlike most other action recognition datasets, Breakfast Actions includes both coarse-grained action annotations and (partially complete) fine-grained action annotations, which makes it useful for learning hierarchical action representations. The HERA paper \cite{morais2020learning} argues that there are several flaws in the original hierarchical annotations for Breakfast Actions. In particular, the original fine-grained annotations only cover a subset of videos, and the HERA authors argue that the fine-grained actions do not form a very meaningful semantic hierarchy with the coarse-grained actions. This may be because the coarse- and fine-grained annotations were produced by two independent annotation teams. Morais et al. \cite{morais2020learning} address this issue by releasing an improved set of hierarchical action annotations for Breakfast Actions, which they refer to as \emph{Hierarchical Breakfast}.

\textbf{THUMOS} \cite{THUMOS14} is a dataset comprising both temporally trimmed and untrimmed videos. Specifically, all videos in the training set are trimmed videos, i.e., each video corresponding to one single action, while the validation and test sets are composed of untrimmed videos. The authors propose two benchmarks based on this dataset -- action recognition and temporal action detection. Given that action anticipation requires sequential actions in untrimmed videos, prior work typically trains models on the validation set and evaluate them on the test set, using the 20-class action labels from temporal actoin detection benchmark.

\textbf{MultiTHUMOS} \cite{yeung2018every} extends the THUMOS'14 dataset \cite{THUMOS14} to denser annotations and multiple action labels per frame.
The original THUMOS dataset includes 20 action classes (including actions related to sports, instruments, and basic motions) and 6,365 action segment labels; MultiTHUMOS increases the total to 65 classes and 38,690 annotations, including overlapping action segments and more fine-grained actions.
Yeung et al. \cite{yeung2018every} argue that new resulting simultaneous action labels are both better at describing human action and useful for capturing the interdependence between actions.
For instance, someone taking both the ``drive car'' and ``fry egg'' actions is likely to take the ``crash car'' and ``wipe off eggs'' actions in the near future.

\textbf{ActivityNet} \cite{caba2015activitynet} is a large dataset of untrimmed online videos depicting everyday activities. The action classes correspond to activities from the American Time Use Survey (ATUS) \cite{bls2023atus}, which is a fine-grained survey of the way that Americans spend their time outside of work. The videos were obtained by searching for queries related to ATUS activities on YouTube and related platforms. They were then labeled with action segment labels, resulting in an average of 1.4 action instances per video. Unfortunately the small number of action instances means that ActivityNet is of limited use for action anticipation, and has only been used in a handful of papers. Furthermore, it's worth noting that the annotated action segments span only 68.8 hours of the 849 total hours of video, so most frames in the untrimmed ActivityNet videos are unlabeled.

\textbf{Brain4Cars} \cite{jain2016brain4cars} consists of natural driving videos from two views -- inside view that captures drivers' head and hand movements, and outside view captures the road ahead. There are 700 annotated events including lane change, turn, etc. However, this dataset is only used by very few papers because it only covers the driving situation and the action categories are also limited.

\textbf{TV Series} \cite{de2016online} is another dataset that curates data from TV shows. The dataset comprises 16-hour videos with 30 labeled actions. Different than TVHI \cite{patron2010high,patron2012structured} that focuses on social interactions, all actions in TV Series can be conducted by the subject alone (e.g., drink, wave and write). It's worth noting that TV Series also includes lots of auxiliary annotations and metadata, such as whether there are more than one person in the video, whether there's a shot cut and whether there's any occlusion. These kinds of information are very important to exclude some outliers and noise in model training.

\textbf{Charades} \cite{sigurdsson2016hollywood} depicts people acting out sequences of simple actions in their own homes.
Sigurdsson et al. created the dataset by asking one set of crowd workers to write scripts depicting plausible household activities.
They then distributed the scripts to another set of crowd workers, and asked the crowd workers to film themselves acting out the scripts in their own home.
Finally, they annotated the videos with action segment labels, active object classes, and video-level natural language descriptions.
The resulting dataset is notable for the wide variety of behaviors and environments that it depicts, although the fact that subjects merely act out activities does make the behavior somewhat artificial.

\textbf{Narrated Instruction Videos (NIV)} \cite{alayrac2016unsupervised} is a dataset of annotated how-to-videos from five categories, including making coffee, performing CPR, and jump-starting a car.
The videos, which were scraped from YouTube, include ASR captions that have been manually corrected by the authors.
The authors also defined a generic sequence of steps for each task by consulting how-to websites, then labeled the start and end times for each of those steps in each video.
Surprisingly, only 6\% of the steps appear out of the order implied by the authors' scripts, even though the videos for a given task were likely all produced by different people.
Thus, while this dataset is technically not scripted by the authors, the depicted behaviors do occur in a very predictable order.

The \textbf{NCAA Basketball} dataset \cite{ramanathan2016detecting} contains hundreds of long videos of basketball games.
The dataset is annotated with timestamps and classes for 11 significant basketball events (such as success or failure of a shot, or stealing a ball).
Some frames in which a shot is taken include an annotation of the ball location, while other frames include player bounding boxes.
The authors argue that the dataset is challenging in part because of the need to attend only to relevant people and object during multi-person interaction, and they propose a method that uses the additional person and ball annotations for this purpose.

The \textbf{Joint Attention in Autonomous Driving (JAAD)} dataset \cite{kotseruba2016joint} is a dataset of dashcam videos that focuses on modeling driver--pedestrian interactions.
It includes action segment annotations for drivers and pedestrians, pedestrian bounding boxes, and metadata describing pedestrian appearance attributes and road conditions.
The videos are short, at 5-15 seconds each, but still have relatively dense annotations, which makes the dataset useful for training and evaluating action anticipation models.

\textbf{Ikea Furniture Assembly (IkeaFA)} \cite{han2017human,toyer2017human} shows recruited actors repeatedly assembling and disassembling the same Ikea table.
The videos average around 2.85 minutes and typically contain around 21 action segments.
It's worth noting that although the videos are moderately long, the action sequences are still highly predictable, since most videos contain the same actions in the same order.

\textbf{Stanford-ECM} \cite{nakamura2017jointly} is another dataset containing various activities of daily living, but with accelerometer and heart rate data as additional modalities.
The dataset has labeled action segments, and each action class is associated with a particular Metabolic Energy Expenditure (MET), which makes it possible to predict the camera-wearer's approximate energy expenditure.
Notably, the dataset is not heavily scripted: subjects were instructed to go about their daily activities while wearing a camera, but were otherwise free to act however they pleased.

\textbf{Atomic Visual Actions (AVA)} \cite{gu2018ava} is a large dataset of film clips with multi-person action annotations and bounding boxes.
Each video in AVA is a 15 minute clip of a longer film taken from YouTube.
The videos are annotated at 1Hz with action labels and bounding boxes for each person, and the bounding boxes are linked across frames to form tracks.
The action taxonomy allows a given person to take multiple actions in a single frame, which allows the dataset to simultaneously capture the state of each actor as well as their interactions with other people and objects.

\textbf{YouCook2} \cite{zhou2018towards} is a dataset of YouTube cooking videos annotated with natural language descriptions of the steps.
Each natural language description is an imperative English sentence, and has an associated start time and end time.
The videos have 7.7 segments on average, with a mean duration of around 20 seconds.
Zhou et al. \cite{zhou2018towards} use the dataset to learn how to segment videos, although it could be used for natural language action anticipation as well.

\textbf{Epic-Kitchens 55 (EK55)} \cite{damen2018scaling} is a dataset of egocentric videos of kitchen activities.
The activities are unscripted: subjects were simply given a head-mounted camera and asked to record themselves each time they entered the kitchen.
The dataset includes action annotations of the familiar verb-plus-nouns form, as well as object bounding boxes corresponding to the ``active'' noun in each action.
Due to the way the dataset was constructed, it also includes supplementary natural language descriptions of the actions (as opposed to just categorical labels), which may be useful for language-based methods.

\textbf{Epic-Kitchens 100 (EK100)} \cite{damen2020rescaling} is an extension of Epic-Kitchens 55 which includes all original videos as well as an extra 45 hours of new video.
The annotations of EK100 are denser and more accurate thanks to an improved annotation process.
The new videos in EK100 also include accelerometer and gyroscope data, but no manually-annotated active object bounding boxes (although the dataset ships with automatically estimated object bounding boxes and hand masks to make up for this).

\textbf{CrossTask} \cite{zhukov2019cross} is a dataset of cooking, car maintenance, and craft videos which is designed to examine cross-task knowledge transfer.
Each task corresponds to a WikiHow article, which was used to derive an ordered, human-curated sequence of actions describing the task.
Videos for each task were collected from YouTube.
The dataset is split into a set of 18 \emph{primary} tasks, which have fully annotated videos with action labels and temporal boundaries, and 63 \emph{related} tasks, which have only unannotated videos and are intended for unsupervised pretraining.
The set of tasks was chosen to have many shared actions, and a near-deterministic ordering of actions within each task.
While the near-deterministic ordering of steps enables transfer of knowledge between tasks, it's worth noting that it also makes the dataset less challenging for action anticipation.

\textbf{COIN} \cite{tang2019coin} is another action recognition dataset of YouTube videos.
COIN is relatively large (477h), and spans a wide variety of tasks (driving, nursing, sports, etc.).
It also has human-verified labels, unlike HowTo100M.
The action labels are also arranged into a semantic hierarchy, in which each action label (level 3) belongs to a task (level 2) and a domain. There is no sharing of action labels across tasks.

\textbf{Pedestrian Intention Estimation (PIE)} \cite{rasouli2019pie} is a video dataset of pedestrian's actions captured by a car-mounted camera. The dataset has bounding box labels of pedestrians, vehicles and many elements on the street, such as traffic lights, signs and zebra crossings. The action labels include standing, looking, crossing, etc. In prior work of action anticipation, PIE data is used to forecast whether or not the pedestrian will cross the street. Similar to Brain4Cars \cite{jain2016brain4cars}, PIE is also only used in very few papers because of the small amount of action categories.

\textbf{Kinetics-700-2020} \cite{smaira2020short} is the most recent release of a series of datasets for action classification in trimmed videos \cite{carreira2017quo,kay2017kinetics,carreira2018short,carreira2019short,smaira2020short}.
The videos in Kinetics are short ($\sim$10s) clips from YouTube videos, while the class taxonomy is relatively broad and includes labels taken from existing dataset (like AVA \cite{gu2018ava} and Epic Kitchens \cite{damen2018scaling}) as well as labels suggested by crowd workers.
Kinetics is not generally used as a benchmark for action anticipation because it does not precisely localize actions within the source YouTube video, or exhaustively label every action that occurs within a given clip. However, it's commonly used as a pretraining dataset for action anticipation models, much as ImageNet is used as a pretraining dataset for image-based prediction tasks \cite{carreira2017quo}.
Thus we include details of the most recent iteration of Kinetics to give an idea of the scale and contents of the dataset.

\textbf{FineGym} \cite{shao2020finegym} is a new dataset built on top of gymnasium videos. The most important distinction of FineGym compared with existing datasets is the hierarchical action labels. All videos are annotated using a three-level hierarchy. The high-level action (e.g., balance beam) are further annotated with several sub-actions (e.g., leap-jumphop, beam-turns, flight-salto, flight-handspring, and dismount). Each sub-action is then mapped to a pre-defined class category. This feature makes FineGym an appropriate dataset for goal and final action anticipation.

\textbf{Oops!} \cite{epstein2020oops} is another dataset used for goal and final action anticipation. It contains 20,723 videos collected from YouTube totaling up more than 50 hours. The video duration ranges from 1 second to 30 seconds with the duration of most videos shorter than 10 seconds. The unique feature of Oops! is that this dataset contains both intentional and unintentional actions as well as the corresponding labels. The moment when the action becomes unintentional is also annotated in each video.

\textbf{BDD100k (BDD)} \cite{yu2020bdd100k} is an additional driving video dataset in addition to Brain4Cars \cite{jain2016brain4cars} and PIE \cite{rasouli2019pie}. The videos capture the road condition in front of the car during driving. Though lots of annotations are provided in this dataset, it is mainly used by Neumann et al. \cite{neumann2019future} to forecast whether the car will stop in the near future.

\textbf{Stanford -TRI Intention Prediction (STIP)} \cite{liu2020spatiotemporal} is also a video dataset collected by car-mounted cameras. STIP contains aligned videos captured in three views (i.e., left, right and front), which enables more investigation of multi-view fusion. It also has the largest number of pedestrian bounding boxes (around 350,000) making it the ideal testbed for the forecast of intention-to-cross.

\textbf{MECCANO} \cite{ragusa2021meccano} is a multi-modal egocentric video dataset that captures the subject building a toy model of a motorbike in the industrial-like environment. 8,857 video segments are annotated with the actions labels from 61 classes. The dataset also has bounding boxes of hands and active objects. It's worth noting that the gaze and depth data is also recorded simultaneously with the RGB videos.

\textbf{Assembly101} \cite{sener2022assembly101} is the first mutli-view action dataset with synchronized egocentric and exocentic videos. The participants were asked to assemble and disassemble 101 toy vehicles without specific instructions, which improves the variation and diversity of the action sequences. The videos are recorded by eight static cameras and four wearable cameras simultaneously totaling 513 hours of footage. 202 coarse action classes and 1,380 fine-grained action classes are pre-defined and labeled on Assembly101 with 3D hand poses.

\textbf{Ego4D} \cite{grauman2022ego4d} is a new large-scale egocentric video dataset collected globally. It covers a broad range of human activities including dancing, cooking, operating machines, painting the house, etc. There are two pre-defined action anticipation benchmarks on Ego4D -- short-term object interaction anticipation and long-term anticipation. Short-term object interaction anticipation forecasts a verb indicating the future interaction of the subject and object as well as predicts when the interaction will begin. Long-term action anticipation requires predicting the future action sequence which is the same as our definition. Ego4D has bounding boxes of the active objects and short human narrations for each action. It enables object-centric forecasting approaches and visual-language pre-training. Song et al. \cite{song2024ego4d} make a new set of annotations -- \textbf{Ego4D Goal-Step} to Ego4D videos, which includes dense annotations (hierarchical action labels) for 4,800 video segments (totaling 480 hours) and goal annotations for 2,807 hours of videos. The large amount of annotations of various granularities can be used for goal anticipation and procedure planning.

\section{Metrics}\label{app:metrics}

\subsection{Classification Metrics}

Assuming there are $C$ classes of actions, we denote the number of samples of $i$-th class as $N_i$, true positive samples of $i$-th class as $TP_i$, false positive samples of $i$-th class as $FP_i$ and false negative samples of $i$-th class as $FN_i$. 

\textbf{Top-k Accuracy.} Given that forecasting future action categories is essentially an classification problem. The most common metric is top-k accuracy. The data sample is seen as correct when the ground truth is in the top-k predictions. Then the accuracy is calculate by
\begin{equation}
    Top\mbox{-}k\ \ Accuracy = \frac{\sum_{i=1}^C TP_i}{\sum_{i=1}^C N_i}.
\end{equation}
Usually prior work uses top-1 accuracy and top-5 accuracy.

\textbf{Mean Top-k Recall}. One drawback of top-k accuracy is that it can be easily biased to frequent action categories when it's used on datasets with long-tail action distribution. Mean top-k recall is used in class-imbalanced situations, which can be written as
\begin{equation}
    Mean\ \ Top\mbox{-}k\ \ Recall = \frac{1}{C}\sum_{i=1}^{C}\frac{TP_i}{N_i}.
\end{equation}.

\textbf{Mean-over-class Accuracy.} Mean-over-class (MoC) accuracy is usually used for dense action anticipation (i.e., forecasting on each video frame). MoC accuracy is essentially the \textit{mean top-1 recall} over all future frames.

\textbf{Mean Average Precision.} For the $i$-th action class, the recall and precision can be formulated as
\begin{align}
    Recall_i &= \frac{TP_i}{TP_i + FP_i}, \label{eq:recall} \\
    Precision_i &= \frac{TP_i}{TP_i + FN_i}.
\end{align}
A precision-recall curve can be obtained by adjusting the classification threshold. If we denote the precision as a function of recall on the curve. Average precision (AP) is the area under curve and mean average precision (mAP) is the average AP across all classes. Specifically, AP of $i$-th class and mAP are formulated as
\begin{align}
    AP_i &= \int_0^1 Precision_i(r)dr, \label{eq:ap} \\
    mAP &= \frac{1}{C}\sum_{i=1}^C AP_i. \label{eq:map}
\end{align}

\textbf{Calibrated Average Precision.} Calibrated average precision (cAP) is a variant of mAP that considers the class imbalance problem. In cAP, recall is calculated following Equation \ref{eq:recall}, while there's a modification in precision calculation. A parameter $w_i$, which is the ratio of negative samples to positive samples, is used to balance the positive samples and negative samples. In this way, the calibrated precision of $i$-th class is written as
\begin{equation}
    Calibrated\ \ Precision_i = \frac{TP_i}{TP_i+\frac{FN_i}{w_i}}.
\end{equation}
Then cAP is obtained following Equation \ref{eq:ap} and Equation \ref{eq:map}. Intuitively, cAP is the mAP metric calculated as if there is a balanced number of positive and negative data samples.

\subsection{Language and Sequential Classification Metrics}

\begin{table*}[t]
\renewcommand{\arraystretch}{1.3}
\caption{
    Action anticipation results on Breakfast Actions \cite{kuehne2014language} and 50Salads \cite{stein2013combining}. $\frac{x\%}{y\%}$ in the column headings means that the model is given $x$\% of the video as observed contexts, and must forecast action labels for the next $y$\% of the video. The performance is measured by mean-over-classes (MoC) accuracy (\%), which is the top-1 accuracy averaged across future frames and then across classes. We group the existing approaches by the input types.
}
\label{tab:res-ba-50s}
\centering
\begin{threeparttable}
\resizebox{\linewidth}{!}{
\begin{tabular}{llccccccccccccccccc}
\toprule
& & & \multicolumn{8}{c}{\textbf{Breakfast Actions}} & \multicolumn{8}{c}{\textbf{50Salads}} \\
\cmidrule(lr){4-11} \cmidrule(l){12-19}
\textbf{Methods} & \textbf{Input} & & \multicolumn{4}{c}{\textbf{20\%}} & \multicolumn{4}{c}{\textbf{30\%}} & \multicolumn{4}{c}{\textbf{20\%}} & \multicolumn{4}{c}{\textbf{30\%}} \\
\cmidrule(lr){4-7} \cmidrule(lr){8-11} \cmidrule(lr){12-15} \cmidrule(l){16-19}
 & & & \textbf{10\%} & \textbf{20\%} & \textbf{30\%} & \textbf{50\%} & \textbf{10\%} & \textbf{20\%} & \textbf{30\%} & \textbf{50\%} & \textbf{10\%} & \textbf{20\%} & \textbf{30\%} & \textbf{50\%} & \textbf{10\%} & \textbf{20\%} & \textbf{30\%} & \textbf{50\%} \\
\midrule
CNN \cite{abu2018will}                  & Ground Truth of Past Actions     &       & 58.0                 & 49.1                 & 44.0                 & 39.3                 & 60.3                 & 50.1                 & 45.2                 & 40.5                 & 36.1                 & 27.6                 & 21.4                 & 15.5                 & 37.4                 & 24.8                 & 20.8                 & 14.1 \\
RNN \cite{abu2018will}                  & Ground Truth of Past Actions     &       & 60.4                 & 50.4                 & 45.3                 & 40.4                 & 61.5                 & 50.3                 & 44.9                 & 41.8                 & 42.3                 & 31.2                 & 25.2                 & 16.8                 & 44.2                 & 29.5                 & 20.0                 & 10.4 \\
Ke et al. \cite{ke2019time}                       & Ground Truth of Past Actions     &       & 64.5                 & 56.3                 & 50.2                 & 44.0                 & 66.0                 & 55.9                 & 49.1                 & 44.2                 & 45.1                 & 33.2                 & 27.6                 & 17.3                 & 46.4                 & 34.8                 & 25.2                 & 13.8 \\
UAAA (Mode) \cite{abu2019uncertainty}   & Ground Truth of Past Actions     &       & 53.0                 & 44.1                 & 39.7                 & 34.9                 & 54.0                 & 44.5                 & 40.2                 & 35.6                 & 38.1                 & 30.1                 & 26.3                 & 16.5                 & 40.0                 & 29.3                 & 23.2                 & 15.5 \\
UAAA (Top1) \cite{abu2019uncertainty}   & Ground Truth of Past Actions     &       & 78.8                 & 72.8                 & 66.3                 & 63.5                 & 82.0                 & 72.8                 & 69.1                 & 62.4                 & 74.9                 & 58.8                 & 46.1                 & 35.7                 & 67.4                 & 52.4                 & 46.7                 & 36.6 \\
Gammulle et al. \cite{gammulle2019forecasting}          & Ground Truth of Past Actions     &       & 87.2                 & 85.2                 & 81.0                 & 75.5                 & 87.9                 & 85.8                 & 82.1                 & 76.3                 & 70.0                 & 64.3                 & 62.7                 & 52.2                 & 68.1                 & 62.3                 & 61.2                 & 56.7 \\
TempAgg \cite{sener2020temporal,sener2021technical} & Ground Truth of Past Actions        &               & 65.5                 & 55.5                 & 46.8                 & 40.1                 & 67.4                 & 56.1                 & 47.4                 & 41.5                 & 47.2                 & 34.6                 & 30.5                 & 19.1                 & 44.8                 & 32.7                 & 23.5                 & 15.3 \\
MAVAP \cite{loh2022long}                & Ground Truth of Past Actions     &       & 69.1                 & 54.1                 & 45.4                 & 35.1                 & 70.9                 & 56.2                 & 47.6                 & 38.1                 & 43.3                 & 35.4                 & 28.4                 & 17.4                 & 43.8                 & 31.8                 & 27.2                 & 14.2 \\

\arrayrulecolor{black!20}\midrule\arrayrulecolor{black}

CNN \cite{abu2018will}                  & RNN-HMM \cite{richard2017weakly} Inferred Actions    &        & 17.9                 & 16.4                 & 15.4                 & 14.5                 & 22.4                 & 20.1                 & 19.7                 & 18.8                 & 21.2                 & 19.0                 & 16.0                 & 9.9                  & 29.1                 & 20.1                 & 17.5                 & 10.9 \\
RNN \cite{abu2018will}                  & RNN-HMM \cite{richard2017weakly} Inferred Actions    &        & 18.1                 & 17.2                 & 15.9                 & 15.8                 & 21.6                 & 20.0                 & 19.7                 & 19.2                 & 30.1                 & 25.4                 & 18.7                 & 13.5                 & 30.8                 & 17.2                 & 14.8                 & 9.8 \\
Ke et al. \cite{ke2019time}                       & RNN-HMM \cite{richard2017weakly} Inferred Actions    &        & 18.4                 & 17.2                 & 16.4                 & 15.8                 & 22.8                 & 20.4                 & 19.6                 & 19.8                 & 32.5                 & 27.6                 & 21.3                 & 16.0                 & 35.1                 & 27.1                 & 22.1                 & 15.6 \\
UAAA (Mode) \cite{abu2019uncertainty}   & RNN-HMM \cite{richard2017weakly} Inferred Actions    &        & 16.7                 & 15.4                 & 14.5                 & 14.2                 & 20.7                 & 18.3                 & 18.4                 & 16.9                 & 24.9                 & 22.4                 & 19.9                 & 12.8                 & 29.1                 & 20.5                 & 15.3                 & 12.3 \\
UAAA (Top1) \cite{abu2019uncertainty}   & RNN-HMM \cite{richard2017weakly} Inferred Actions    &        & 28.9                 & 28.4                 & 27.6                 & 28.0                 & 32.4                 & 31.6                 & 32.8                 & 30.8                 & 53.5                 & 43.0                 & 50.5                 & 33.7                 & 56.4                 & 42.8                 & 35.8                 & 30.2 \\
AGG \cite{piergiovanni2020adversarial}  & RNN-HMM \cite{richard2017weakly} Inferred Actions    &        & -                    & -      
              & -                    & -                    & -                    & -                    & -                    & -      
              & 40.4                 & 33.7                 & 25.4                 & 20.9                 & 40.7                 & 40.1   & 26.4                 & 19.2 \\
TempAgg \cite{sener2020temporal,sener2021technical}            & RNN-HMM \cite{richard2017weakly} Inferred Actions    &        & 18.8                 & 16.9                 & 16.5                 & 15.4                 & 23.0                 & 20.0                 & 19.9                 & 18.6                 & 32.7                 & 26.3                 & 21.9                 & 15.6                 & 32.3                 & 25.5                 & 22.7                 & 17.1 \\

\arrayrulecolor{black!20}\midrule\arrayrulecolor{black}

Ng et al. (WS) \cite{ng2020forecasting}      & Fisher Vector Features                           &        & 18.6                 & 16.7   
              & 14.8                 & 14.7                 & 23.8                 & 21.2                 & 19.3                 & 16.8   
              & 36.4                 & 26.3                 & 23.4                 & 15.5                 & 35.4                 & 26.4   
              & 23.7                 & 18.4 \\
Ng et al. (FS) \cite{ng2020forecasting}      & Fisher Vector Features                           &        & 18.8                 & 18.4   
              & 17.8                 & 17.0                 & 24.0                 & 21.9                 & 19.7                 & 19.6   
              & 39.3                 & 31.4                 & 27.0                 & 23.9                 & 41.7                 & 32.7   
              & 31.4                 & 26.4 \\
TempAgg \cite{sener2020temporal,sener2021technical}    & Fisher Vector Features                 &        & 15.6                 & 13.1                 & 12.1                 & 11.1                 & 19.5                 & 17.0                 & 15.6                 & 15.1                 & 25.5                 & 19.9                 & 18.2                 & 15.1                 & 30.6                 & 22.5                 & 19.1                 & 11.2 \\
TempAgg \cite{sener2020temporal,sener2021technical}    & Fisher Vector + RNN-HMM Actions        &        & 25.0                 & 21.9   
              & 20.5                 & 18.1                 & 23.0                 & 20.5                 & 19.8                 & 19.8   
              & 34.7                 & 25.9                 & 23.7                 & 15.7                 & 34.5                 & 26.1   
              & 19.0                 & 15.5 \\

\arrayrulecolor{black!20}\midrule\arrayrulecolor{black}

Ng et al. (WS) \cite{ng2020forecasting}      & I3D \cite{carreira2017quo} Features             &        & 21.7                 & 18.9                 & 16.7                 & 14.6                 & 26.2                 & 21.8                 & 20.4                 & 16.5                 & -                    & -                    & -                    & -                    & -                    & -                    & -                    & - \\
Ng et al. (FS) \cite{ng2020forecasting}      & I3D \cite{carreira2017quo} Features             &        & 23.0                 & 22.3                 & 22.0                 & 20.9                 & 26.5                 & 25.0                 & 24.1                 & 23.6                 & -                    & -                    & -                    & -                    & -                    & -                    & -                    & - \\
TempAgg \cite{sener2020temporal,sener2021technical}   & I3D \cite{carreira2017quo} Features    &        & 24.2                 & 21.2                 & 20.0                 & 18.1                 & 30.4                 & 26.3                 & 23.8                 & 21.2                 & -                    & -                    & -                    & -                    & -                    & -                    & -                    & - \\
SRL \cite{qi2021self}                        & I3D \cite{carreira2017quo} Features             &        & 25.6                 & 21.0   
              & 18.5                 & 16.0                 & 27.3                 & 23.6                 & 20.8                 & 17.3   & -                    & -                    & -                    & -                    & -                    & -      & -                    & - \\
A-ACT \cite{gupta2022act}                    & I3D \cite{carreira2017quo} Features             &        & 26.7                 & 24.3                 & 23.2                 & 21.7                 & 30.8                 & 28.3                 & 26.1                 & 25.8                 & 35.4                 & 29.6                 & 22.5                 & 16.1                 & 35.7                 & 25.3                 & 20.1                 & 16.3 \\
FUTR \cite{gong2022future}                   & I3D \cite{carreira2017quo} Features             &        & 27.7                 & 24.6                 & 22.8                 & 22.0                 & 32.3                 & 29.9                 & 27.5                 & 25.9                 & 39.6                 & 27.5                 & 23.3                 & 17.8                 & 35.2                 & 24.9                 & 24.2                 & 15.3 \\
TempAgg \cite{sener2020temporal,sener2021technical}   & I3D \cite{carreira2017quo} Features + Inferred Actions$^\dagger$   &   & 37.1                 & 31.8                 & 30.1                 & 27.1                 & 39.8                 & 34.2                 & 31.9                 & 27.8                 & -                    & -                    & -                    & -                    & -      
              & -                    & -                    & - \\

\arrayrulecolor{black!20}\midrule\arrayrulecolor{black}

SRL \cite{qi2021self}                        & TCN \cite{lea2017temporal} Features             &        & -                    & -                    & -                    & -                    & -                    & -                    & -                    & -                    & 37.9                 & 28.8                 & 21.3                 & 11.1                 & 37.5                 & 24.1                 & 17.1                 & 9.1 \\
MAVAP \cite{loh2022long}                     & MS-TCN++ \cite{li2020ms} Features               &        & 69.1                 & 52.6                 & 43.5                 & 32.9                 & 68.0                 & 53.7                 & 44.6                 & 35.5                 & 43.5                 & 32.4                 & 27.5                 & 16.1                 & 42.5                 & 30.8                 & 25.4                 & 14.8 \\
AAP \cite{zhao2020diverse}                   & TCN \cite{lea2017temporal} Feat. + GT of Past Actions &  & 72.2                 & 62.4   
              & 56.2                 & 46.0                 & 74.1                 & 71.3                 & 65.3                 & 52.4   
              & 46.6                 & 35.6                 & 31.9                 & 21.4                 & 46.1                 & 36.4   
              & 33.1                 & 19.5 \\
Abu Farhar et al. \cite{abu2021long}                          & TCN \cite{lea2017temporal} Feat. + TCN Inferred Actions & & 25.9                 & 23.4                 & 22.4                 & 21.5                 & 29.7                 & 27.4                 & 25.6                 & 25.2                 & 34.8                 & 28.4                 & 21.8                 & 15.3                 & 34.4                 & 23.7                 & 19.0                 & 15.9 \\

\arrayrulecolor{black!20}\midrule\arrayrulecolor{black}

ANTICIPATR \cite{nawhal2022rethinking}       & ViT \cite{vaswani2017attention} Features        &        & 37.4                 & 32.0   
              & 30.3                 & 28.6                 & 39.9                 & 35.7                 & 32.1                 & 29.4   
              & 41.1                 & 35.0                 & 27.6                 & 27.3                 & 42.8                 & 42.3   & 28.5                 & 23.6 \\
DIFFANT \cite{zhong2023diffant}              & ViT \cite{vaswani2017attention} Features        &        & 25.3                 & 24.6                 & 24.4                 & 22.7                 & 32.1                 & 31.8                 & 31.2                 & 30.8                 & 36.1                 & 34.0                 & 30.5                 & 25.3                 & 34.1                 & 30.1                 & 26.3                 & 20.2 \\
S-GEAR \cite{diko2024semantically}           & ViT \cite{vaswani2017attention} Features        &               & -                     & -                   & -                    & -                    & -                    & -                    & -                     & -       & 41.0                 & 28.5             
   & 21.5                 & 15.3                 & 41.0                 & 27.8                 & 21.4                 & 16.7 \\

\arrayrulecolor{black!20}\midrule\arrayrulecolor{black}

Zhang et al. \cite{zhang2024object}          & SlowFast \cite{feichtenhofer2019slowfast} Features   &   & -                     & -                   & -                    & -                    & -                    & -                    & -                     & -                   & 37.4                 & 28.9                 & 24.2                 & 18.1                 & 28.0                  & 24.0                & 24.3                 & 19.3 \\
\bottomrule
\end{tabular}
}

\begin{tablenotes}
\item $\dagger$ The actions in the observed video segment are inferred by TempAgg own action detection model.
\end{tablenotes}
\end{threeparttable}
\end{table*}

\textbf{BLEU.} BLEU \cite{papineni2002bleu} is typically used for models that forecast future actions in natural language. It measures how well the model output approximates human-annotated description. BLEU calculates precision scores based on n-grams (contiguous sequences of n words). Precision is calculated separately for different n-gram lengths, typically from 1-grams (individual words) up to 4-grams. Precision for n-grams is the ratio of correctly predicted n-grams to the total number of n-grams in the model-generated language.

\textbf{METEOR.} Another evaluation metric for natural language measurement is METEOR \cite{banerjee2005meteor} which aims to improve upon BLEU by considering synonyms, stemming and paraphrasing in its matching process. It calculates an F1 score based on precision and recall of n-grams, while also applying a penalty for fragmented word ordering. As a result, METEOR provides a more nuanced evaluation of translation quality that often correlates better with human judgment.

\textbf{Edit Distance.} Edit distance (ED), also known as Levenshtein Distance, is firstly used in \cite{grauman2022ego4d} to measure the similarity between the predicted action sequence and ground truth action sequence. It calculates the minimum number of edits required to transform one sequence into another. These edits can include insertions, deletions, substitutions and transpositions of any two predicted actions. Intuitively, ED tolerates some error in the action order in predicted sequence. Note that a lower edit distance suggests a better performance.

\textbf{Success Rate.} Success rate (SR) is used in procedure planning problem to measure if the predicted in-between actions can successfully lead to the given goal. The plan is considered as a success if all predicted actions are the same as ground truth \cite{chang2020procedure}. Success rate is the number of successful plans over the number of all predicted plans.

\textbf{Step Accuracy.} Step accuracy measures the similarity of two sequences by measuring the accuracy at each timestep. This metric only focuses on each individual step regardless of the global assessment.

\textbf{Mean Intersection over Union.} Mean intersection over union (mIoU) treats the predicted sequence and ground truth sequence as bags of words regardless of the action order. It counts the numbers of actions separately in the intersection set and union set of prediction and ground truth. Then mIoU is calculated by the intersection number divided by the union number. This is the least strict metric among all sequence metrics.

\begin{table*}[t]
\renewcommand{\arraystretch}{1.3}
\caption{
    Action anticipation results on THUMOS14 \cite{THUMOS14}. The performance is measured by calibrated Average Precision (cAP) (\%), which is proposed in \cite{de2016online}. The best performance is highlighted by boldface.
}
\label{tab:res-thumos}
\centering
\begin{tabular}{lccccccccc}
\toprule
\textbf{Methods}              & \textbf{0.25s} & \textbf{0.5s}    & \textbf{0.75s}    & \textbf{1s}      & \textbf{1.25s}    & \textbf{1.5s}    & \textbf{1.75s}    & \textbf{2s}      & \textbf{Avg.} \\
\hline
ED \cite{gao2017red}         & 43.8          & 40.9    & 38.7     & 36.8    & 34.6     & 33.9    & 32.5     & 31.6    & 36.6 \\
RED \cite{gao2017red}        & 45.3          & 42.1    & 39.6     & 37.5    & 35.8     & 34.4    & 33.2     & 32.1    & 37.5 \\
Zhong et al. \cite{zhong2018unsupervised}    & -   & -   & -  & 44.1    & -        & -       & -        & -       & -    \\
TRN \cite{xu2019temporal}    & 45.1          & 42.4    & 40.7     & 39.1    & 37.7     & 36.4    & 35.3     & 34.3    & 38.9 \\
LAP-Net \cite{qu2020lap}     & 49.0          & 47.4    & 45.3     & 43.2    & 41.3     & 39.7    & 38.3     & 37.0    & 42.6 \\
OadTR \cite{wang2021oadtr}   & 50.2          & 49.3    & 48.1     & 46.8    & 45.3     & 43.9    & 42.4     & 41.1    & 45.9 \\
OadTR (Kinetics Pretraining) \cite{wang2021oadtr} & 59.8 & 58.5 & 56.6 & 54.6 & 52.6 & 50.5 & 48.6     & 46.8    & 53.5 \\
LSTR \cite{xu2021long}       & -             & -       & -        & -       & -        & -       & -        & -       & 50.1 \\
TeSTra \cite{zhao2022real}   & \textbf{64.7} & \textbf{61.8}    & \textbf{58.7}     & \textbf{55.7}    & \textbf{53.2}     & \textbf{51.1}    & \textbf{49.2}     & \textbf{47.8}    & 55.3 \\
MAT \cite{wang2023memory}    & -             & -       & -        & -       & -        & -       & -        & -       & 57.3 \\
MAT (Kinetics Pretraining) \cite{wang2023memory}  & - & - & - & -      & -        & -       & -        & -       & \textbf{58.2} \\
\bottomrule
\end{tabular}
\end{table*}

\begin{table*}[t]
\renewcommand{\arraystretch}{1.3}
\caption{
    Action anticipation results on TVSeries~ \cite{de2016online}. The performance is measured by calibrated Average Precision (cAP) (\%). The best performance is highlighted by boldface.
}
\label{tab:res-tvseries}
\centering
\begin{tabular}{lccccccccc}
\toprule
\textbf{Methods}             & \textbf{0.25s}     & \textbf{0.5s}      & \textbf{0.75s}       & \textbf{1s}      & \textbf{1.25s}      & \textbf{1.5s}       & \textbf{1.75s}       & \textbf{2s}       & \textbf{Avg.} \\
\hline
ED~ \cite{gao2017red}        & 78.5               & 78.0               & 76.3                 & 74.6             & 73.7                & 72.7                & 71.7                 & 71.0              & 74.6 \\
RED~ \cite{gao2017red}       & 79.2               & 78.7               & 77.1                 & 75.5             & 74.2                & 73.0                & 72.0                 & 71.2              & 75.1 \\
TRN~ \cite{xu2019temporal}   & 79.9               & 78.4               & 77.1                 & 75.9             & 74.9                & 73.9                & 73.0                 & 72.3              & 75.7 \\
LAP-Net~ \cite{qu2020lap}    & 82.6               & 81.3               & 80.0                 & 78.9             & 77.9                & 77.1                & 76.3                 & 75.5              & 78.7 \\
OadTR~ \cite{wang2021oadtr}  & 81.9               & 80.6               & 79.4                 & 78.2             & 77.1                & 76.0                & 75.2                 & 74.3              & 77.8 \\
OadTR (Kinetics Pretraining)~ \cite{wang2021oadtr} & \textbf{84.1} & \textbf{82.6} & \textbf{81.3} & \textbf{80.1} & \textbf{78.9} & \textbf{77.7} & \textbf{76.7}        & \textbf{75.7}     & 79.1 \\
LSTR~ \cite{xu2021long}      & -                  & -                  & -                    & -                & -                   & -                   & -                    & -                 & 80.8 \\
MAT~ \cite{wang2023memory}   & -                  & -                  & -                    & -                & -                   & -                   & -                    & -                 & 81.5 \\
MAT (Kinetics Pretraining)~ \cite{wang2021oadtr} & -         & -       & -                    & -                & -                   & -                   & -                    & -                 & \textbf{82.6} \\
\bottomrule
\end{tabular}
\end{table*}

\begin{table*}[t]
\renewcommand{\arraystretch}{1.3}
\caption{
    Precedure planning results on CrossTask \cite{zhukov2019cross}, COIN \cite{tang2019coin} and NIV \cite{alayrac2016unsupervised}. The performance is measured by three metrics -- Success Rate (SR) (\%), Accuracy (\%) and mIoU (\%). The best performance is highlighted by boldface.
}
\label{tab:res-ct-coin-niv}
\centering
\resizebox{\linewidth}{!}{
\begin{tabular}{lcccccccccccccccccc}
\toprule
& \multicolumn{6}{c}{\textbf{CrossTask}} & \multicolumn{6}{c}{\textbf{COIN}} & \multicolumn{6}{c}{\textbf{NIV}} \\
\cmidrule(lr){2-7} \cmidrule(l){8-13} \cmidrule(l){14-19}
\textbf{Methods}   & \multicolumn{3}{c}{\textbf{T=3}}    & \multicolumn{3}{c}{\textbf{T=4}}     & \multicolumn{3}{c}{\textbf{T=3}}     & \multicolumn{3}{c}{\textbf{T=4}}    & \multicolumn{3}{c}{\textbf{T=3}}    & \multicolumn{3}{c}{\textbf{T=4}} \\
\cmidrule(lr){2-4} \cmidrule(lr){5-7} \cmidrule(lr){8-10} \cmidrule(l){11-13}  \cmidrule(l){14-15}  \cmidrule(l){16-19}
  & \textbf{SR}     & \textbf{Acc.}     & \textbf{mIoU}     & \textbf{SR}     & \textbf{Acc.}     & \textbf{mIoU}  & \textbf{SR}   & \textbf{Acc.}   & \textbf{mIoU}     & \textbf{SR}       & \textbf{Acc.}   & \textbf{mIoU}     & \textbf{SR}      
  & \textbf{Acc.}   & \textbf{mIoU}   & \textbf{SR}       & \textbf{Acc.}     & \textbf{mIoU} \\
\hline
DDN \cite{chang2020procedure}       & 12.18 & 31.29 & 47.48 & 5.97  & 27.10 & 48.46 & -     & -     & -     & -     & -     & -     & -     & -     & -     & -     & -     & - \\
Bi et al. \cite{bi2021procedure}    & 21.27 & 49.46 & 61.70 & 16.41 & 43.05 & 60.93 & -     & -     & -     & -     & -     & -     & -     & -     & -     & -     & -     & - \\
P$^3$IV \cite{zhao2022p3iv}         & 23.34 & 49.96 & \textbf{73.89} & 13.40 & 44.16 & \textbf{70.01} & 15.40 & 21.70 & \textbf{76.30} & 11.30 & 18.90 & \textbf{70.50} & 24.68 & 49.01 & \textbf{74.29} & 20.14 & 38.36 & \textbf{67.29} \\
PlaTe \cite{sun2022plate}           & 16.00 & 36.17 & 65.91 & 14.00 & 35.29 & 55.36 & -     & -     & -     & -     & -     & -     & -     & -     & -     & -     & -     & - \\
PDPP (Base) \cite{wang2023pdpp}     & 26.47 & 55.35 & 58.95 & 15.40 & 49.42 & 56.99 & -     & -     & -     & -     & -     & -     & -     & -     & -     & -     & -     & - \\
PDPP (HowTo100) \cite{wang2023pdpp} & \textbf{37.20} & \textbf{64.67} & 66.57 & \textbf{21.48} & \textbf{57.82} & 65.13 & \textbf{21.33} & \textbf{45.62} & 51.82 & \textbf{14.41} & \textbf{44.10} & 51.39 & \textbf{31.25} & \textbf{49.26} & 57.92 & \textbf{26.72} & \textbf{48.92} & 59.04 \\
\bottomrule
\end{tabular}
}
\end{table*}

\subsection{Time and Duration Prediction Metrics}

 We assume the predicted time and ground truth time of the $i$-th sample are $\hat{t}_i$ and $t_i$, and there are $N$ data samples.
 
\textbf{Root Mean Square Error.} Root mean square error (RMSE) is a widely-used metric for regression models. It is formulated as
\begin{equation}
    RMSE = \sqrt{\frac{1}{N}\sum_{i=1}^N\left(t_i-\hat{t}_i\right)^2}.
\end{equation}

\textbf{Mean Absolute Error.} Mean Absolute Error (MAE) directly calculates the difference of model output and ground truth. It is formulated as
\begin{equation}
    MAE = \frac{1}{N}\sum_{i=1}^N\left|t_i-\hat{t}_i\right|.
\end{equation}

\begin{table*}[t]
\renewcommand{\arraystretch}{1.3}
\caption{
    Comparison of results on the Epic-Kitchens 55 \cite{damen2018scaling} validation and test sets with anticipation time $\tau_a=1$ second. We report the top-5 accuracy (\%) of verb, noun and action. For the validation set, we additionally report mean Average Precision (mAP) (\%) for frequent, rare and all action categories. For each model, we report the best result over different choices of architecture and input modalities reported in the corresponding paper. \\
}
\label{tab:res-ek55}
\centering
\begin{threeparttable}
\begin{tabular}{lccccccccc|ccc}
\toprule
\multirow{2}{*}{\textbf{Methods}} & \multicolumn{3}{c}{\textbf{Val (Top-5 Acc.)}} & \multicolumn{3}{c}{\textbf{S1 (Top-5 Acc.)}} & \multicolumn{3}{c}{\textbf{S2 (Top-5 Acc.)}} & \multicolumn{3}{c}{\textbf{Val (mAP)}} \\
\cmidrule(lr){2-4} \cmidrule(lr){5-7} \cmidrule(lr){8-10} \cmidrule(lr){11-13}
& \textbf{Verb} & \textbf{Noun} & \textbf{Action} & \textbf{Verb} & \textbf{Noun} & \textbf{Action} & \textbf{Verb} & \textbf{Noun} & \textbf{Action} & \textbf{Freq} & \textbf{Rare} & \textbf{All} \\
\midrule
VNMCE \cite{furnari2018leveraging}                      & 74.1           & 39.1           & 26.0           & -           & -            & -           & -           & -           & -         & -              & -              & -       \\
Miech et al. \cite{miech2019leveraging}                 & 73.5           & 34.6           & 19.1           & -           & -            & -           & 70.0        & 32.2        & 19.3      & -              & -              & -       \\
Ke et al. \cite{ke2019time}                             & -              & -              & -              & -           & -            & 33.9        & -           & -           & -         & -              & -              & -       \\
RU-LSTM \cite{furnari2019would, furnari2020rolling}     & \textbf{79.6}  & 51.8           & 35.3           & 79.6        & 51.0         & 33.7        & 69.6        & 34.4        & 21.2      & -              & -              & -       \\
Ego-OMG \cite{dessalene2020egocentric}                  & -              & -              & -              & -           & -            & 34.5        & -           & -           & 23.8      & -              & -              & -       \\
IAI \cite{zhang2020egocentric}                          & 79.2           & 50.6           & 33.4           & -           & -            & -           & 70.1        & 35.5        & 21.4      & -              & -              & -       \\
Liu et al. \cite{liu2020forecasting}                    & -              & -              & -              & 79.2        & 52.0         & 34.3        & 71.8        & 39.0        & 23.7      & -              & -              & -       \\
Ego-Topo \cite{nagarajan2020ego}                        & -              & -              & -              & -           & -            & -           & -           & -           & -         & 56.9           & 29.2           & 38.0    \\
ImageRNN \cite{wu2020learning}                          & -              & -              & 35.6           & 79.7        & 52.1         & 35.0        & 70.7        & 35.8        & 22.2      & -              & -              & -       \\
TempAgg \cite{sener2020temporal,sener2021technical}     & -              & -              & 36.4           & 79.7        & 54.0         & 36.1        & 70.1        & 37.8        & 23.4      & -              & -              & -       \\
SF-RU-LSTM \cite{osman2021slowfast}                     & -              & -              & 36.1           & -           & -            & -           & -           & -           & -         & -              & -              & -       \\
Zatsarynna et al. \cite{zatsarynna2021multi}            & -              & -              & -              & 79.5        & 51.9         & 34.4        & 72.0        & 36.7        & 21.7      & -              & -              & -       \\
Camporese et al. \cite{camporese2021knowledge}          & -              & -              & 35.9           & 79.6        & 52.9         & 35.0        & 70.7        & 36.6        & 21.3      & -              & -              & -       \\
Fernando et al. \cite{fernando2021anticipating}         & -              & -              & 39.2           & -           & -            & -           & -           & -           & -         & -              & -              & -       \\
AVT \cite{girdhar2021anticipative}                      & -              & -              & 31.7$^{\dagger}$ & -         & -            & -           & -           & -           & -         & -              & -              & -       \\
SRL \cite{qi2021self}                                   & 78.9           & 47.7           & 31.7           & 79.6        & 52.0         & 34.6        & 71.9        & 36.8        & 22.1      & -              & -              & -       \\
Mahmud \cite{mahmud2021prediction}                      & -              & -              & 29.6           & -           & -            & -           & -           & -           & -         & -              & -              & -       \\
MM-Transformer \cite{roy2021action}                     & 79.5           & 52.1           & 35.7           & 78.6        & 57.7         & 30.8        & 70.4        & \textbf{44.2} & 20.0    & -              & -              & -       \\
A-ACT \cite{gupta2022act}                               & -              & -              & 34.8           & 80.1        & 53.5         & 36.7        & 71.1        & 36.5        & 23.5      & -              & -              & -       \\
Tai et al. \cite{tai2022unified}                        & 78.7           & 47.6           & 32.0           & -           & -            & -           & -           & -           & -         & -              & -              & -       \\
DCR \cite{xu2022learning}                               & -              & -              & 41.2           & -           & -            & \textbf{38.5} & -         & -           & \textbf{24.8} & -          & -              & -       \\
Roy et al. \cite{roy2022predicting}                     & 77.1           & \textbf{53.8}  & \textbf{43.9}  & \textbf{82.6} & \textbf{58.0} & 38.3     & \textbf{73.1} & 41.6      & 24.2      & -              & -              & -       \\
HRO \cite{liu2022hybrid}                                & -              & -              & 37.4           & -           & -            & -           & -           & -           & -         & -              & -              & -       \\
ANTICIPATR \cite{nawhal2022rethinking}                  & -              & -              & -              & -           & -            & -           & -           & -           & -         & 58.1           & 29.1           & 39.1    \\
OCT \cite{liu2022joint}                                 & 73.9           & 45.9           & 24.4           & -           & -            & -           & -           & -           & -         & -              & -              & -       \\
DIFFANT \cite{zhong2023diffant}                         & -              & -              & -              & -           & -            & -           & -           & -           & -         & 55.0           & 31.0           & 38.7    \\
AntGPT \cite{zhao2024antgpt}                            & -              & -              & -              & -           & -            & -           & -           & -           & -         & \textbf{58.8}  & \textbf{31.9}  & \textbf{40.1} \\
S-GEAR \cite{diko2024semantically}                      & -              & -              & 43.2           & -           & -            & -           & -           & -           & -         & -              & -              & -       \\
\bottomrule
\end{tabular}

\begin{tablenotes}
\item $\dagger$\ AVT results on Epic-Kitchens 55 only use RGB frames as input.
\end{tablenotes}
\end{threeparttable}
\end{table*}

\begin{table*}[t]
\renewcommand{\arraystretch}{1.3}
\caption{
    Comparison of results on the Epic-Kitchens 100 \cite{damen2020rescaling} validation and test sets with anticipation time $\tau_a=1$ second. We report the mean top-5 recall (\%) of verb, noun and action. For each model, we report the best result over different choices of architecture and input modalities reported in the corresponding paper. \\
}
\label{tab:res-ek100}
\centering
\begin{tabular}{lcccccc}
\toprule
\multirow{3}{*}{\textbf{Methods}} & \multicolumn{3}{c}{\textbf{Val}} & \multicolumn{3}{c}{\textbf{Test}} \\
\cmidrule(lr){2-4} \cmidrule(lr){5-7}
& \textbf{Verb} & \textbf{Noun} & \textbf{Action} & \textbf{Verb} & \textbf{Noun} & \textbf{Action} \\
\midrule
TransAction \cite{gu2021transaction}              & 35.0            & 35.5              & 16.6               & 36.2               & 32.2             & 13.4 \\
Zatsarynna et al. \cite{zatsarynna2021multi}                        & -               & -                 & -                  & 21.5               & 26.8             & 11.0 \\
AVT \cite{girdhar2021anticipative}                & -               & -                 & 15.9               & -                  & -                & 12.6 \\
Tai el al. \cite{tai2022unified}                             & 28.1            & 31.2              & 14.8               & -                  & -                & -    \\
DCR \cite{xu2022learning}                         & -               & -                 & 18.3               & -                  & -                & 17.3 \\
Roy et al. \cite{roy2022predicting}               & -               & -                 & -                  & 31.4             & 30.1             & 14.3 \\
MeMViT \cite{wu2022memvit}                        & 32.2            & 37.0              & 17.7               & -                  & -                & -    \\
TeSTra \cite{zhao2022real}                        & 30.8            & 35.8              & 17.6               & -                  & -                & -    \\
OCT \cite{liu2022joint}                           & 21.9            & 27.6              & 12.4               & -                  & -                & -    \\
MAT \cite{wang2023memory}                         & 35.0            & 38.8              & 19.5               & -                  & -                & -    \\
AFFT \cite{zhong2023anticipative}                 & 22.8            & 34.6              & 18.5               & 20.7               & 31.8             & 14.9 \\
RAFTformer \cite{girase2023latency}               & 33.8            & 37.9              & 19.1               & 30.1               & 34.1             & 15.4 \\
NAOGAT \cite{thakur2024leveraging}                & 25.3            & 27.9              & 10.7               & -                  & -                & -    \\
InAViT \cite{roy2024interaction}                  & 52.5            & 51.9              & 25.9               & \textbf{49.1}      & \textbf{50.0}    & \textbf{23.8} \\
UADT \cite{guo2024uncertainty}                    & 43.5            & 46.6              & 23.0               & -                  & -                & -    \\
PlausiVL \cite{mittal2024can}                     & \textbf{55.6}   & \textbf{54.2}     & \textbf{27.6}      & -                  & -                & -    \\
S-GEAR \cite{diko2024semantically}                & 30.2            & 37.0              & 19.9               & 26.6               & 32.6             & 15.5 \\
\bottomrule
\end{tabular}

\end{table*}

\section{Quantitative Comparison of Methods}

\subsection{Action Anticipation on Breakfast Actions and 50Salads}

The action anticipation results on Breakfast Actions \cite{kuehne2014language} and 50Salads \cite{stein2013combining} are demonstrated in Table \ref{tab:res-ba-50s}. Breakfast Actions and 50Salads are two early exocentric video dataset with labels of action categories, start time and end time. They are typically used for evaluation of long-term action anticipation and action segment action anticipation. There are also a few models for other anticipation tasks (next action or time anticipation, short-term action anticipation and fixed-interval action anticipation) validated on the two datasets, which are also included in Table \ref{tab:res-ba-50s}. Typically, the input to existing models may be the ground truth action categories in the observed videos, the pre-computed video features from pre-trained encoders or the combination of the two inputs. We group prior work by the type of input to provide a clear comparison. It's worth noting that there are some work \cite{miech2019leveraging, mehrasa2019variational,morais2020learning,ke2021future,fernando2021anticipating,roy2021action,roy2022action,xu2022learning,manousaki2023vlmah} tested on Breakfast Actions and 50Salads with different setting (e.g., evaluating the performance of action anticipation after $\tau_a=1$ second). They are not covered in this section.

\subsection{Action Anticipation on THUMOS14}

The results on THUMOS14 \cite{THUMOS14} are listed in Table \ref{tab:res-thumos}. THUMOS14 is mainly used for evaluation of short-term action anticipation and fixed-interval action anticipation. Prior models typically evaluate the performance at different anticipation time ranging from 0.25 second to 2 seconds. Note that JOADAA \cite{guermal2024joadaa} is tested with a different setting so it's not included in the table.

\subsection{Action Anticipation on TVSeries}

The results on TVSeries \cite{de2016online} are shown in Table \ref{tab:res-tvseries}. TVSeries is mostly used as the testbed together with THUMOS14 for evaluation of short-term action anticipation and fixed-interval action anticipation. They also share the same setting for validation.

\subsection{Action Anticipation on CrossTask, COIN and NIV}

The results on CrossTask \cite{zhukov2019cross}, COIN \cite{tang2019coin} and NIV \cite{alayrac2016unsupervised} are demonstrated in Table \ref{tab:res-ct-coin-niv}. All the models are designed for procedure planning.

\subsection{Action Anticipation on Epic-Kitchens 55/100}

The results on Epic-Kitchens 55/100 \cite{damen2018scaling,damen2020rescaling} are demonstrated in Table \ref{tab:res-ek55} and Table \ref{tab:res-ek100}. Multiple values for anticipation time may be investigated in Epic-Kitchens 55. We only summarize the performance when anticipation time is equal to 1 second, which is usually considered as the primary setting. For Epic-Kitchen 100, there's a default setting for action anticipation that is directly used by prior work. Most models evaluated on Epic-Kitchens are designed for short-term action anticipation. However, we observe that there are some approaches for fixed-interval anticipation and long-term action anticipation reporting the performance on Epic-Kitchens under the same setting. We also include these approaches for a thorough comparison. Note that some work evaluates their models on Epic-Kitchens 55 using other metrics instead of top-5 accuracy \cite{guan2020generative,zhao2020diverse,ke2021future,ghosh2023text,tan2023multiscale}. Table \ref{tab:res-ek55} and Table \ref{tab:res-ek100} don't cover these approaches.

\begin{table*}[t]
\renewcommand{\arraystretch}{1.3}
\caption{
    Comparison of results on the EGTEA Gaze+ \cite{li2018eye} dataset with anticipation time $\tau_a=0.5$ second. We report the top-1 accuracy (\%), mean class accuracy (\%) of verb, noun and action as well as top-5 accuracy of action. We also show mean Average Precision (mAP) (\%) of frequent, rare and all action categories. For each model, we report the best result over different choices of architecture and input modalities reported in the corresponding paper. \\
}
\label{tab:res-egtea}
\centering
\begin{tabular}{lccccccc|ccc}
\toprule
\multirow{2}{*}{\textbf{Methods}} & \multicolumn{3}{c}{\textbf{Top-1 Acc.}} & \multicolumn{3}{c}{\textbf{Mean Class Acc.}} & \textbf{Top-5 Acc.} & \multicolumn{3}{c}{\textbf{mAP}} \\
\cmidrule(lr){2-4} \cmidrule(lr){5-7} \cmidrule(lr){8-8} \cmidrule(lr){9-11}
& \textbf{Verb} & \textbf{Noun} & \textbf{Action} & \textbf{Verb} & \textbf{Noun} & \textbf{Action} & \textbf{Action} & \textbf{Freq} & \textbf{Rare} & \textbf{All} \\
\midrule
RU-LSTM \cite{furnari2019would, furnari2020rolling}    & -            & -              & -              & -             & -           & -               & 71.8              & -              & -            & -     \\
Liu et al. \cite{liu2020forecasting}                   & 49.0         & 45.5           & 36.6           & 32.5          & 32.7        & 25.3            & -                 & -              & -            & -     \\
Ego-Topo \cite{nagarajan2020ego}                       & -            & -              & -              & -             & -           & -               & -                 & 80.7           & 54.7         & 73.5  \\
ImageRNN \cite{wu2020learning}                         & -            & -              & -              & -             & -           & -               & 72.3              & -              & -            & -     \\
SF-RU-LSTM \cite{osman2021slowfast}                    & -            & -              & -              & -             & -           & -               & 72.2              & -              & -            & -     \\
Camporese et al. \cite{camporese2021knowledge}         & -            & -              & -              & -             & -           & -               & 75.2              & -              & -            & -     \\
AVT \cite{girdhar2021anticipative}                     & 54.9         & 52.2           & 43.0           & 49.9          & 48.3        & 35.2            & -                 & -              & -            & -     \\
SRL \cite{qi2021self}                                  & -            & -              & -              & -             & -           & -               & 78.0              & -              & -            & -     \\
AVT+Distillation \cite{ghosh2023text}                  & -            & -              & 45.5           & -             & -           & -               & -                 & -              & -            & -     \\
Roy et al. \cite{roy2022predicting}                    & 64.8         & 65.3           & 49.8           & 63.4          & 55.6        & 37.4            & -                 & -              & -            & -     \\
HRO \cite{liu2022hybrid}                               & -            & -              & -              & -             & -           & -               & \textbf{79.2}     & -              & -            & -     \\
ANTICIPATR \cite{nawhal2022rethinking}                 & -            & -              & -              & -             & -           & -               & -                 & 83.3           & 55.1         & 76.8  \\
AFFT \cite{zhong2023anticipative}                      & 53.4         & 50.4           & 42.5           & 42.4          & 44.5        & 35.2            & 72.5              & -              & -            & -     \\
DIFFANT \cite{zhong2023diffant}                        & -            & -              & -              & -             & -           & -               & -                 & 83.5           & 61.4         & 77.3  \\
InAViT \cite{roy2024interaction}                       & \textbf{79.3}& \textbf{77.6}  & \textbf{67.8}  & \textbf{79.2} & \textbf{76.9}  & \textbf{58.2}& -                 & -              & -            & -     \\
Zhang et al. \cite{zhang2024object}                    & -            & -              & 44.6           & -             & -           & 36.4            & -                 & -              & -            & -     \\
AntGPT \cite{zhao2024antgpt}                           & -            & -              & -              & -             & -           & -               & -                 & \textbf{84.8}  & \textbf{72.9}& \textbf{80.2}  \\
S-GEAR \cite{diko2024semantically}                     & -            & -              & 45.7           & -             & -           & -               & -                 & -              & -            & -     \\
\bottomrule
\end{tabular}

\end{table*}

\begin{table*}[t]
\renewcommand{\arraystretch}{1.3}
\caption{
    Comparison of results on Ego4D \cite{grauman2022ego4d}. We report the Edit Distance (ED), mean Average Precision (mAP) (\%) of verb, noun and action. $\downarrow$ denotes a lower value in ED indicates a better performance. For each model, we report the best result over different choices of architecture and input modalities reported in the corresponding paper. \\
}
\label{tab:res-ego4d}
\centering
\begin{tabular}{lccccccccc}
\toprule
\multirow{2}{*}{\textbf{Methods}} & \multicolumn{3}{c}{\textbf{Val (ED $\downarrow$ )}} & \multicolumn{3}{c}{\textbf{Test (ED $\downarrow$ )}} & \multicolumn{3}{c}{\textbf{Test (mAP $\uparrow$ )}} \\
\cmidrule(lr){2-4} \cmidrule(lr){5-7} \cmidrule(lr){8-10}
& \textbf{Verb} & \textbf{Noun} & \textbf{Action} & \textbf{Verb} & \textbf{Noun} & \textbf{Action} & \textbf{Verb} & \textbf{Noun} & \textbf{Action} \\
\midrule
\multicolumn{10}{c}{\textit{Ego4D v1}} \\
\midrule
SlowFast+Transformer \cite{grauman2022ego4d}    & 74.5             & 77.9              & \textbf{94.1}     & 73.9             & 78.0           & 94.3                & -              & -                & -     \\
Video+CLIP \cite{das2022video+}                 & 71.3             & 74.7              & -                 & 73.9             & 76.9           & 94.1                & -              & -                & -     \\
MVP \cite{tan2023multiscale}                    & -                & -                 & -                 & 72.4             & 80.9           & 94.3                & -              & -                & -     \\
H3M+I-CVAE \cite{mascaro2023intention}          & -                & -                 & -                 & 74.1             & 73.9           & 93.0                & -              & -                & -     \\
EgoT2 \cite{xue2023egocentric}                  & -                & -                 & -                 & 72.0             & 76.0           & 93.0                & -              & -                & -     \\
HierVL \cite{ashutosh2023hiervl}                & -                & -                 & -                 & 72.4             & 73.5           & 92.8                & -              & -                & -     \\
NAOGAT \cite{thakur2024leveraging}              & -                & -                 & -                 & -                & -              & -                   & \textbf{32.8}  & \textbf{33.2}    & \textbf{15.1}  \\
Zhang et al. \cite{zhang2024object}                       & 72.8             & 77.1              & -                 & 72.7             & 74.0           & 92.9                & -              & -                & -     \\
AntGPT \cite{zhao2024antgpt}                    & -                & -                 & -                 & \textbf{65.8}    & \textbf{65.5}  & \textbf{88.1}       & -              & -                & -     \\
PlausiVL \cite{mittal2024can}                   & \textbf{67.9}    & \textbf{68.1}     & -                 & -                & -              & -                   & -              & -                & -     \\
\midrule
\multicolumn{10}{c}{\textit{Ego4D v2}} \\
\midrule
AntGPT \cite{zhao2024antgpt}                    & -                & -                 & -                 & 65.0             & 65.0           & 87.7                & -              & -                & -     \\
\bottomrule
\end{tabular}

\end{table*}

\subsection{Action Anticipation on EGTEA Gaze+}

The results on EGTEA Gaze+ \cite{li2018eye} are listed in Table \ref{tab:res-egtea}. EGTEA Gaze+ is mainly used for short-term action anticipation, long-term action anticipation and fixed-interval action anticipation. Though different anticipation time may be applied in anticipation problem, we show the results with anticipation time $\tau_a=0.5$ second, which is the most widely-used setting. RAFTformer \cite{girase2023latency} uses a different metric (top-5 recall) so it's not included in Table \ref{tab:res-egtea}.

\subsection{Action Anticipation on Ego4D}

The action anticipation results on Ego4D \cite{grauman2022ego4d} long-term action anticipation benchmark are listed in Table \ref{tab:res-ego4d}. Most existing work uses the default setting on this benchmark and measures the performance by Edit Distance (ED). AntGPT conducts experiments on both the first version and second version of Ego4D. We include both of them in the table for thorough comparison.

\end{document}